\documentclass[lettersize,journal]{IEEEtran}
\usepackage{amsmath,amsfonts}
\usepackage{amssymb}
\usepackage{subfigure}
\usepackage{array}
\usepackage{textcomp}
\usepackage{stfloats}
\usepackage{url}
\usepackage{verbatim}
\usepackage{graphicx}
\usepackage{booktabs}
\usepackage{tabularx}
\usepackage[table]{xcolor}
\usepackage{multirow}
\usepackage{makecell}
\usepackage{url}
\usepackage[pagebackref,breaklinks=true,colorlinks,citecolor=blue,urlcolor=blue,linkcolor=blue,bookmarks=false]{hyperref}
\usepackage{orcidlink}

\newcommand*{\belowrulesepcolor}[1]{%
  \noalign{%
    \kern-\belowrulesep 
    \begingroup 
      \color{#1}%
      \hrule height\belowrulesep 
    \endgroup 
    \vspace{-0.03mm}
  }%
} 
\newcommand*{\aboverulesepcolor}[1]{%
  \noalign{%
  \vspace{-0.03mm}
    \begingroup 
      \color{#1}%
      \hrule height\aboverulesep 
    \endgroup 
    \kern-\aboverulesep 
  }%
}

\begin{document}

\title{Rethinking Multi-Label Image Classification With Deep Learning: Taxonomy, Challenge, and Outlook}

\author{

{
Xuelin Zhu$^{\orcidlink{0000-0001-7676-2843}}$, 
Xiu-Shen Wei$^{\orcidlink{0000-0002-8200-1845}}$, 
Jiawei Ge$^{\orcidlink{0000-0001-7268-7815}}$, 
Shuai Xu$^{\orcidlink{0000-0002-5734-3616}}$, 
Bing Wang$^{\orcidlink{0000-0003-0977-0426}}$
}

\thanks{This work is supported by the National Natural Science Foundation of China (42301520), the Research Grants Council of Hong Kong (25206524, 15212925), the Unmanned Autonomous Systems Research Centre (P0049516), and the Smart Cities Research Institute (P0051028). \textit{(Corresponding author: Bing Wang.)}}

\thanks{Xuelin Zhu and Bing Wang are with the Spatial Intelligence Group, The Hong Kong Polytechnic University, Hong Kong, China (e-mail: zhuxuelin23@gmail.com; bingwang@polyu.edu.hk).}

\thanks{Xiu-Shen Wei is with the School of Computer Science and Engineering, Southeast University, Nanjing, China (e-mail: weixs@seu.edu.cn).}

\thanks{Jiawei Ge is with the School of Cyber Science and Engineering, Southeast University, Nanjing, China (e-mail: jiawei\_ge@seu.edu.cn).}

\thanks{Shuai Xu is with the College of Computer Science and Technology, Nanjing University of Aeronautics and Astronautics, Nanjing, China (e-mail: xushuai7@nuaa.edu.cn).}


}
\markboth{}%
{Shell \MakeLowercase{\textit{et al.}}: A Sample Article Using IEEEtran.cls for IEEE Journals}


\maketitle

\begin{abstract}
Multi-label image classification (MLIC), a fundamental task in computer vision, focuses on identifying multiple objects or concepts within an image, underpinning numerous read-world applications, such as autonomous driving, disease diagnosis, recommendation system, and mobile service robot. Over the past decade, deep learning paradigms based on convolutional neural networks, recurrent neural networks, and Transformers have significantly advanced this field, owing to their powerful capability in visual representation and relationship modeling. These advances have markedly improved the robustness, scalability, and generalization ability of MLIC models across diverse datasets and application domains. In this survey, we provide a comprehensive review of the deep learning-based literature on MLIC. Concretely, we first revisit the background, including problem definition, datasets, backbones and evaluation metrics. Next, we develop a plausible taxonomy for the deep learning-based MLIC approaches, organizing them into six groups: region-oriented methods, label-oriented methods, architecture-oriented methods, representation-oriented methods, learning-oriented methods, and data-oriented methods. Finally, we provide an insightful exposition of the underlying learning game in MLIC and its implications for other vision domains, and we empirically summarize the key challenges and research directions in MLIC while outlining promising avenues for future development. We believe this survey offers the research community a holistic and systematic perspective on MLIC, thereby facilitating subsequent exploration and innovation in this field and beyond.

\end{abstract}

\begin{IEEEkeywords}
Multi-label image classification, Multiple object recognition, Image tagging, Image annotation, Survey.
\end{IEEEkeywords} 




\section{Introduction}

\IEEEPARstart{M}{ulti}-label image classification, which aims to recognize multiple objects or concepts within an image, has attracted much research attention for many years, driven by numerous practical applications such as recommendation system \cite{yang2015pinterest}, mobile robot \cite{cartucho2018robust}, disease diagnosis \cite{ge2021multi}, and autonomous driving \cite{li2021ml}. As a fundamental task in computer vision (CV), MLIC underpins a wide range of visual research, including
vision-language pretraining (VLP) \cite{huang2024tag2text} and self-supervised learning \cite{zhu2023multi}, 
Beyond generic benchmarks, MLIC techniques have been widely applied in domain-specific imaging scenarios for identifying specialized labels, such as medical imaging \cite{pal2025label}, remote sensing \cite{dai2022feature}, 
and weather imaging \cite{afxentiou2023multi}.
Overall, MLIC plays a pivotal role in both academic research and real-world systems, driving progress across diverse fields and industries.


Single-label image classification (SLIC) \cite{he2016deep} refer to assigning a single label to an image containing one dominant object. In contrast, MLIC predicts multiple labels for images characterized by multiple objects and diverse scenes, making the task more challenging. Beyond inheriting from SLIC the reliance on robust visual feature extraction, MLIC encounters many task-specific challenges. Typically, objects in an image tend to follow intricate co-occurrence patterns, the pixel regions they occupy are spatially correlated, their associated textual labels are semantically interdependent, and the surrounding background carries rich contextual cues. These characteristics have a pronounced impact on the accuracy and robustness of label prediction, influence architectural design choices and model learning strategies, and collectively shape and drive the evolution of research in the MLIC community.

For MLIC, the research community has undergone a technical shift from statistical machine learning to deep learning. Early methods mainly rely on hand-crafted visual features and statistical learning models \cite{devkar2017survey} that learn decision boundaries and produce label rankings. With the advent of deep learning, convolutional neural networks (CNNs) \cite{he2016deep}, recurrent neural networks (RNNs) \cite{hochreiter1997long}, and Transformers \cite{vaswani2017attention,dosovitskiy2020image} have become dominant across diverse vision tasks, owing to their exceptional capabilities in feature extraction and sequence modeling, as well as their flexibility in handling complex data structures. Accordingly, a substantial body of work has been devoted to developing network architectures and learning strategies for MLIC over the past decade. However, the literature remains fragmented and uncoordinated, highlighting the urgent need for a comprehensive and up-to-date review of MLIC to better guide future research.


The only survey \cite{devkar2017survey} dedicated exclusively to MLIC is now outdated, covering just 37 studies published before 2017 that were dominated by statistical learning methods. Notably, it misses the rapid progress of MLIC in the era of deep learning. In this review, we distill key techniques with broad impact on MLIC, and employ them as main threads to organize the literature. To the best of our knowledge, this is the first survey dedicated to deep learning-based MLIC methods. Our main contributions are threefold: 1) We provide a comprehensive and systematic review of deep learning-based methods, offering readers a panoramic view on MLIC; 2) We reveal a latent learning game in MLIC and generalize it to a broad range of visual tasks, offering insights that may shape future vision studies; 3) We conduct experimental validation and analysis on key challenges and promising directions, and articulate a forward-looking outlook on the future of MLIC in the era of VLP and large language models (LLMs).


The remainder of this survey is organized as follows. Sec. \ref{sec:background} introduces the background of MLIC, including problem definition, datasets, backbones, and evaluation metrics. Sec. \ref{sec:method} then comprehensively reviews, categorizes and summarizes deep learning-based MLIC methods. Finally, Sec. \ref{sec:future} outlines future directions, and Sec. \ref{sec:conclusion} concludes the survey.

\section{Background}
\label{sec:background}

In this section, we first present the formal problem definition of the MLIC task, followed by an introduction to commonly used benchmark datasets, backbones, and evaluation metrics.

\textbf{Problem Definition.} For convenience, we denote a multi-label image dataset as $\mathcal{D}$ and its label set as $\mathcal{C}=\{c_0,c_1,\cdots,c_{N-1}\}$, where $N$ is the total number of labels. We use $\mathcal{I}$ and $\mathcal{Y}$ to represent the RGB image space and the multi-hot encoding space. For a specific sample ($I\in\mathcal{I},\mathbf{y}\in\mathcal{Y}$), $I$ is an image and $\mathbf{y}=[y_0,y_1,\cdots,y_{N-1}]^\top$ is its annotations, where $y_i\in\{0,1\}$ for $i\in\{0,1,\cdots,n-1\}$ is a binary indicator, $y_i=1$ if the label $c_i$ presents in the image $I$ and $0$ otherwise. Thus, the dataset can be defined as $\mathcal{D}=\{(I^m,\mathbf{y}^m)\}_{m=1}^M$ with $M$ being the total number of samples. The goal of MLIC is to learn a function from the dataset $\mathcal{D}$ that maps images from the RGB space to labels in the multi-hot encoding space, represented as: $f:\mathcal{I}\rightarrow\mathcal{Y}$. Giving a new image $I'$, the output of the mapping function $\mathbf{y}'=f(I')\in\mathcal{Y}$ serves as its predicted labels.

\textbf{Datasets.} In MLIC, Pascal VOC \cite{everingham2010pascal}, MS-COCO \cite{lin2014microsoft}, and NUS-WIDE \cite{chua2009nus} are the most widely used benchmarks for evaluating model performance. These datasets differ in the scale of images and the number of labels, and together they cover a broad spectrum of real-world multi-label scenarios. Detailed statistics (e.g., number of images, categories), dataset splits, and additional benchmarks used in MLIC are provided in the supplementary materials.

\textbf{Backbones.} In MLIC, most frameworks are built on deep convolutional networks such as ResNet \cite{he2016deep} and transformer-based architectures such as Vision Transformer (ViT) \cite{dosovitskiy2020image}, typically pretrained on ImageNet \cite{deng2009imagenet}. These two backbone families form the foundation of modern MLIC pipelines. ResNets provide strong hierarchical feature extractors with good inductive biases for local patterns, while ViTs offer flexible, global context modeling through self-attention. Their details and additional backbone networks adopted in MLIC are provided in the supplementary materials.

\textbf{Evaluation Metrics.} In MLIC, the most common evaluation metrics include the average precision (AP) per label and the mean average precision (mAP) over all labels, measuring label-wise performance and overall performance, respectively. Their details and additional evaluation metrics used in MLIC are provided in the supplementary materials.

\begin{figure*}
    \centering
    \includegraphics[width=1\linewidth]{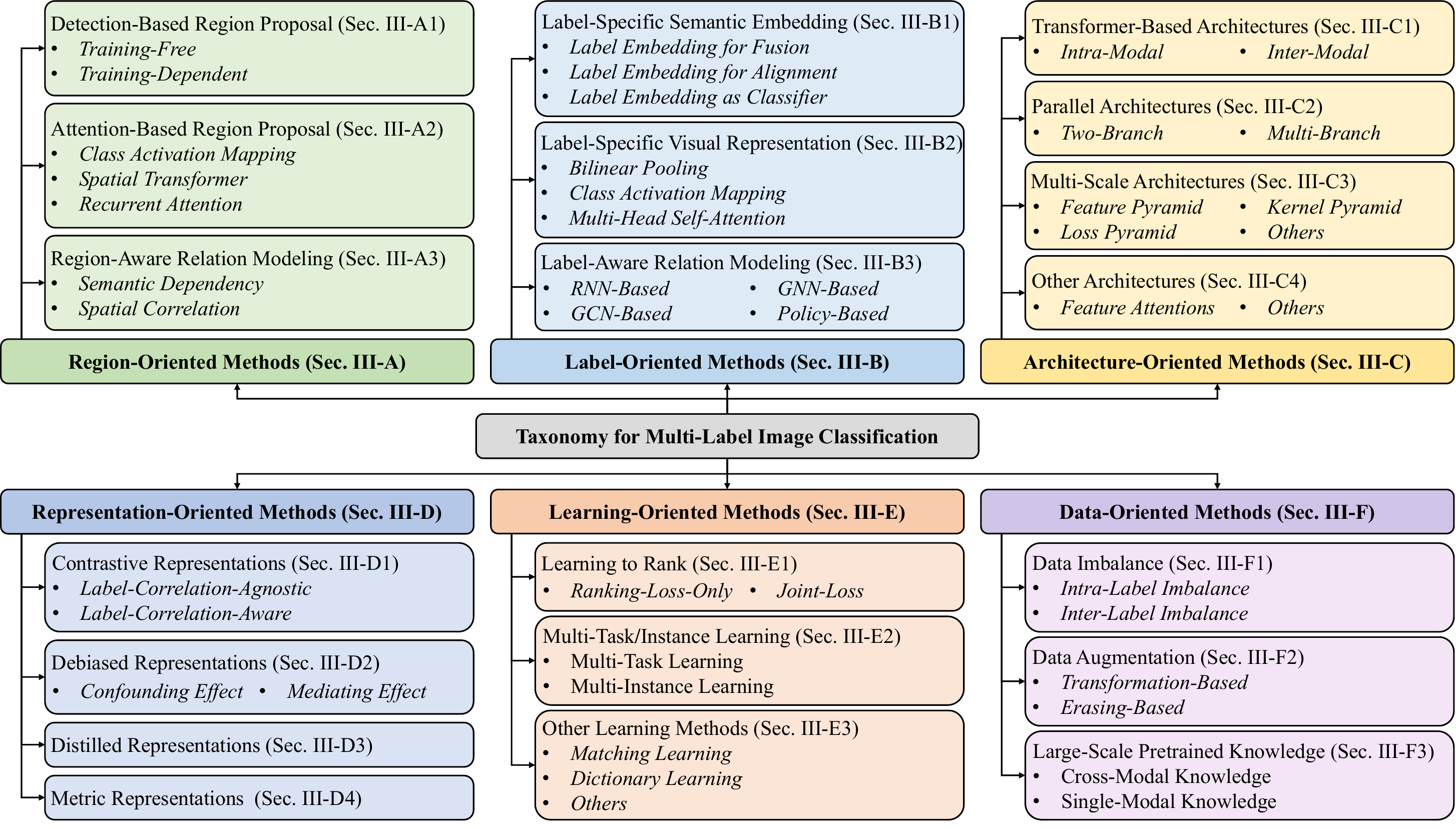}
    \caption{Overview of our taxonomy for MLIC methods. We categorize these methods into six groups according to their key innovations and core contributions.}
    \label{fig:texonomy}
\end{figure*}

\section{Methods: A Survey}
\label{sec:method}

In this section, we comprehensively survey deep learning-based MLIC methods. As illustrated in Fig. \ref{fig:texonomy}, we organize existing approaches into six aspects: region-oriented methods in Subsec. \ref{sec:region-methods}, label-oriented methods in Subsec. \ref{sec:label-methods}, architecture-oriented methods in Subsec. \ref{sec:arch-methods}, representation-oriented methods in Subsec. \ref{sec:representation-methods}, learning-oriented methods in Subsec. \ref{sec:learning-methods}, and data-oriented methods in Subsec. \ref{sec:data-methods}.

\subsection{Region-Oriented Methods\label{sec:region-methods}}

Region-oriented methods transform MLIC into a series of region-specific multi-class classification problems by identifying informative regions within images.

\subsubsection{Detection-Based Region Proposal}

Object detection is commonly used to mine potential informative regions from an image, which can be either training-free or training-based.

\textbf{Training-Free.}
Most training-free methods follow a three-stage pipeline, consisting of region proposal, feature extraction, and label prediction. For region proposal, early studies \cite{wei2015hcp,Zhao2016RegionalGN,yang2016exploit,yu2017combining,liu2018multi} rely on off-the-shelf techniques to detect object-related image regions. Specifically, BING \cite{cheng2014bing} and Edge Boxes \cite{zitnick2014edge} are adopted in \cite{wei2015hcp,yu2017combining,Zhao2016RegionalGN,liu2018multi} for hypotheses generation due to their high computation efficiency and recall in object detection. While  Yang et al. \cite{yang2016exploit} employ Selective Search \cite{uijlings2013selective} in a unsupervised way, Xu et al. \cite{xu2020joint} apply a pre-trained Mask R-CNN \cite{he2017mask} to recognize and localize objects in images. To extract regional features, common solutions \cite{wei2015hcp,yang2016exploit,yu2017combining} involve cropping the detected regions from images, resizing them, and processing them with a shared CNN, which, however, might be computationally intensive. Drawing inspiration from Faster R-CNN \cite{ren2016faster}, several methods \cite{Zhao2016RegionalGN,liu2018multi} implement a region of interest (RoI) pooling operation on the convolutional feature maps, followed by a fully-connected layer to produce regional representations for each proposal. For label prediction, many studies \cite{wei2015hcp,Zhao2016RegionalGN,yu2017combining} leverage cross-region max-pooling to aggregate the predictions over all regions while minimizing potential prediction noise. Yang et al. \cite{yang2016exploit} build a pool of ground-truth object bounding boxes and utilize the labels of neighboring regions in the pool to enhance each region proposal, culminating in the creation of a Fisher vector. In \cite{liu2018multi}, the predicted region-wise label logits are summed to obtain final prediction.

\textbf{Training-Based.} Instead of relying on off-the-shelf region proposal tools, several approaches \cite{zhang2018multilabel,zhang2022spatial,zhang2023spatial} integrate an object detection module into the network and train the entire framework end-to-end using bounding box annotations and ground-truth labels. RLSD \cite{zhang2018multilabel} develops a fully convolutional localization layer to locate regions, and then extracts regional features via bilinear interpolation and fuses region-wise predictions by max-pooling for final output. Subsequent studies \cite{zhang2022spatial,zhang2023spatial} introduce an object detection branch following Faster R-CNN \cite{ren2016faster} and extract regional features with RoI pooling to estimate object presence confidence, while Shen et al. \cite{shen2023improved} adopt YOLOv5s to enhance both flexibility and effectiveness.

\subsubsection{Attention-Based Region Proposal}

Attention models are also an effective tool for approximately locating informative regions within input images for multi-label prediction.


\textbf{Class Activation Mapping.} Class activation mapping \cite{zhou2016learning} (CAM) highlights image regions that contribute most to a model's predictions. Building on this, several studies \cite{gao2021learning,zhan2022global} learn a 1$\times$1 convolution layer to generate label-specific activation maps. After min-max normalization, these activation maps are binarized with a constant threshold to delineate discriminative object regions. The region-wise predictions produced by a shared CNN are subsequently aggregated via category-wise max-pooling to obtain the final label predictions. Furthermore, Zhan et al. \cite{zhan2022global} introduce a weakly supervised loss and an energy-based RoI selection mechanism to enhance the quality of the CAM-based region proposals for MLIC.

\textbf{Spatial Transformer.} Spatial transformer \cite{jaderberg2015spatial} is a neural network module that enables explicit spatial manipulation on input feature maps. It offers the capability to perform geometric transformations, such as translation, scaling, rotation, and more complex warpings, by learning an affine transformation matrix during the training process. Building on this concept, several works \cite{wang2017multi,ablavatski2017enriched,nie2022multi,dai2023global} incorporate a spatial transformer layer into multi-label frameworks to sequentially localize object regions and perform label prediction. This design allows object localization and MLIC to be jointly optimized in a unified end-to-end architecture. The resulting regional predictions are finally aggregated via label-wise max-pooling to produce the final multi-label outputs.

\textbf{Recurrent Attention.} Recurrent attention is motivated by human strategies for visual sequence recognition tasks. Several methods \cite{chen2018order,lyu2019attend,yazici2020orderless} employ long short-term memory (LSTM) \cite{hochreiter1997long} networks to iteratively discover relevant regions by modeling the correlations between hidden states and image features. DRAM \cite{ba2014multiple} processes multi-resolution crops at specific locations and updates its internal state at each time step, simultaneously producing label predictions and the next glimpse location. Following the anchor strategy in Faster R-CNN \cite{ren2016faster}, Chen et al. \cite{chen2018recurrent} extend DRAM by yielding multiple regions with various scales and aspect ratios centered at each location, thus better covering all potential objects for MLIC.



\subsubsection{Region-Aware Relation Modeling\label{sec:region-relation}} 

Correlations between object regions are critical in MLIC, primarily in the form of semantic dependency and spatial correlation.

\textbf{Semantic Dependency.} Region-aware semantic dependency in multi-label images mainly arises from contextual cues and semantic interactions between object regions. For example, a ``board" in a snowy scene is likely a ``snowboard", whereas in a street scene it is more likely a ``skateboard". To exploit such dependency, typical studies \cite{zhang2018multilabel,chu2021multi} generate multiple object regions and then process them sequentially using LSTMs to capture latent semantic relations. Many approaches, including reinforcement learning-based \cite{ba2014multiple,chen2018recurrent}, spatial transformer-based \cite{wang2017multi,ablavatski2017enriched,nie2022multi} and visual attention-based \cite{chen2018order,lyu2018coarse,li2018multi,li2018attentive,lyu2019multi,lyu2019attend,yazici2020orderless}, iteratively discover a sequence of informative regions associated with different semantic objects and model their semantic dependencies through the recurrent dynamics of LSTMs, allowing the learned semantic dependencies guide the localization of subsequent regions.

\textbf{Spatial Correlation.} Region-aware spatial correlation primarily manifests in the spatial layout among object regions. For example, a device with a screen next to a ``keyboard" is likely to be a ``computer", whereas a similar device next to a remote control is more likely a ``TV".
To explore such correlations, several methods \cite{zhao2021transformer,zhang2023spatial,zhou2023datran} use self-attention \cite{vaswani2017attention} to model interactions between objects at different positions. Specifically, Zhao et al. \cite{zhao2021transformer} apply relative positional encoding on channel-wise features and then use a Transformer encoder \cite{vaswani2017attention} to capture position-wise spatial relationships. Zhang et al. \cite{zhang2023spatial} define positional embeddings as the intersection over union (IoU) among object regions and fuse them with object semantics through self-attention, incorporating both appearance and geometry for MLIC. Similarly, Subsequent methods \cite{zhou2023datran,zhou2023feature} adopt multi-head self-attention (MSA) to capture spatial correlations in multi-label images. In \cite{cheng2022mltr}, a dedicated multi-label Transformer is construct, where cross-window attention is tailored to effectively exploit spatial dependencies within images in a self-attention manner.

\subsection{Label-Oriented Methods\label{sec:label-methods}}

Label-oriented methods emphasize learning semantic embeddings or visual representations for each label, converting MLIC into a set of label-specific binary classification tasks.

\subsubsection{Label-Specific Semantic Embedding\label{sec:label-embedding-classifier}} 

Label-specific semantic embedding aims to learn continuous semantic representations for labels based on their one-hot encodings.

\textbf{Label Embedding for Fusion.} 
A common goal of label embedding is to fuse with image features, yielding enhanced joint representations for MLIC. In \cite{wang2016cnn}, one-hot label vectors are mapped into a continuous space using a learnable matrix and processed by an LSTM unit, whose hidden states are then concatenated with image features for label prediction. In \cite{liu2017semantic,yazici2020orderless}, image features are concatenated with the embedding of the previously predicted label and then fed into LSTMs for current prediction. Subsequent methods \cite{wang2020fast,xie2022label,wang2022cross} fuse image features and label embeddings by multi-modal factorized bilinear (MFB) pooling, which maps them into a unified space and combines them using a Hadamard product followed by sum pooling. In \cite{wang2020multi}, lateral connections are built between a graph neural network (GCN) and a CNN at multiple layers, injecting label embeddings into the process of image representation learning for label prediction.


\textbf{Label Embedding for Alignment.} Label semantic embeddings are also widely leveraged to align visual features across modalities. Typical methods \cite{chen2019learning,zhu2023scene,zhu2024semantic} employ pretrained language models, e.g., GloVe \cite{pennington2014glove} and Bert \cite{devlin2018bert}, to represent labels as word vectors, leveraging their textual semantics to guide bilinear pooling toward semantically relevant image regions. Instead, Transformer-based approaches \cite{liu2021query2label,dao2021multi,zhu2022two,dao2023contrastively,huang2023cross,chen2024pursuit,wang2024dynamic} treat label embeddings as learnable parameters that steer MSA to aggregate visual features in a label-aware manner. To acquire correlation-aware label embeddings for cross-modal alignment, You et al. \cite{you2020cross} propose to minimize the mean square error (MSE) between the cosine similarity of label embeddings and their statistical co-occurrence probabilities, while Qu et al. \cite{qu2023multi} derive refined label embeddings using a graph attention network (GAT) with differentiable graph pooling for MLIC.

\textbf{Label Embedding as Classifier.} Label semantic embeddings are also often learned directly as classifiers, where each embedding serves as a weight vector whose inner product with the image features yields the corresponding label score. In \cite{wang2016cnn}, the learned label embeddings are not only fused with image features, but also additionally utilized as classification weights that compute inner-product scores with the fused representations. In GCN-based approaches \cite{chen2019multi,meng2019multi,qing2019learning,wang2020multi,li2020multi,vu2020privacy,nguyen2021modular,singh2022multi,singh2022iml,wang2022stmg,sun2023attention,yuan2023graph,singh2024multi,wu2024multi}, label embeddings are commonly initialized using pretrained language models and then propagated through a GCN, yielding a set of semantic correlation-aware interdependent object classifiers that are matched with image features for MLIC.

\subsubsection{Label-Specific Visual Representation\label{sec:label-representation-learning}}

Label-specific visual representation aims to aggregate visual features for each label from the semantically related image regions.

\textbf{Bilinear Pooling.} Distinct from cosine similarity \cite{you2020cross,liang2022multi,kuang2025two,ye2025multi} that operates in an inner product space, bilinear pooling is commonly used for cross-modal fusion in MLIC via a Hadamard product. Typical methods \cite{chen2022label,qu2023multi} feed label embeddings and image features into a bilinear pooling module followed by softmax, and use the normalized logits to re-weight and aggregate image features, yielding a visual representation for each label. To reduce computational cost, many studies \cite{chen2019learning,chen2020knowledge,wu2020adahgnn,zhou2021multi,chen2022label,zhu2023scene,pu2023semantic,zhu2024semantic,gu2024hypergraph,fu2024generative} adopt low-rank bilinear pooling \cite{kim2016hadamard}, which first projects label embeddings and image features into a low-dimensional space and then applies bilinear pooling to obtain spatial attention maps. Specifically, Pu et al. \cite{pu2023semantic} apply low-rank bilinear pooling on feature maps to re-weight them along both spatial and channel dimensions, followed by a global max pooling to obtain label-specific visual representations.

\textbf{Class Activation Mapping.} CAM \cite{zhou2016learning} produces an activation map on the input image for each label, underscoring image regions that contribute significantly to label recognition.
Typical methods \cite{ye2020attention,cao2021multi,chen2024msfa} use CAM in a parameter-free manner to reveal implicit attentions for each label. Instead, most studies \cite{zhu2017learning,meng2019multi,kuang2023multi,kuang2024multi,zhao2021transformer,chen2022sst,wu2022smart,zhou2023mining,huang2023cross,li2023improved,chen2025modeling} learn label-specific activation maps with a convolutional module composed of several convolution layers with ReLU, followed by a final 1$\times$1 convolution layer whose number of kernels matches the number of labels. After applying softmax normalization along the spatial dimension, label-specific activation maps are obtained and serve as attentional weights to aggregate image features, thereby producing label-specific visual representation for MLIC.

\textbf{Multi-Head Self-Attention.} As a core component in Transformer \cite{vaswani2017attention,dosovitskiy2020image}, MSA has been widely exploited to produce visual representations for each label. Typical methods \cite{liu2021query2label,dao2021multi,zhu2022two,dao2023contrastively,huang2023cross,chen2024pursuit,wang2024dynamic} leverage Transformer decoders for this purpose. In MSA, each cross-attention layer maps label embeddings to queries and image features to keys and values. MSA then independently compute similarity between queries and keys to obtain attention weights over the values, and their outputs are then concatenated and fused to update the label embeddings. As the layers stack, label embeddings progressively aggregate semantically correlated features from images, yielding label-specific visual representations.

\subsubsection{Label-Aware Relation Modeling\label{sec:label-relation}}

Many methods model label relations via diverse networks and learning strategies.

\textbf{RNN-Based.} Early methods model label dependencies by storing label contexts in LSTM memory, casting MLIC as sequence generation under a fixed label order, such as frequent-first \cite{wang2016cnn}, rare-first \cite{jin2016annotation,liu2017semantic}, and dictionary order \cite{jin2016annotation}. 
In \cite{wang2016cnn}, the previously predicted label is used to inform the current prediction at each step for capturing label dependencies. Furthermore,
Jin et al. \cite{jin2016annotation} introduce \textit{start} and \textit{stop} signals to generate label sequences of proper length.
While Liu et al. \cite{liu2017semantic} develop a regularized embedding layer between a CNN and a RNN to ease joint training,
Lyu et al. \cite{lyu2019attend} retain a dynamic visual context by soft attention, focusing on relevant image regions for each label.
To reduce MLIC’s sensitivity to label order,
Chen et al. \cite{chen2019multi3} feed the channel-wise object features into an LSTM for sequential prediction without specifying a label order.
In \cite{dutta2020recurrent}, previously predicted ground-truth labels are treated as negatives in later steps, enabling to learn diverse label dependencies without any fixed order.
While Chen et al. \cite{chen2018order} predict all labels at each step and maintain a candidate pool to prevent duplicates,
Yazici et al. \cite{yazici2020orderless} dynamically select the label order that minimizes the training loss by rearranging ground-truth labels to best match the predictions.

\textbf{GNN-Based.} GNN-based methods model label correlations via graph propagation, typically building a label graph from label co-occurrence. Early methods \cite{chen2019learning,chen2020knowledge} initialize nodes with label-specific visual representations and update them by a gated recurrent mechanism, enhancing high co-occurrence interactions and suppressing others.
While Chen et al. \cite{chen2022label} update label semantic embeddings to pull together highly co-occurring labels, Zhu et al. \cite{zhu2023scene} introduce an entropy-based loss for unsupervised scene detection, enabling scene-aware label interactions.
Subsequent studies \cite{zhou2023double,huang2025multi,ye2025multi} use GATs to automatically learn label correlations from neighbors’ features without predefined graphs. Li et al. \cite{li2026multi} construct a dynamic-static label correlations fusion module with a GAT to optimize label features and establish better label–object mappings.
Additionally, several methods \cite{wu2021gmmlic,wu2023semantic} jointly model spatial relations, label semantics, and instance–label assignments within a unified GNN framework for MLIC.

\begin{table*}
    \centering
    \scriptsize
    \caption{Summarization and comparison of GCN-based methods in modeling label correlation for MLIC.}
    \renewcommand{\tabcolsep}{0pt}
    \begin{tabularx}{\textwidth}{Xp{3.5cm}p{5.5cm}p{2.5cm}p{2.2cm}p{1.6cm}}
        \toprule
            \belowrulesepcolor{gray!30!}
            \rowcolor{gray!30!} Method & Node Representation & Correlation Matrix & Denoising & Over-smoothing & Normalization \\
            \aboverulesepcolor{gray!30!}
        \midrule
            \belowrulesepcolor{gray!15!}
            \rowcolor{gray!15!} \multicolumn{6}{c}{\textbf{Hand-crafted Correlation Matrix}} \\
            \aboverulesepcolor{gray!15!}
        \midrule
            ML-GCN \cite{chen2019multi} & GloVe embeddings & Co-occurrence probability & Binarization 
            & Re-weighting & Laplacian
        \\ \rowcolor{gray!10!} 
            Meng et al. \cite{meng2019multi} & GloVe embeddings & Co-occurrence probability & Binarization 
            & Re-weighting & Laplacian
        \\
            \multirow{2}{*}{TSGCN \cite{xu2020joint}} & GloVe embeddings & Co-occurrence probability & Binarization & Re-weighting & -
            \\ 
            & Region features & IoU + relative direction & - & - & - 
        \\ \rowcolor{gray!10!} 
            F-GCN \cite{wang2020fast} & GloVe embeddings & Co-occurrence probability & Binarization 
            & Re-weighting & Laplacian
        \\ 
            \multirow{2}{*}{KSSNet \cite{wang2020multi}} & GloVe embeddings & Co-occurrence probability &      \multirow{2}{*}{Filtering} & \multirow{2}{*}{Identity matrix} & \multirow{2}{*}{Laplacian}
            \\
            & ConceptNet & Semantic relation & & & 
        \\ \rowcolor{gray!10!} 
            P-GCN \cite{chen2021learning} & label-specific visual features & Co-occurrence probability & Binarization & Re-weighting & Laplacian
        \\ 
            G-CAM \cite{wang2021g} & GloVe embeddings & Co-occurrence probability & Binarization & Re-weighting & Laplacian 
        \\ \rowcolor{gray!10!} 
            &  & Co-occurrence probability & & &
            \\ \rowcolor{gray!10!} 
            &  & Wordnet structure & & & 
            \\ \rowcolor{gray!10!} 
            \multirow{-3}{*}{VSGCN \cite{deng2022beyond}} & \multirow{-3}{*}{\makecell[l]{GloVe embeddings \\ Visual prototypes}} & Cosine similarity & \multirow{-3}{*}{Filtering} & \multirow{-3}{*}{-} & \multirow{-3}{*}{$\ell_1$-norm}
        \\ 
            IML-GCN \cite{singh2022iml} & Visual prototypes & Co-occurrence probability & Binarization & Re-weighting & Laplacian
        \\ \rowcolor{gray!10!} 
            CFMIC \cite{wang2022cross} & GloVe embeddings & Co-occurrence probability & Binarization 
            & Re-weighting & Laplacian 
        \\ 
            FLNet \cite{sun2023attention} & GloVe embeddings & Co-occurrence probability & - & - & Laplacian 
        \\ \rowcolor{gray!10!}
            Wu et al. \cite{wu2024multi} & GloVe embeddings & Co-occurrence probability & Binarization & - & -
        \\
        \midrule
            \belowrulesepcolor{gray!15!}
            \rowcolor{gray!15!} \multicolumn{6}{c}{\textbf{Adaptive Correlation Matrix}} \\
            \aboverulesepcolor{gray!15!}
        \midrule
            A-GCN \cite{qing2019learning,li2020learning} & Glove embeddings & Dot production & - & $\ell_1$-norm loss & Laplacian 
        \\ \rowcolor{gray!10!} 
            MCSAF \cite{li2020multi} & Glove embeddings & Cosine similarity & - & - & - 
        \\
            STMG \cite{wang2022stmg} & Bert embeddings & Dot production & - & $\ell_2$-norm loss & Laplacian 
        \\ \rowcolor{gray!10!} 
            LGLM \cite{xie2022label} & Glove embeddings & Dot production & - & $\ell_2$-norm loss & Laplacian 
        \\
            SRDL \cite{pu2023semantic} & Glove embeddings & Random initialization & - & - & Laplacian 
        \\ \rowcolor{gray!10!} 
            SGTN \cite{vu2020privacy} & dpUGC embeddings & Cosine similarity & Binarization & Identity matrix & Laplacian 
        \\
            MGTN \cite{nguyen2021modular} & Glove embeddings & Co-occurrence probability & Interval Binarization & Identity matrix & Laplacian 
        \\ \rowcolor{gray!10!} 
            GATN \cite{yuan2023graph} & Glove embeddings & Cosine similarity & Binarization & Re-weighting & Laplacian 
        \\
        \midrule
            \belowrulesepcolor{gray!15!}
            \rowcolor{gray!15!} \multicolumn{6}{c}{\textbf{Dynamic Correlation Matrix}} \\
            \aboverulesepcolor{gray!15!}
        \midrule
            \multirow{2}{*}{ADD-GCN \cite{ye2020attention}} & \multirow{2}{*}{Label-specific visual features} & Random initialization & Binarization & Re-weighting & Laplacian
            \\
            & & Estimated from input features & - & - & - 
        \\ \rowcolor{gray!10!} 
            & & Random initialization & - & - & - 
            \\ \rowcolor{gray!10!} 
            \multirow{-2}{*}{2S-DGCN \cite{cao2021multi}} & \multirow{-2}{*}{Label-specific visual features} & Estimated from input features & - & - & - 
        \\
            \multirow{2}{*}{Object-GCN \cite{liu2024multi}} & GloVe embeddings & Random initialization & - & - & - 
            \\
            & Label-specific visual features & Estimated from input features & - & - & - 
        \\ \rowcolor{gray!10!} 
            & GloVe embeddings & Statistical label Co-occurrence probability & - & - & - 
            \\ \rowcolor{gray!10!} 
            \multirow{-2}{*}{IA\_GCN \cite{wang2021instance}} & Label-specific visual features & Inner products between label-wise region features & - & - & - 
        \\ 
            M-GCN \cite{yao2022m} & Label-specific visual features & Estimated from input features & - & - & - 
        \\ \rowcolor{gray!10!}
            &  & Co-occurrence probability & - & - & - 
            \\ \rowcolor{gray!10!}
            &  & Estimated from paired features & - & - & Softmax 
            \\ \rowcolor{gray!10!}
            \multirow{-3}{*}{ML-AGCN \cite{singh2022multi,singh2024multi}} & \multirow{-3}{*}{Label-specific visual features} & Cosine similarity & - & - & - 
        \\ 
            & Label-specific visual features & Dot product between input features & - & - & Softmax 
            \\ 
            \multirow{-2}{*}{DRGN \cite{zhou2023mining}} & Global image feature & Learnable Laplacian matrix & - & - & - 
        \\ \rowcolor{gray!10!}
            MGSGN \cite{kuang2023multi} & Label-specific visual features & Cosine similarity between projected input features & Filtering & - & - 
        \\ 
            MMDSR \cite{kuang2024multi} & Label-specific visual features & Cosine similarity between projected input features & Filtering & - & - 
        \\ \rowcolor{gray!10!}
            PCMT \cite{li2024pyramidal} & Label-specific visual features & Outer production of predicted score vector & - & - & Softmax 
        \\
        \bottomrule
    \end{tabularx}
    \label{tab:gcn}
\end{table*}

\begin{figure}
    \centering
    \includegraphics[width=1\linewidth]{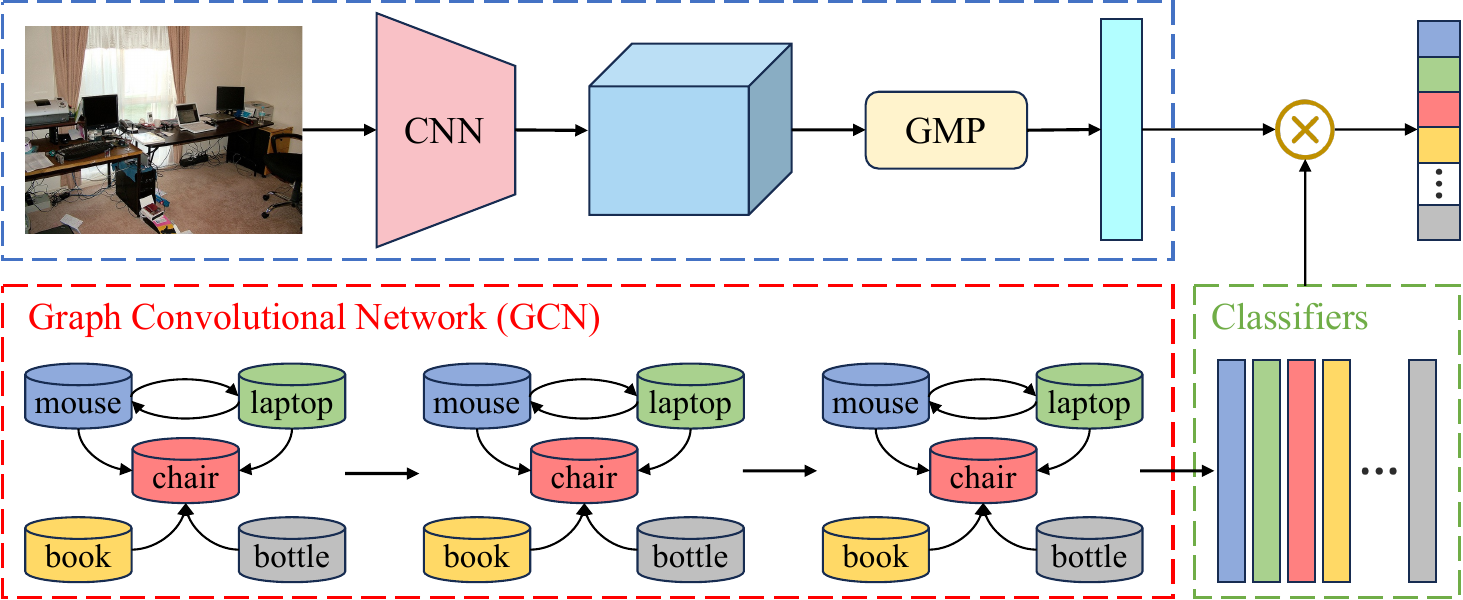}
    \caption{The pipeline of the ML-GCN. Figure is reproduced from \cite{chen2019multi}. GMP indicates the global max-pooling operation. Best viewed in color.}
    \label{fig:mlgcn}
\end{figure}

\textbf{GCN-Based.} GCN-based MLIC methods capture label correlations using graph convolution. ML-GCN \cite{chen2019multi} first introduces GCNs into MLIC by estimating a label co-occurrence matrix, binarizing it into an adjacency matrix with the entity $a'_{ij}=0$ if $p_{ij}<\tau$ and $1$ otherwise, and re-weighting it as
\begin{equation}
    a_{ij} = \left\{
        \begin{array}{lr}
            p/\sum_{j=0,j\neq i}^{N-1}a'_{ij}, & \mathrm{if}\ i\neq j, \\
            1 - p, & \mathrm{if}\ i=j,
        \end{array}
    \right.
\end{equation}
where $\tau\in[0,1]$ filters noisy edges and $p$ balances a label against its neighbors. A GCN layer is then defined as $\mathbf{H}^{l+1} = \sigma(\mathbf{A}\mathbf{H}^l\mathbf{W}^l)$,
where $\mathbf{W}^l$ is a parameter matrix; $\sigma(\cdot)$ is a non-linear activation; $\mathbf{H}^0$ is label embeddings initialized by GloVe \cite{pennington2014glove}. Stacking such layers enables ML-GCN learn label correlations and inter-dependent label classifiers for matching image features, as illustrated in Fig. \ref{fig:mlgcn}.

\raisebox{0.3ex}{\scriptsize $\bullet$} \textit{Hand-Crafted Correlation Matrix}. Following ML-GCN, many subsequent studies manually construct correlation matrices for GCN-based relation modeling, primarily aiming to enhance visual features and improve their alignment with label classifiers, such as attention-based feature re-weighting \cite{meng2019multi,sun2023attention}, label-specific feature pooling \cite{xu2020joint}, multi-scale cross-modal matching \cite{liang2022multi,wang2025establishing}, MFB-pooling-based cross-modal matching \cite{wang2020fast,wang2022cross}, label-enhanced visual representation \cite{wang2020multi,sun2023attention}, label-specific representation learning \cite{chen2021learning}, GCN-based visual attention consistency \cite{wang2021g}, CAM-based image segment generation \cite{wu2024multi}, data-drive label-wise prototype learning \cite{singh2022iml}, pairnorm-based normalization \cite{chauhan2023tackling}, self-attention-based dynamic fusion \cite{li2023improved}, and direction-guided multi-label ranking \cite{zhou2021multiple}. 
To reduce label-imbalance bias, multi-view correlation matrices are handcrafted to enrich label relationships beyond simple co-occurrence. In \cite{xu2020joint}, an object spatial graph is built with nodes being object regions connected by IoU and relative direction. A ConceptNet-based knowledge graph is combined with co-occurrence in \cite{wang2020multi}. In \cite{deng2022beyond}, two extra matrices is derived from WordNet distances among labels and cosine similarities of their embeddings, based on which a multi-head GCN is developed to jointly model label correlations. Please refer to the upper part of Table~\ref{tab:gcn} for more details.

\raisebox{0.3ex}{\scriptsize $\bullet$} \textit{Adaptive Correlation Matrix}. To solve the limited flexibility of hand-crafted correlation matrices, many methods \cite{qing2019learning,li2020learning,wang2022stmg,xie2022label} learn an adaptive label correlation matrix by applying $1\times1$ convolutions to label embeddings followed by dot products. Some works \cite{qing2019learning,li2020learning} imposes an $\ell_1$ sparsity constraint to reduce over-smoothing, while others \cite{wang2022stmg,xie2022label} use an $\ell_2$ penalty. Beyond this,
adaptive correlations can also be constructed by computing cosine similarity between label embeddings \cite{li2020multi}, by treating the correlation matrix as a learnable module \cite{pu2023semantic}, or by initializing it from external knowledge \cite{lin2025adaptive}.
However, these methods rely on hard binarization, which drops all values below a threshold and may render label representations indistinguishable. To alleviate this issue,
typical methods \cite{vu2020privacy,nguyen2021modular} apply multiple real-value thresholds to individually binarize the co-occurrence matrix into several binary matrices, process them with convolutions, and fuse the outputs via element-wise multiplication.
In \cite{yuan2023graph}, MSA is applied to a cosine-similarity correlation matrix to explore diverse subgraphs, which are multiplied to form an adaptive correlation matrix for label correlation modeling.
Further details are provided in the middle part of Table~\ref{tab:gcn}.

\raisebox{0.3ex}{\scriptsize $\bullet$} \textit{Dynamic Correlation Matrix}. Both adaptive and hand-crafted correlation matrices encode only static label relations, which limits their ability to generalize across diverse images. To address this, many work instead learns dynamic, image-conditioned correlation matrices. In two-stage architectures \cite{ye2020attention,cao2021multi,liu2024multi}, a first GCN with a static matrix models global label co-occurrence, while a second, dynamic GCN is driven by an image-dependent correlation matrix to capture fine-grained, instance-specific relations.
In one-stage designs, several works \cite{wang2021instance,yao2022m} augment the co-occurrence matrix by an instance-aware correlation matrix and a memory GCN respectively, both derived from dot products of label-specific visual features. Similarly, object-level and image-level GCNs are built in \cite{zhou2023mining} with correlations from inner products of object features. In \cite{singh2022multi,singh2024multi}, a dynamic matrix is created by summing co-occurrence, attention-based, and similarity-based correlation matrices. Related works  \cite{kuang2023multi,kuang2024multi} project label-specific features into multiple subspaces and compute average cosine-similarity matrices across these spaces as the adjacency for dynamic GCNs. Recently, \cite{li2024pyramidal} uses the outer product of predicted label scores as the correlation matrix, enabling a dynamic GCN to capture complex label interactions. \cite{zhao2025towards} introduces average message passing to avoid over-emphasizing co-occurrence and to purify label representations for MLIC. More information can be found in the bottom part of Table~\ref{tab:gcn}.

\raisebox{0.3ex}{\scriptsize $\bullet$} \textit{Correlation-Matrix-Free}. Many attempts have been made to explore correlation-matrix-free GCNs to model label correlations. Wu et al. \cite{wu2020adahgnn} create an adaptive hypergraph that uses hyperedge convolution to capture high-order label correlations, with the incidence matrix initialized from GloVe embeddings and optimized end-to-end. In \cite{gu2024hypergraph}, vertices and hyperedges are refined to highlight co-occurring labels and uses Transformer-style message passing for richer node features. Recently, a fully graph-convolutional model \cite{yao2025gkgnet} is developed where each node adaptively connect to a variable number of neighbors, effectively handling multi-scale, irregular regions and label correlations via GCNs based on group KNN (K-nearest neighbor). Kuang et al. \cite{kuang2026multi} enhance ViT-based MLIC by introducing a Vision GNN branch that creates a KNN graph over image patches, aggregates semantically related regions, and fuses these graph features for MLIC.

\textbf{Policy-Based.} 
Many policies have also been proposed to model relationships between labels.
In \cite{he2018reinforced}, MLIC is formulated as a Markov Decision Process, where label correlations are captured by a state that combines image feature with previously predicted labels to guide subsequent predictions.
In \cite{dutta2020recurrent}, the ground-truth labels that have been predicted are treated as negatives in later loss computation, compelling the model to learn multiple inter-label dependencies.
While label-correlation aware loss \cite{chu2021multi} encourages the frequently co-occurring labels to have compatible predictions and penalizes outputs that violate known label correlations,
label mask training strategy \cite{lanchantin2021general} enforces correlation learning by randomly masking a subset of labels and requiring the model to predict them from the remaining ground-truth labels. Recently, 
Chen et al. \cite{chen2025modeling} use partial labels and masked modeling policy to infer full labels, thereby capturing label correlations for MLIC.

\subsection{Architecture-Oriented Methods\label{sec:arch-methods}}

Architecture-oriented methods focus on developing effective and specialized network architectures for MLIC.

\subsubsection{Transformer-Based Architectures}

Transformers \cite{vaswani2017attention,dosovitskiy2020image} have achieved remarkable success in both NLP and CV, and their MSA architectures have been extensively explored for MLIC in both intra-modal and inter-modal settings.

\textbf{Intra-Modal.} Transformer-based intra-modal architectures focus on modeling spatial correlations among image patches. Early methods \cite{zhao2021transformer,wu2022smart,chen2022sst} directly employ Transformer encoder for this purpose. Specifically, Wu et al. \cite{wu2022smart} exploit masked attention to restrict MSA to high-confidence labels and avoid deviation from low-confidence labels, whereas Chen et al. \cite{chen2022sst} employ two independent Transformer encoders to capture both spatial and semantic label correlations. The multi-label Transformer framework \cite{cheng2022mltr} captures both local and global dependencies in images by window partitioning, in-window pixel attention, and cross-window attention.
In \cite{dong2022towards}, label-specific tokens are assigned to generate class-level attention maps, improving both interpretability and performance.
Recent studies \cite{zhou2023feature,zhou2023datran,liu2025transformer} explore Transformer architectures to fuse multi-scale convolutional feature maps and capture their spatial dependencies for MLIC.

\textbf{Inter-Modal.} Transformer-based inter-modal architectures model the cross-modal correlations between images and labels.
Early methods \cite{lanchantin2021general,zhao2021m3tr} concatenate image features with label embeddings and feed them into a Transformer encoder to capture their correlations. Subsequent works \cite{liu2021query2label,zhu2022two,zhou2023aligning}, use a Transformer decoder to pool semantically related image features and update label embeddings, yielding label-specific visual representations.
Additionally,
Graph Transformers are introduced to dynamically refine the label correlation matrix for adaptive inter-label relation modeling in \cite{yuan2023graph}, and to capture intra-modal correlations within object regions and labels as well as latent instance–label correspondences in \cite{wu2023transformer}.
To better mine cross-modal correlations, interactive visual-linguistic attention \cite{ouyang2023hsvlt} is framed to integrate cross-modal interactions for joint representations. In \cite{li2024pyramidal}, a pyramidal visual guidance layer and a hybrid modal interaction layer are created to capture hierarchical visual-label dependencies and alleviate semantic disparities across modalities.
In \cite{wang2026topic}, multiple prompt tokens that encode topic relations among label combinations are hierarchically inserted into Transformer blocks to inject latent topic information for MLIC.


\subsubsection{Parallel Architectures}

Parallel MLIC architectures are typically either two-branch or multi-branch designs that process different feature modalities or semantic levels in parallel.

\textbf{Two-Branch.} In two-branch architectures, each branch typically serves a distinct role (e.g., one captures object spatial information while the other models image context \cite{zhang2022spatial}; or one focuses on intra-image dependencies while the other exploits inter-image semantic relations \cite{zhou2023mining}), with global–local and two-stream designs being particularly prevalent.

\raisebox{0.3ex}{\scriptsize $\bullet$} \textit{Global-Local.} Global–local architectures comprise two branches: one for global features and one for local regions.
Many studies \cite{yu2017combining,zhu2017learning,yu2019delta,gao2021learning} use category-wise max-pooling to gather regional predictions in local branch, which are then fused with global-branch predictions for the final result.
In \cite{yu2019delta}, multi-scale features are produced in local branch and spatially pooled to cooperate with global branch for prediction.
Gao et al. \cite{gao2021learning} reduce the number of attentional regions while maintaining high diversity to bridge the gap between global and local branches.
While Chu et al. \cite{chu2021multi} simply combine the global and regional predictions via element-wise multiplication, zhan et al. \cite{zhan2022global} apply self-attention between global and local features to force message passing between branches and capture label correlations.
Recent methods \cite{kuang2023multi,kuang2024multi} exploit correlations between global information and local multi-scale features to enhance performance.

\raisebox{0.3ex}{\scriptsize $\bullet$} \textit{Two-Stream.} Two-stream architectures use two branches tailored to different modalities, typically learning complementary visual and semantic representations for MLIC. Early approaches explore two-stream GCNs \cite{xu2020joint,cao2021multi}, with one stream capturing semantic correlations within label space and the other modeling spatial relationships among object regions.
Zhu et al. \cite{zhu2022two} present a two-stream Transformer, which includes a spatial stream that explores correlations among image patches and a semantic stream that aggregates spatial features to update label semantics.
In \cite{hu2024dual}, one branch extracts features from image segments and the other enhances label semantics with CLIP \cite{radford2021learning}.
In addition, dual-stream architectures \cite{li2018improving,luo2019visual} add an independent branch to capture saliency or scene information as a complementary modality to the visual branch.
More recently, a two-stream semantic alignment network \cite{kuang2025two} is constructed, where one stream extracts image patch features and the other aligns them with label semantics, updating both bidirectionally via MSA.

\textbf{Multi-Branch.} 
Early multi-branch methods \cite{chen2013multi,wei2015hcp} extract multiple object-related image regions and feed them into separate branches that are jointly used for final label prediction.
Zhou et al. \cite{zhou2023attention} present a three-branch network: a categorical memory branch that augments features using cross-image context, a channel-relation branch that models inter-channel correlations, and a spatial-relation branch that captures pixel dependencies to highlight salient objects.
In \cite{coulibaly2022deep}, MLIC is treated as parallel subtasks processed by multiple branches, either assigning distinct classifier branches to different label groups or fusing features from several pre-trained CNN branches into a single prediction head.
Yin et al. \cite{yin2024tfad} construct an image decoder with image–label cross-attention, a text decoder handling textual descriptions, and an image–text decoder with MSA, whose outputs are combined to produce final multi-label prediction scores.

\begin{figure}[t]
  \centering
  \subfigure[Feature pyramid.]{
    \label{fig:feature-pyramid}
    \includegraphics[width=0.24\linewidth]{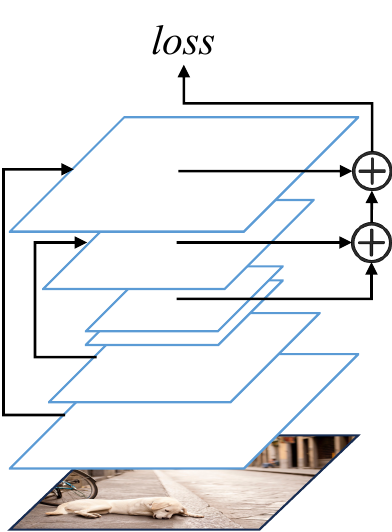}
    }
  \hspace{-1.2em}
  \subfigure[Kernel pyramid.]{
    \label{fig:kernel-pyramid}
    \includegraphics[width=0.49\linewidth]{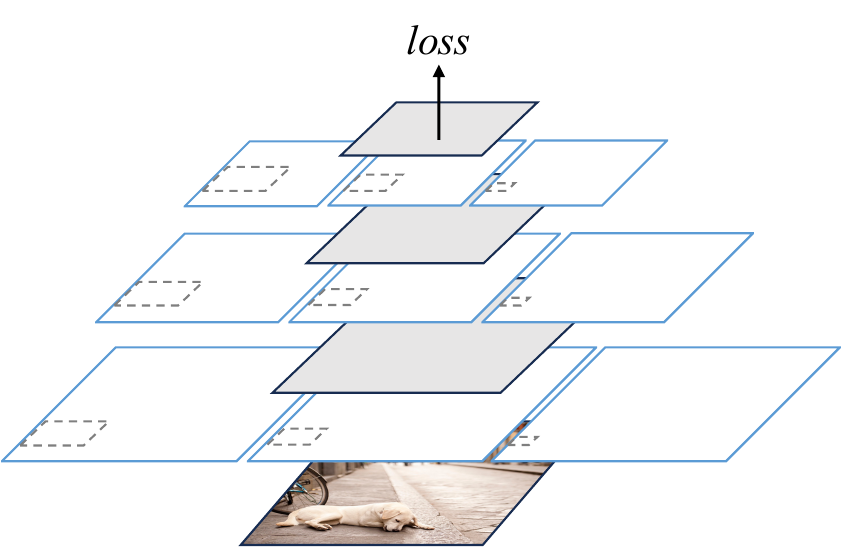}
    }
  \hspace{-1.2em}
  \subfigure[Loss pyramid.]{
    \label{fig:loss-pyramid}
    \includegraphics[width=0.23\linewidth]{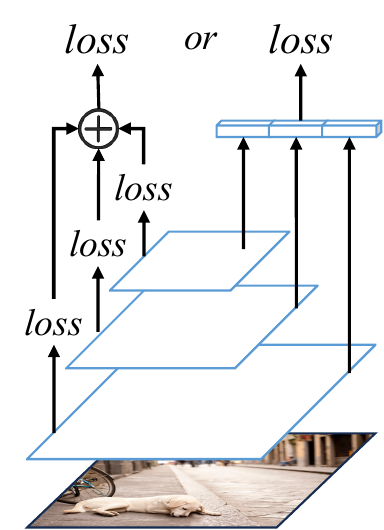}
    }
  \caption{The pipeline of pyramidal multi-scale architectures in MLIC.}
  \label{fig:multi-scale-arch}
\end{figure}

\subsubsection{Multi-Scale Architectures}

Multi-scale architectures exploit both high-level and low-level CNN features for MLIC.

\textbf{Feature Pyramid.} Feature pyramid is common in fusing multi-scale CNN feature maps. As shown in Fig. \ref{fig:feature-pyramid}, feature maps from different CNN layers are resized to a common resolution and then fused for MLIC.
Typical methods \cite{yan2019multi,zhang2022spatial,wang2024semantic} generate same-resolution feature maps through a feature pyramid network \cite{lin2017feature}, using top-down and lateral connections to normalize spatial resolution and align semantic levels across the pyramid.
Other studies \cite{gu2022multilabel,dai2023global,zhou2023feature,chen2024msfa,chen2025modeling,zhong2025multi} achieve multi-scale feature maps with consistent resolution by employing spatial dimension sampling techniques, such as bilinear-interpolation-based up-sampling \cite{gu2022multilabel,dai2023global,zhong2025multi} and down-sampling \cite{chen2025modeling}, nearest-neighbor-based up-sampling \cite{chen2024msfa}, and deconvolution-based up-sampling \cite{zhou2023feature}. In \cite{gu2022multilabel}, pyramidal feature maps are used to produce distribution maps of salient semantics, suppressing object-irrelevant responses and denoising features.

\textbf{Kernel Pyramid.} Kernel pyramid, as shown in Fig. \ref{fig:kernel-pyramid}, refers to a split-transform-merge process that utilizes convolutional kernels with different sizes to produce multi-scale feature maps on either a single layer or multiple layers of CNNs.
In \cite{yu2019delta}, the spatial pyramid convolutional transfer layer contains several parallel convolutions with various kernel sizes and operates on convolutional feature maps for capturing object features at multiple scales.
Yang et al. \cite{xianhua2021image} build bottleneck modules on the last two backbone layers by replacing ordinary convolution with pyramid convolution. Ye et al. \cite{ye2025multi} employ convolution kernels with different size to convert the dimensions of multi-scale feature maps, which are then fused to enhance features of small objects for MLIC.

\textbf{Loss Pyramid.} Loss pyramid (Fig. \ref{fig:loss-pyramid}) either sums the scale-wise losses or calculates the loss over the concatenated multi-scale features.
Many studies \cite{li2020multi,you2020cross,liang2022multi,chen2024msfa,wang2025establishing} first derive scale-specific features from multi-resolution feature maps via global pooling or attention, then predict label scores at each scale and combine scale-wise losses into a pyramidal loss for training. In \cite{wu2020adahgnn,kuang2023multi}, a hypergraph neural network or MSA is applied to enhance multi-scale feature maps, which are then pooled and concatenated to compute a pyramidal loss.
In \cite{li2020multi}, feature maps at each scale are pooled to predict label scores, and the resulting multi-scale scores are combined via weighted summation to compute the cross-entropy loss.
MSRN \cite{qu2023multi} obtains local and group-shared semantic representations at multiple resolutions, then concatenates them across scales for pyramidal loss.

\textbf{Others.} Many work \cite{zhou2020classify,zhao2021transformer,kuang2024multi,zhou2023datran,zhou2023feature,liu2025transformer,jiu2025multi} develops cross-scale attention to explicitly model the correlations among feature maps at different scales. Specifically, typical methods \cite{zhao2021transformer,kuang2024multi} up-sample multi-scale feature maps to a common resolution and apply cross-attention to model position-wise relationships across scales, enhancing structure information of small objects. Recent studies \cite{zhou2023datran,zhou2023feature,liu2025transformer} use MSA to capture cross-scale correlations by treating high-level feature maps as queries and low-level feature maps as keys and values. Jiu et al. \cite{jiu2025multi} integrate multi-order geometric contexts via cross-scale feature aggregation, enabling effective modeling for complex scenes.
Besides cross-scale attention, Zhao et al. \cite{zhao2020double} extract multi-scale image features using downsamplings and upsamplings in parallel branches at each CNN layer, learning a scaling policy to capture scale variations in images. In \cite{ouyang2023hsvlt}, multi-scale feature maps are gathered to learn interpretable global context and rescale the spectrum’s concentration. Wang et al. \cite{wang2016beyond} apply stochastic scaling and cropping to images, while Park et al. \cite{park2020marsnet} handle images of varying sizes using horizontal and vertical pooling. In \cite{wang2021semantic}, adaptive pooling is applied to CNN outputs, yielding multi-scale feature maps for MLIC.

\subsubsection{Other Architectures} 

Extensive attempts have also been made to develop other effective architectures for MLIC.

\textbf{Feature Attentions.} Diverse attention modules are explored in MLIC for feature enhancement. Typical approaches \cite{liu2019decoupling,zhou2023double,dai2023global} adopt channel attention modules to model inter-channel dependencies of feature maps, such as attention sub-network in \cite{liu2019decoupling} that learns channel-wise attention maps and GAT in \cite{zhou2023double} that captures correlations among channels. In \cite{li2020multi}, a spatial attention module is created to exploit spatial relationships of feature maps. Furthermore, many work \cite{woo2018cbam,zhao2020double,zhou2023datran,zhou2023mining,zhou2023attention,pu2023semantic} integrates both channel and spatial attention to jointly model channel-wise and spatial dependencies of feature maps. Recently, cross-modal attention modules, such as dual-modal attention \cite{tan2024pvlr,tan2024sspa}, double attention \cite{zhou2023double}, and image-text powered attention \cite{yin2024tfad}, are proposed to enhance cross-modal interactions and representations.
NOAH \cite{li2024noah} sequentially splits, transforms and merges features to capture spatially dense category-specific attentions at local to global scales for improved performance in MLIC.

\textbf{Others.} Many alternative architectures have been explored specifically for MLIC. Early methods \cite{vallet2015multi,le2016fully,song2018deep,cheng2018multi,wang2019baseline,lydia2020multi} are largely confined to modest extensions of CNNs. Moving beyond simple CNN variants, subsequent approaches focus on customized classification heads for multi-label predictions, such as multi-label K neighborhood classification head \cite{li2019classification}, multi-head residual attention \cite{zhu2021residual}, Transformer decoder-based scalable head \cite{ridnik2023ml}, and deep context-aggregation head \cite{jiu2025deep}. In addition, a variety of distinct architectural paradigms have been investigated for MLIC, including hierarchical architecture \cite{wang2022hierarchical}, interpretable architectures \cite{dong2022towards,sovatzidi2023towards}, automated black-box testing framework \cite{hu2023atom}, Mixture-of-experts architecture \cite{yin2024hybrid}, fully graph-based architecture \cite{yao2025gkgnet}, and Mamba-based architectures \cite{du2025mlmamba,zhu2025mambaml}.

\subsection{Representation-Oriented Methods\label{sec:representation-methods}}

In this subsection, we focus on approaches that advance MLIC by learning strong representations. 

\subsubsection{Contrastive Representations}

Contrastive learning (CL) produces strong representations by pulling semantically similar instances together and pushing dissimilar ones apart.

\textbf{Label-Correlation-Agnostic.} Typical CL methods in MLIC learn discriminative label-wise visual representations without accounting for label correlations. 
Early methods \cite{dao2021multi,dao2023contrastively} aim to pull features of identical labels closer and push different labels apart.
Subsequent studies \cite{hassanin2022learning,ma2023semantic} maintain label prototypes for CL. 
MCL \cite{hassanin2022learning} minimizes Euclidean distances between label representations and their prototypes while maximizing distances between different prototypes.
In \cite{ma2023semantic}, sample-sample and sample-prototype contrasts are performed within and cross images in a batch.
Instead, Fu et al. \cite{fu2024generative} generate contrastive samples by fitting label semantic space with a Gaussian mixture model (GMM).
In \cite{wang2023mumic}, CLIP-style image-text contrast is applied with a tempered sigmoid binary cross-entropy (BCE) loss.
While Xu et al. \cite{xu2023research} contrast label quantities to control positive and negative similarities,
cross-image foreground–background CL \cite{zhou2025drtn} is developed to align foreground features and push them away from background, yielding label-agnostic activation maps.

\textbf{Label-Correlation-Aware.} Many research focuses on label-correlation-aware CL for MLIC. Typical methods \cite{sajedi2023end,sajedi2024probmcl} form positive sets by selecting samples that share enough labels with an anchor image, pulling them closer and pushing others apart to capture label dependencies, and adopt GMMs to model epistemic uncertainty by encoding the presence, spatial location, and statistical complexity of labels.  Sajedi et al. \cite{sajedi2023end} use the Bhattacharyya coefficient to measure similarity between label representations, forming positive and negative pairs for kernel-based CL that increases the similarity of frequently co-occurring labels. They \cite{sajedi2024probmcl} further utilize a correlation coefficient as the similarity metric between Gaussian mixtures and define a probabilistic CL to emphasize samples sharing more labels with the anchor. Zhang et al. \cite{zhang2024multi} capture multi-label correlations at the image level by weighting sample–anchor relations according to label overlap.

\subsubsection{Debiased Representations}

In MLIC, object context can bias label classifiers, causing false negatives in rare contexts and false positives for frequent co-occurrences, mainly through confounding effect or mediating effect.

\begin{figure}[t]
  \centering
  \subfigure[Confounding effect.]{
    \label{fig:confounding-effect}
    \includegraphics[width=0.28\linewidth]{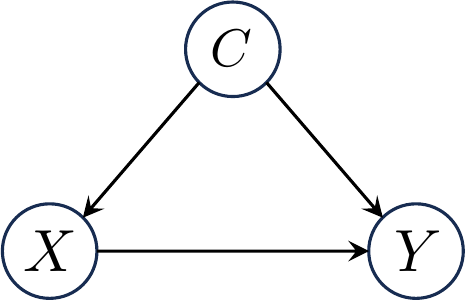}
    }
  \hspace{2em}
  \subfigure[Mediating effect.]{
    \label{fig:mediating-effect}
    \includegraphics[width=0.28\linewidth]{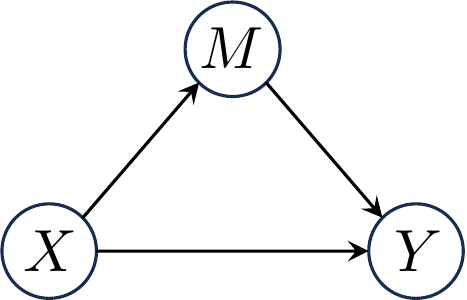}
    }
  \caption{Confounding effect and mediating effect in causal theory. (a) $C$ is the confounder set that builds a spurious causal relation between $X$ and $Y$. (b) $M$ is the mediator set that offers an indirect effect between $X$ and $Y$.}
  \label{fig:causal-inference}
\end{figure}

\textbf{Confounding Effect.} In MLIC, confounders are prior contextual knowledge from biased datasets, pretrained models, or training process, which distort the causality between an object and its prediction, as illustrated in Fig. \ref{fig:confounding-effect}. For de-confounded training, backdoor adjustment \cite{liu2022contextual} is commonly used:
\begin{equation}
    P(Y|do(X))=\sum_c P(Y|X,C=c)P(C=c).
    \label{eq:do-operation}
\end{equation}
Here $C$ is the confounder set; $do(X)$ performs causal intervention by cutting off the link from the confounder $C$ to the cause $X$, pursuing the causality between the cause $X$ and the effect $Y$ without confounding. However, $C$ is unobservable and potentially infinite, making it infeasible to enumerate all confounders.
Early methods \cite{liu2022contextual,liu2023causality} reformulate Eq. (\ref{eq:do-operation}) via inverse probability weighting, where the reverse weight uncovers a mapping between the values of $(y, x)$ and the confounder $c$, allowing $P(c)$ to be bypassed when conditioning on $X$. CCD \cite{liu2022contextual} uses an energy model to approximate the intervened effect by multi-head sampling of confounded logits. IDA \cite{liu2023causality} obtains sample-specific class-aware features by multiple sampling to re-weight samples, compensating spatial attention against confounders and yielding robust class-specific representations.
Recent studies \cite{tian2023causal,chen2024pursuit} approximate Eq. (\ref{eq:do-operation}) by normalized weighted geometric mean, requiring explicitly modeling the confounder set $C$ and its prior $P(C)$.
For $C$, CMLL \cite{tian2023causal} uses a predefined confounder set where each entry is the average regional feature of a label over the whole dataset, while Chen et al. \cite{chen2024pursuit} cluster spatial features of all images from a pretrained CNN.
For $P(C)$, CMLL initializes it from the label distribution in the confounder set and updates it during training, while \cite{chen2024pursuit} estimates it from data.
With $C$ and $P(C)$ ready, CMLL and \cite{chen2024pursuit} perform causal intervention by self-attention and  cross-attention, respectively.

\textbf{Mediating Effect.} The mediator set $M$ lies between cause $X$ and effect $Y$, explaining how $X$ affects $Y$. It exploits context to boost performance but also makes predictions rely on mixed object–context cues rather than pure instance features, as shown in Fig. \ref{fig:mediating-effect}. To mitigate this effect, counterfactual reasoning is used to measure the Total Direct Effect (TDE) \cite{liu2022contextual}, strengthening the direct causal effect along $X\rightarrow Y$:
\begin{equation}
\begin{aligned}
    \mathrm{TDE}(Y) =\ &P(Y|X=x,M=m) \\ & - P(Y|X=x_0,M=m),
    \label{eq:tde}
\end{aligned}
\end{equation}
where $x_0$ indicates turning off the link $X\rightarrow Y$, thus the prediction relies solely on the mediators $M$. 
For $P(Y|X=x_0,M=m)$, CCD \cite{liu2022contextual} uses early-training per-label mean features as contextual confounders, then weights label prototypes by their probabilities to form an image-specific context as the mediator $M$.
PAT \cite{xie2024counterfactual} uses correlative features between co-occurring objects as the mediator by masking the object $x$ from an image using a patch-based strategy. With the mediator $M$ constructed, both methods \cite{liu2022contextual,xie2024counterfactual} perform TDE inference to train a model free from contextual bias.

\subsubsection{Distilled Representations}
Knowledge Distillation (KD) transfers knowledge from a teacher model to a student one to produce distilled representations for MLIC. In \cite{liu2018multi}, both feature-level and prediction-level knowledge are distilled from a detection model to boost a MLIC model by minimizing the $\ell_2$ loss between their region features and predictions. A calibrated branch is introduced in \cite{song2021handling} for label prediction and uncertainty estimation, guiding a student network to handle difficult labels via uncertainty distillation. Subsequent methods \cite{xu2022boosting,yang2023multi,wang2024beyond} focus on capturing label correlations. By decomposing MLIC into sub-tasks, PSD \cite{xu2022boosting} uses category-specific sub-task networks as teachers and optimizes the student with MSE-based KD, balancing overfitting and label correlations. L2D \cite{yang2023multi} creates intra-class and inter-class relations and explores structural knowledge through label-wise embedding distillation with Huber loss. In \cite{wang2024beyond}, knowledge is distilled at both the batch and instance levels to capture multi-order label-pair dependencies, boosting transfer from teacher to student. Li et al. \cite{li2024multi} bridge the teacher–student architecture using mutual information, and integrate structured information, logits, and BCE distillation for MLIC.

\subsubsection{Metric Representations} 
Deep metric learning measures the similarity between samples by learning a representation function. In MLIC, Shi et al. \cite{shi2017training} propose a max-margin loss to enforce a margin between positive and negative labels, a max-correlation loss to align image features with ground-truth labels, and a correntropy loss to minimize training errors. In \cite{li2019reconstruction}, a bidirectional deep distance metric is learned so that an image is closer to its true label embedding than to other label embeddings or to image embeddings of neighboring labels. Subsequent methods \cite{zhang2019multi,chen2019multi2,chen2020disentangling,zhou2021multi} focus on metric learning for capturing label correlations. Zhang et al. \cite{zhang2019multi} propose a view-specific distance metric with manifold regularization to preserve intrinsic feature-space geometry, then concatenate all views to learn a unified distance metric. Chen et al. \cite{chen2019multi2} minimize Euclidean distances between correlated labels while pushing uncorrelated labels away from the correlated centroid. They further introduce a ranking loss \cite{chen2020disentangling} to encourage more compact and discriminative label vectors by enforcing shorter distances to correlated than to uncorrelated labels. CPCL \cite{zhou2021multi} measures relative appearance discrepancies of features under category-specific composing and decomposing transformations for MLIC.

\subsection{Learning-Oriented Methods\label{sec:learning-methods}}

This subsection reviews approaches that mainly focus on diverse and specialized  learning paradigms for MLIC.

\subsubsection{Learning to Rank}
Learning-to-rank reformulates MLIC from a set of binary decisions into a label-ranking task, shifting supervision from per-label correctness to global ordering consistency to exploit relative relevance among labels.

\textbf{Ranking-Loss-Only.}
Sampling-based methods \cite{gong2013deep,li2017improving,ma2022deep} learn relative preferences between positive and negative labels by sampling label pairs. Gong et al. \cite{gong2013deep} propose a weighted approximate ranking loss that penalizes positives more when they rank low, repeatedly sampling negatives until a violation occurs. While LSEP \cite{li2017improving} offers a smooth, differentiable approximation to the pairwise hinge loss to ease optimization, Ma et al. \cite{ma2022deep} impose inter-label margins at selected layers using any $\ell_p$ ($p\geq1$) norm, making the loss adaptable to diverse architectures. Both \cite{li2017improving,ma2022deep} sample only a fixed number of label pairs, resulting in computational complexity that scales linearly with the number of labels. To avoid the complexity and randomness of sampling, a dynamic weighted Euclidean loss \cite{wang2016beyond} is framed to adaptively assign larger penalties to ground-truth labels with low-ranked scores, pushing them to the top without explicit negative sampling.

\textbf{Joint-Loss.}
Another line of work \cite{shi2017training,song2018deep,seymour2018multi,lyu2018coarse,lyu2019multi,wen2020multilabel} jointly optimizes BCE and ranking losses, endowing models both strong classification accuracy and explicit ranking capability, thus better handling label imbalance and dependencies. Additionally, several approaches \cite{yeh2019multilabel,zhou2021multiple} feed an image into a CNN to produce a transformation matrix that maps label embeddings into an image-conditioned space, where labels are ranked by their $\ell_2$ norms. This enables the use of any pairwise ranking loss, effectively performing learning-to-rank in an image-adaptive label-embedding space.

\subsubsection{Multi-Task/Instance Learning}
Multi-instance and multi-task learning exploit structure beyond image–label pairs, offering richer and more informative supervision that yields better representations and improved generalization in MLIC.

\textbf{Multi-Task Learning.} MLIC can be recognized as a special case of multi-task learning, treating each label as a separate classification task.
Huang et al. \cite{huang2013multi} extend one classification task of traditional network into multiple binary classification tasks by defining the output layer with a negative and positive node per label, enabling intermediate representations shared across tasks.
In \cite{coulibaly2022deep}, multiple CNN-based sub-networks are trained jointly and fused via multi-output heads or multi-feature fusion to learn shared and complementary task representations.
Markatopoulou et al. \cite{markatopoulou2018implicit} present each concept as a sparse, shared linear combination of latent features, and model label correlations by a structured-output loss, capturing inter-concept relations at both feature and label levels.
Yin et al. \cite{yin2024hybrid} treat label heterogeneity as negative transfer and use a mixture-of-experts Transformer with task-specialized and shared experts to share information among related labels while decoupling conflicting ones.
Xu et al. \cite{xu2022boosting} decompose MLIC into sub-tasks via spectral clustering on label co-occurrence graphs, train parallel sub-models on different label subsets to learn joint and category-specific patterns, and then distills their knowledge into a global model for MLIC.

\textbf{Multi-Instance Learning.}
In MLIC, multi-instance learning offers a natural way to model images as bags of regions with multiple associated labels. Chen et al. \cite{chen2013multi} build a two-stage neural network, where the first MLP learns region–label relationships and the second MLP aggregates regional outputs to capture label correlations. Yan et al. \cite{yan2016new} first extract latent topic features for each bag via probabilistic latent semantic analysis and then apply a neural network for label prediction, thus filtering low-level noise and enhancing semantic discriminability. As deep learning emerges, Guo et al. \cite{guo2018deep} integrate multi-instance learning into CNNs for MLIC, enabling end-to-end training on real-world multi-labeled images. Song et al. \cite{song2018deep} extend this idea with a deep multi-modal CNN, where jointly processes images and label-group descriptions by constructing multi-modal instances and learning an instance-level scoring function for image-level predictions.

\subsubsection{Other Learning Methods}

Many efforts have also been made to explore diverse learning paradigms for MLIC.

\textbf{Matching Learning.} Matching learning views MLIC as a matching task between two sets and learns a scoring function that assigns higher scores to correct pairs than to incorrect ones.
Typical methods \cite{wu2021gmmlic,wu2023semantic} build a dynamic instance spatial graph and a label semantic graph, merge them into an assignment graph by linking each instance to every label, and produce instance–label matching scores via message passing, enabling to jointly model spatial relations, label semantics, and their correspondences.
Wu et al. \cite{wu2023transformer} further replace hand-crafted graphs by a Transformer, using region proposals plus a global instance as the instance set, where self-attention models instance and label structures and cross-attention infers instance–label matches end-to-end.
In \cite{li2023patchct}, MLIC is formulated as set-matching problem between patch and label embeddings, which are aligned in a shared space via a bidirectional conditional transport distance.

\textbf{Dictionary Learning.}
In MLIC, dictionary learning seeks to acquire a set of visual atoms, represent each image as a sparse combination of these atoms, and associate the resulting sparse codes to labels. By sharing atoms across different labels, it can capture inter-label correlations and enhance the robustness of label recognition. Recent approaches \cite{zhou2021deep,zhang2024dbddl} have attempted to extend dictionary learning from traditional statistical learning \cite{cao2015sled,jing2016multi,niu2019coupled} to deep learning for MLIC. Specifically,
Zhou et al. \cite{zhou2021deep} learn a semantic dictionary from class-level semantics using auto-encoder, represent CNN features with label embeddings in this dictionary space, and optimize the dictionary and codes via alternating updates. 
Zhang et al. \cite{zhang2024dbddl} separately build semantic and detail dictionaries via dedicated neural modules, then merge them into a comprehensive dictionary for MLIC.

\textbf{Others.} 
Diverse learning mechanisms have been explored for MLIC. He et al. \cite{he2018reinforced} introduce curriculum learning into MLIC, mimicking human behavior by labeling images from easy to complex and using a reinforcement learning agent to sequentially predict labels. In \cite{zhou2021multi}, compositional learning is introduced to represent each image as category-related features and learn to compose and decompose them with category prototypes for MLIC.
In addition, Li et al. \cite{li2016conditional} formulate MLIC as conditional graphical Lasso inference problem and propose a unified Bayesian framework that jointly learns the graph structure and parameters conditioned on image features.
Marino et al. \cite{marino2017more} cast MLIC as reasoning over a knowledge graph, where detector scores initialize nodes in an object–attribute graph and message passing is performed to leverage external knowledge.
While ensemble learning is employed in \cite{wang2019baseline,abulfaraj2025deep} as a voting system that combines multiple classifiers, adversarial learning is introduced in \cite{zhou2020classify} to train networks robust to occlusion.
Moreover, privacy-preserving learning \cite{xu2022privacy} is investigated to provide a more comprehensive understanding of MLIC.

\subsection{Data-Oriented Methods\label{sec:data-methods}}

This subsection reviews approaches that focus on the statistical distribution and tailored augmentation of multi-label data, as well as the rich knowledge of large-scale data.

\subsubsection{Data Imbalance}
Data imbalance is a key challenge in MLIC, appearing at both intra-label and inter-label levels.
 
\textbf{Intra-Label Imbalance.} Intra-label imbalance means that negative samples greatly outnumber positive ones for a given label. To solve this, Yan et al. \cite{yan2019imbalance} build rectification methods to offset or constrain the tendency, which detects class imbalances by feeding random noise into label classifiers.
ASL \cite{ridnik2021asymmetric} seperately adjust the decay rates for positive and negative samples, boosting the contribution of positive samples to the loss. Further, the strictly proper asymmetric loss \cite{cheng2024towards} enhances per-sample calibration constraints for better-calibrated probability estimates.
Yuan et al. \cite{yuan2023balanced} develop a balanced masking strategy for graph-based methods by selectively manipulating node embeddings to exploit majority-sample priors while preserving class balance.
In \cite{yuan2024positive}, models are trained using only positive labels, with adaptive re-balancing factors incorporated into the loss to address label imbalance.

\textbf{Inter-Label Imbalance.}
Inter-label imbalance characterizes a long-tailed distribution of positive samples across labels.

\raisebox{0.3ex}{\scriptsize $\bullet$} \textit{Loss-Based:} Many efforts focus on loss design to alleviate the imbalance. Wu et al. \cite{wu2020distribution} re-weight positive and negative terms based on label co-occurrence and use a negative-tolerant regularizer to reduce over-suppression of negatives. Lin et al. \cite{lin2023probability} introduce probability re-balancing with an adaptive focal term to correct the distortion of cost-sensitive methods. They further present a distributionally robust variant of LSEP \cite{lin2025distributionally} by introducing class-wise LSEP with a negative gradient constraint. While robust asymmetric loss \cite{park2023robust} reduces sensitivity to hyper-parameters, balanced ASL \cite{timmermann2025lm} emphasizes rare and uncertain positive samples. Wei et al. \cite{wei2025delving} capture predictive complex patterns to enrich the representations of tail samples.
Guo et al. \cite{guo2021long} propose collaborative training on uniform and re-balanced samplings with BCE-based losses, logit compensation, and a cross-branch consistency loss. Chen et al. \cite{chen2025towards} apply group sampling and a dynamic loss to equalize gradients across categories, preventing tail overfitting while improving classifier discriminability. Tao et al. \cite{tao2025mlc} use collapse calibration with binarized classifiers to reduce intra-class variation and classification bias, thus shaping the feature geometry to better handle inter-label imbalance.

\raisebox{0.3ex}{\scriptsize $\bullet$} \textit{Semantic-Based:} Many works exploit semantic priors to link head and tail labels. Xia et al. \cite{xia2023lmpt} combine prompt tuning with a class-specific embedding loss to capture head–tail semantic relations. Yan et al. \cite{yan2024category} refine category prompts via progressively injecting visual information. Du et al. \cite{du2025category} enhance class-specific features through selective feature enhancement. While Zhang et al. \cite{zhang2025federated} finetune CLIP to establish head–tail semantic correlations under federated settings, Tang et al. \cite{tang2025unleashing} use CLIP's text encoder to capture label correlations and realign visual-textual modalities, averting overfitting on tail classes without compromising head-class performance.
More recently, Fan et al. \cite{fan2026channel} initialize label embeddings with CLIP to capture inter-class semantic correlations and enforce visual–semantic consistency. In \cite{fan2026exploring}, logit calibration is framed to reduce head–tail probability bias.

\subsubsection{Data Augmentation}

Data augmentation in MLIC mainly includes transformation-based and erasing-based methods.

\textbf{Transformation‑Based.} Transformation‑based approaches mainly apply geometric or appearance transformations to learn robust representations for MLIC.
Wang et al. \cite{wang2019baseline} generate synthetic samples by interpolating pairs of images and labels to increase data variability.
Wang et al. \cite{wang2025splicemix} augment both sample space and batch scale by using semantically blended mixed images to alleviate co-occurrence bias, and splice regular and mixed images to enforce cross-scale consistency training.
Visual attention consistency \cite{guo2019visual,wang2021g} enforces the attentional regions to undergo the same spatial transformation (e.g., flipping) as the input image, whereas semantics consistency \cite{chu2021multi} requires the underlying image semantics to remain unchanged under geometric operations.

\textbf{Erasing‑Based.} Erasing-based studies prevent models from over-reliance on a few discriminative image areas by removing or masking them.
Zhong et al. \cite{zhong2020random} erase a randomly selected image region with random values, 
which may remove only background or entire objects, leading to suboptimal training. Recent studies \cite{pu2023semantic,zhou2023feature,zhou2025drtn} explore attention-guided and object-centered erasing strategies to better regularize network training.
Pu et al. \cite{pu2023semantic} normalize a spatial attention matrix into marginal distributions and erase dominant attention regions of high-confident labels.
Both \cite{zhou2023feature,zhou2025drtn} yield importance maps to highlight discriminative regions, then mask them to force the model to explore remaining regions for MLIC.


\subsubsection{Large-Scale Pretrained Knowledge }

Knowledge derived from pretraining on large-scale data has been widely explored to substantially improve MLIC performance, primarily involving multi-modal and single-modal knowledge.

\textbf{Cross-Modal Knowledge.} In MLIC, cross-modal knowledge is primarily derived from CLIP \cite{radford2021learning}, which is trained on large-scale image–text pairs and acquires a vast, generalizable understanding of visual objects, textual concepts, and their alignments. To leverage such cross-modal knowledge for MLIC,
typical methods \cite{tan2024pvlr,tan2024sspa,rawlekar2024improving} use CLIP to extract both textual embeddings and visual features. 
Specifically, Tao et al. \cite{tan2024pvlr} apply dual-modal attention to model bidirectional interaction between visual and linguistic features, whereas Tan et al. \cite{tan2024sspa} leverage gated dual-modal alignment to refine their cross-modal correspondence.
Rawlekar et al. \cite{rawlekar2024improving} enhance per-class independent classifiers with class co-occurrence information to refine initial image–text predictions.
In addition, Zhang et al. \cite{zhang2024recognize} first build a large-scale dataset containing 6,449 labels from massive image-text pairs, and then use the CLIP text encoder to extract label semantics as queries, which are fed into a Transformer decoder for image tagging. 

\textbf{Single-Modal Knowledge.} In MLIC, single-modal knowledge refers to linguistic and visual knowledge learned from large-scale text corpora and image collections, respectively.

\raisebox{0.3ex}{\scriptsize $\bullet$} \textit{Visual Knowledge:} The most widely recognized source of visual knowledge is ImageNet \cite{deng2009imagenet}, whose pretrained networks have become the default backbones for visual feature extraction in MLIC. Wu et al. \cite{wu2019tencent} build a large-scale multi-label image database to train ResNet-101 for large-scale MLIC. Segmentation priors from SAM \cite{kirillov2023segment}, learned from large-scale semantic segmentation data, have also been leveraged for MLIC. In \cite{li2023semantic}, an image segmentation model is proposed that employs a query-based mask decoder to produce semantic-aware, multi-granularity masks for segmenting and recognizing objects at any desired granularity. Yuan et al. \cite{yuan2024open} further integrate CLIP and SAM into a unified framework, distilling segmentation knowledge from SAM into CLIP while feeding complementary recognition knowledge from CLIP back into SAM, providing an effective tool for instance-level labeling.

\raisebox{0.3ex}{\scriptsize $\bullet$} \textit{Linguistic Knowledge:} In MLIC, most linguistic knowledge comes from the pretrained language models \cite{mikolov2013efficient,pennington2014glove} for label embedding. With the rise of LLMs, their world knowledge acquired from massive text corpora has greatly enhanced MLIC. 
Fang et al. \cite{fang2025amita} use LLMs to obtain category attributes and design prompts to enhance their differences, strengthening semantic distinction among categories.
In \cite{tan2024sspa}, inherent knowledge in LLMs is used to generate detailed descriptions, distinguishing objects from other similar categories and discovering label relationships.
Yin et al. \cite{yin2024tfad} generate multi-granularity image descriptions using LLMs to enhance visual label classification. 
ASM \cite{wang2023all} comprises an image tokenizer and an LLM-based decoder, trained on a large-scale dataset with next-token prediction and region-text alignment to achieve open-world recognition and understanding.
In addition, a new paradigm \cite{guo2023texts,zhu2026autoit} is proposed for MLIC that generates multi-label texts using LLMs for prompt learning or adapter optimization in CLIP embedding space, enabling visual label recognition without manually labeled images.

\section{Challenges and Outlooks}
\label{sec:future}

\subsection{An Open Problem Beyond MLIC}

\begin{figure}
    \centering
    \includegraphics[width=\linewidth]{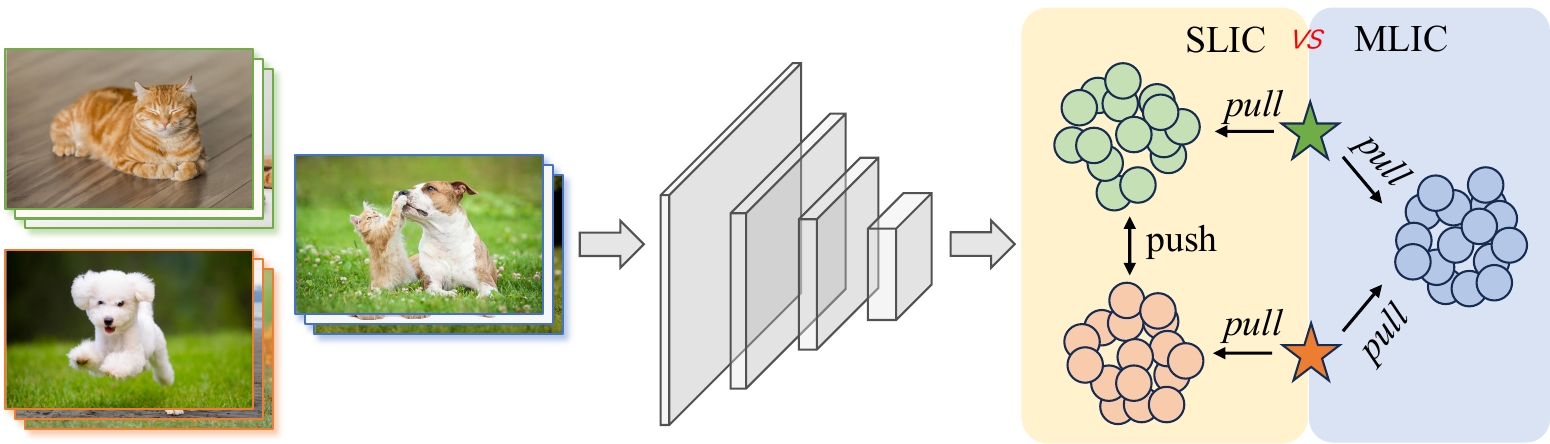}
    \caption{An example of a learning game between single-label and multi-label images. Circles and pentagrams represent image features and classifier vectors, respectively. Images and their representations are associated with color.}
    \label{fig:game}
\end{figure}

\begin{figure}[t]
  \centering
  \subfigure{
    \includegraphics[width=0.46\linewidth]{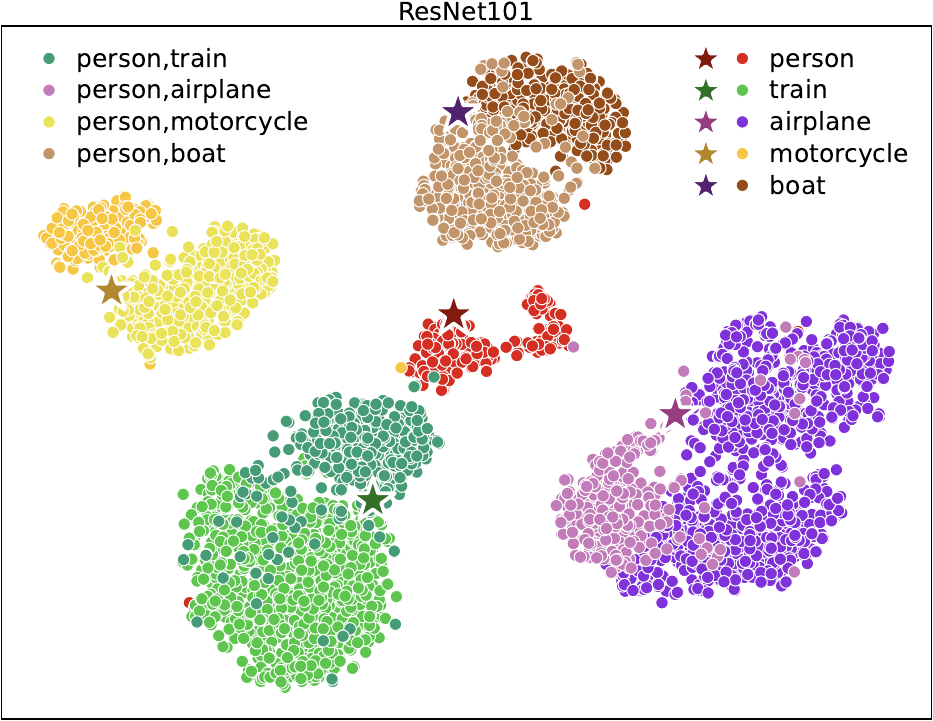}
    }
  \hfill
  \subfigure{
    \includegraphics[width=0.46\linewidth]{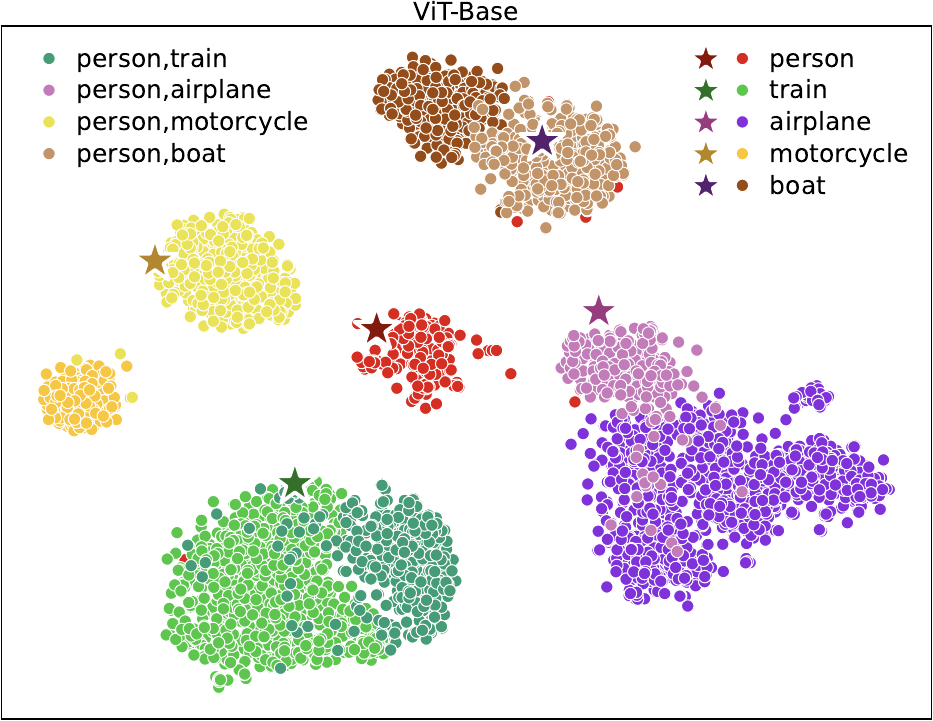}
    }
  \subfigure{
    \includegraphics[width=0.46\linewidth]{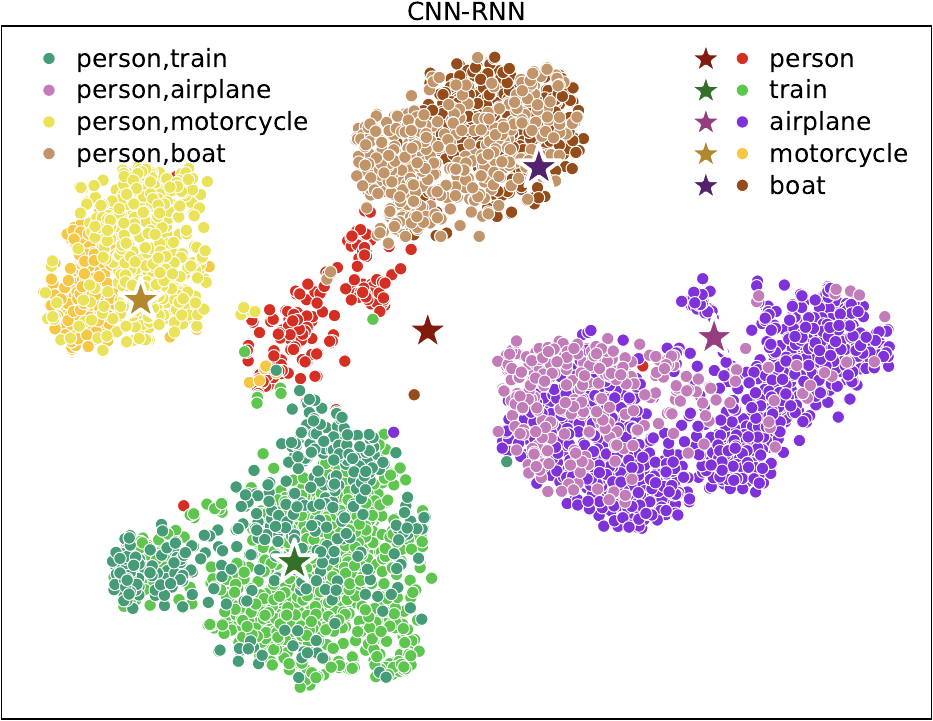}
    }
  \hfill
  \subfigure{
    \includegraphics[width=0.46\linewidth]{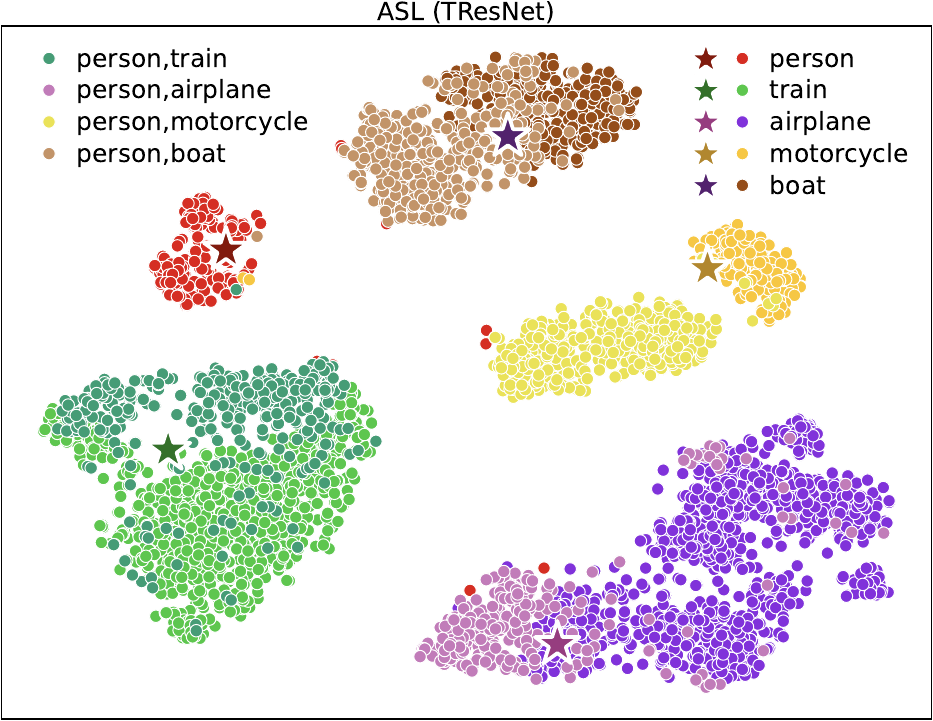}
    }
  \caption{Distributions of image representations and label classifier vectors for selected labels and label combinations from the Microsoft COCO training set. Label classifier vectors are marked with pentagrams.}
  \label{fig:tnse}
\end{figure}

Recent MLIC developments show that label-specific representation learning has become the dominant paradigm and consistently outperforms global representations. However, there is still little insight into why region-level local representations outperform global ones. We argue that this disparity stems from an inherent and often overlooked learning game between SLIC and MLIC, which is implicitly induced when training MLIC models with global representations. 
Consider a toy dataset as in Fig. \ref{fig:game}, where some images contain only cats or dogs, and others contain both. In the global representation paradigm, all images are encoded by a DNN into single feature vectors to train the label classifiers. For single-label images, an ideal solution places each classifier near the center of its own feature cluster and keeps different classifiers well separated. For multi-label images, however, their features must lie close to both the cat and dog classifiers to yield correct predictions, thus creating a conflict: pushing classifiers apart benefits SLIC, while pulling them closer is required by MLIC. We call this tension a learning game between SLIC and MLIC. Fig. \ref{fig:tnse} shows the experimental results for typical MLIC models. The classifiers for each label are not perfectly centered on the corresponding image feature cluster but are shifted due to the pull of image representations with combined labels, empirically revealing the SLIC-MLIC learning game.

Although largely overlooked by the community, the SLIC-MLIC learning game is partially alleviated in methods based on local representations \cite{chen2019learning,liu2021query2label}. By localizing object regions within an image and encoding each region separately, the global image representation are decomposed into several region-level local representations. Training then effectively degenerates into optimizing SLIC-style objectives on individual object regions, thereby sidestepping the direct conflict between SLIC and MLIC. However, region-based methods are inherently tied to object localization quality and create a conceptual gap between MLIC and SLIC, hindering methodological unification and causing MLIC to lag behind SLIC. To solve this issue, a promising direction is to replace static classifiers with image-conditioned dynamic classifiers, inspired the image-conditioned prompt learning \cite{zhou2022conditional}. Under this setting, classifier parameters are adapted on the fly according to the image content, encouraging large separation between classifiers for single-label images while still allowing them to adaptively move together for multi-label images.

The learning game is not confined to MLIC. Any task that learns a shared global representation to support both “pure” instances (associated with a single dominant concept) and “composite” instances (involving multiple objects, attributes, or relations) encounters a similar conflict. Representative examples include image retrieval, scene understanding, visual relationship detection, and video action or event recognition. In these cases, single-concept objectives favor well-separated prototypes in the representation space, whereas multi-concept objectives require the same space to allow certain prototypes to move closer, reproducing the SLIC–MLIC learning game in different guises. Furthermore, the learning game naturally arises in modern multimodal foundation models as well, where a shared embedding space must simultaneously support discriminative alignment for single-concept image–text pairs and flexible compatibility with multi-object or compositional descriptions. Evidently, an in-depth investigation of the SLIC-MLIC game is of great value to the broader CV community.

\subsection{Key Challenges and Research Directions}

\subsubsection{Small Objects} 
\begin{figure}
    \centering
    \includegraphics[width=1\linewidth]{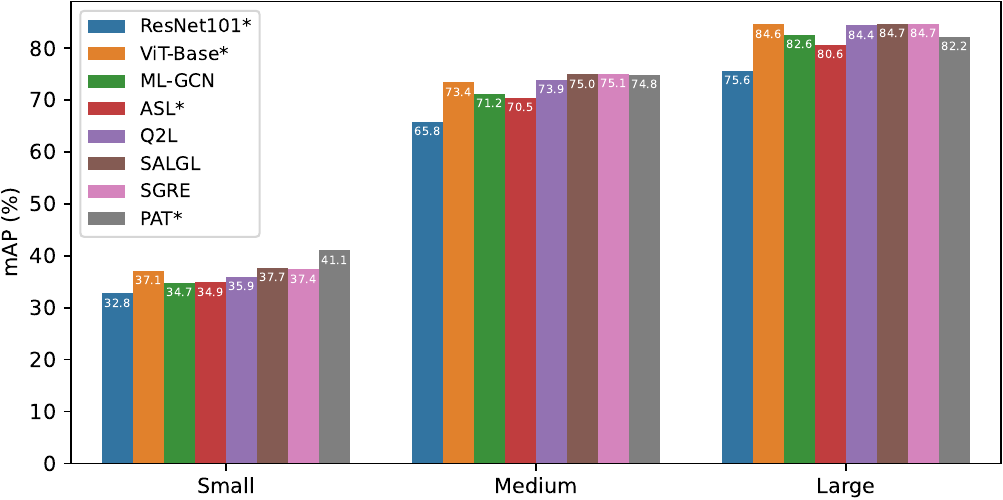}
    \caption{Performance of typical MLIC methods on objects of different sizes on the Microsoft COCO validation set, including ResNet101 \cite{he2016deep}, ViT-Base \cite{dosovitskiy2020image}, ML-GCN \cite{chen2019multi}, ASL \cite{ridnik2021asymmetric}, Q2L \cite{liu2021query2label}, SALGL \cite{zhu2023scene}, SGRE \cite{zhu2024semantic}, and PAT \cite{xie2024counterfactual}. Objects are categorized into three sizes based on area: small (area $<$ 32$\times$32), medium (32$\times$32 $\leq$ area $<$ 96$\times$96), and large (area $\geq$ 96$\times$96). The symbol * denotes models re-trained using official codes. Except for ViT-Base, all models use ResNet101 as the backbone network for fair comparison.}
    \label{fig:size}
\end{figure}

Identifying small objects is particularly challenging because they occupy few pixels that provide limited discriminative information and are easily overshadowed by large objects and cluttered backgrounds. Fig. \ref{fig:size} shows that small-object recognition remains a major performance bottleneck in MLIC. 
Future work may draw inspiration from the observation in Fig. \ref{fig:size} that patch-based methods (ViT-Base \cite{dosovitskiy2020image} and PAT \cite{xie2024counterfactual}) perform better on small-object recognition, and further exploit patch-level modeling to boost small-object sensitivity, such as multi-scale patch tokenization. In addition, exploiting label co-occurrence patterns and semantic relations offers a promising avenue. Contextual cues from reliably detected large objects can be leveraged to refine the predictions for small objects, while semantic embeddings can transfer information from semantically related and easier-to-recognize labels to harder small-object labels.

\subsubsection{Contextual Bias} 
In multi-label images, objects are commonly embedded in rich contexts formed by other objects and backgrounds, which can mislead classifiers by assigning high scores to nonexistent objects due to the presence of frequently co-occurring objects or backgrounds, and low scores to existent objects in rare contexts.
To handle such contextual bias, promising directions include designing explicit context-bias metrics such as per-class prediction stability under background perturbations, developing architectures that factorize objects and contexts, and building scalable causal pipelines for feature-level interventions that enforce stable predictions for an object when only the context changes.

\subsubsection{Attention Mechanism} Attention has become a de facto component of high-performance MLIC models for locating regions of interest. However, we still lack a clear understanding of how attention learns to focus on specific regions and which factors (e.g., label frequency, co-occurrence patterns) drive its behavior, making it hard to systematically diagnose and remedy failure cases.
Future work may develop attention diagnostics and metrics (e.g., alignment with human-annotated regions), design explicitly regularized attention, (e.g., intra-label attention sparsity and inter-label attention separation); and explor counterfactual interventions (e.g., masking attended regions) to assess the causal role of attention in predictions.

\subsubsection{Prompt Learning}
Prompt learning \cite{zhou2022conditional} has achieved remarkable progress in SLIC, yet its development in MLIC still lags behind. Present category-level prompts are weakly adaptive across diverse multi-label contexts, and struggle to capture multi-label characteristics.
Promising directions include adaptive prompt generation that conditions prompts on label co-occurrence patterns and contextual cues, enabling prompts to dynamically capture the specific multi-label configuration of each image. Another avenue is to integrate semantic structure (e.g., label hierarchies) into prompts so that the related labels can mutually reinforce each other while conflicting labels are differentiated. Systematically benchmarking these prompting strategies would be a key step toward improving performance and establishing a novel paradigm for MLIC.

\subsubsection{Novel Architecture}
A promising direction is to explore emerging architectures for MLIC beyond CNNs, RNNs, and ViT, which involve quadratic-complexity attention, rigid spatial grids, or suboptimal inductive bias in sequence modeling. Mamba \cite{gu2023mamba} offers linear-time sequence modeling with the capacity to capture dependencies among labels. RWKV \cite{peng2023rwkv} combines recurrent and attention-like mechanisms in a parallelizable framework, potentially enabling temporally coherent yet order-agnostic modeling of label sequences. KANs \cite{liu2024kan} introduce new function-approximation properties and structured nonlinearities that may better capture high-order interactions among labels. Future research can adapt these novel architectures to MLIC and systematically evaluate their effectiveness and efficiency against conventional baselines.

\subsection{Outlook on the Future of MLIC}

\subsubsection{Domain-Specialized MLIC}
In specialized domains such as medical imaging, meteorology, remote sensing, and chest X-rays, future MLIC approaches are expected to depend heavily on large foundation models. These models provide powerful general visual priors and rich semantics, which can be adapted to expert label systems through domain-specific finetuning, prompt-based adaptation, or parameter-efficient transfer. A key challenge and opportunity will be to tightly couple these models with domain knowledge, so that the resulting MLIC systems are reliable, generalizable, and trustworthy enough for deployment in safety-critical real-world applications.

\subsubsection{Data-efficient Learning}
CLIP \cite{radford2021learning} has enabled a new paradigm in which MLIC can be performed using only textual data \cite{guo2023texts,zhu2026autoit}, thereby bypassing the necessity for manually annotated images. 
This paradigm naturally supports data-efficient MLIC and is particularly beneficial in domains with scarce and costly training data, such as specialized medical imaging, scientific microscopy, and safety‑critical industrial inspection. Future MLIC is expected to develop effective mechanisms that enhance the cross-modal transferability of label prompts within the CLIP embedding space, and to study how limited amounts of high-quality images can be most effectively combined with abundant texts for MLIC.

\subsubsection{General Label Intelligence}
Recent advances in LLMs and VLP are profoundly reshaping the research landscape of MLIC. LLMs, with their strong capabilities in conceptual reasoning, commonsense inference, and structured knowledge manipulation, enable more principled modeling of label semantics, including hierarchies, correlations, and fine-grained attributes. Looking ahead, an important direction is to harness LLMs as a general “label intelligence” to automatically interpret, expand, and restructure label vocabularies. Overall, leveraging LLMs and VLP is expected to drive the next generation of research and development in MLIC toward more scalable, interpretable, and adaptable systems.

\section{Conclusion\label{sec:conclusion}}

This survey presents a comprehensive and systematic review of deep learning-based MLIC methods developed over the past decade and, to the best of our knowledge, is the first survey dedicated specifically to this topic from the joint perspectives of taxonomy, challenge, and outlook. We first introduce the essential background knowledge of MLIC, and then systematically organize and summarize existing methods by distilling broadly influential techniques from six perspectives: label-oriented, region-oriented, architecture-oriented, representation-oriented, learning-oriented, and data-oriented methods. Finally, we reveal the underlying learning game in MLIC, discuss open research challenges and promising directions, and provide an outlook on the future of MLIC. We hope this survey will offer readers a panoramic view of deep learning-based MLIC methods, inspire new research ideas, and ultimately foster further advances in the academic community.

\bibliographystyle{IEEEtran}
\bibliography{reference}

@inproceedings{yang2016exploit,
  title={Exploit bounding box annotations for multi-label object recognition},
  author={Yang, Hao and Tianyi Zhou, Joey and Zhang, Yu and Gao, Bin-Bin and Wu, Jianxin and Cai, Jianfei},
  booktitle={Proceedings of the IEEE/CVF conference on computer vision and pattern recognition},
  pages={280--288},
  year={2016}
}

@inproceedings{wang2017multi,
  title={Multi-label image recognition by recurrently discovering attentional regions},
  author={Wang, Zhouxia and Chen, Tianshui and Li, Guanbin and Xu, Ruijia and Lin, Liang},
  booktitle={Proceedings of the IEEE international conference on computer vision},
  pages={464--472},
  year={2017}
}

@inproceedings{chen2018recurrent,
  title={Recurrent attentional reinforcement learning for multi-label image recognition},
  author={Chen, Tianshui and Wang, Zhouxia and Li, Guanbin and Lin, Liang},
  booktitle={Proceedings of the AAAI conference on artificial intelligence},
  volume={32},
  number={1},
  pages={6730--6737},
  year={2018}
}

@article{zhang2018multilabel,
  title={Multilabel image classification with regional latent semantic dependencies},
  author={Zhang, Junjie and Wu, Qi and Shen, Chunhua and Zhang, Jian and Lu, Jianfeng},
  journal={IEEE Transactions on Multimedia},
  volume={20},
  number={10},
  pages={2801--2813},
  year={2018},
  publisher={IEEE}
}

@inproceedings{wu2021gmmlic,
  title={GM-MLIC: Graph Matching based Multi-Label Image Classification}, 
  author={Yanan Wu and He Liu and Songhe Feng and Yi Jin and Gengyu Lyu and Zizhang Wu},
  booktitle={Proceedings of the International Joint Conference on Artificial Intelligence},
  year={2021},
  pages={1179--1185},
}

@article{wu2023semantic,
  title={Semantic-aware graph matching mechanism for multi-label image recognition},
  author={Wu, Yanan and Feng, Songhe and Wang, Yang},
  journal={IEEE Transactions on Circuits and Systems for Video Technology},
  volume={33},
  number={11},
  pages={6788--6803},
  year={2023},
  publisher={IEEE}
}

@inproceedings{wang2016cnn,
  title={Cnn-rnn: A unified framework for multi-label image classification},
  author={Wang, Jiang and Yang, Yi and Mao, Junhua and Huang, Zhiheng and Huang, Chang and Xu, Wei},
  booktitle={Proceedings of the IEEE/CVF conference on computer vision and pattern recognition},
  pages={2285--2294},
  year={2016}
}

@article{shi2017training,
  title={Training DCNN by combining max-margin, max-correlation objectives, and correntropy loss for multilabel image classification},
  author={Shi, Weiwei and Gong, Yihong and Tao, Xiaoyu and Zheng, Nanning},
  journal={IEEE transactions on neural networks and learning systems},
  volume={29},
  number={7},
  pages={2896--2908},
  year={2017},
  publisher={IEEE}
}

@article{lyu2019attend,
  title={Attend and imagine: Multi-label image classification with visual attention and recurrent neural networks},
  author={Lyu, Fan and Wu, Qi and Hu, Fuyuan and Wu, Qingyao and Tan, Mingkui},
  journal={IEEE Transactions on Multimedia},
  volume={21},
  number={8},
  pages={1971--1981},
  year={2019},
  publisher={IEEE}
}

@inproceedings{chen2018order,
  title={Order-free rnn with visual attention for multi-label classification},
  author={Chen, Shang-Fu and Chen, Yi-Chen and Yeh, Chih-Kuan and Wang, Yu-Chiang},
  booktitle={Proceedings of the AAAI conference on artificial intelligence},
  volume={32},
  number={1},
  pages={6714--6721},
  year={2018}
}

@inproceedings{yazici2020orderless,
  title={Orderless recurrent models for multi-label classification},
  author={Yazici, Vacit Oguz and Gonzalez-Garcia, Abel and Ramisa, Arnau and Twardowski, Bartlomiej and Weijer, Joost van de},
  booktitle={Proceedings of the IEEE/CVF Conference on Computer Vision and Pattern Recognition},
  pages={13440--13449},
  year={2020}
}

@inproceedings{zhu2017learning,
  title={Learning spatial regularization with image-level supervisions for multi-label image classification},
  author={Zhu, Feng and Li, Hongsheng and Ouyang, Wanli and Yu, Nenghai and Wang, Xiaogang},
  booktitle={Proceedings of the IEEE/CVF conference on computer vision and pattern recognition},
  pages={5513--5522},
  year={2017}
}

@inproceedings{chen2019multi,
  title={Multi-label image recognition with graph convolutional networks},
  author={Chen, Zhao-Min and Wei, Xiu-Shen and Wang, Peng and Guo, Yanwen},
  booktitle={Proceedings of the IEEE/CVF conference on computer vision and pattern recognition},
  pages={5177--5186},
  year={2019}
}

@inproceedings{you2020cross,
  title={Cross-modality attention with semantic graph embedding for multi-label classification},
  author={You, Renchun and Guo, Zhiyao and Cui, Lei and Long, Xiang and Bao, Yingze and Wen, Shilei},
  booktitle={Proceedings of the AAAI conference on artificial intelligence},
  volume={34},
  number={07},
  pages={12709--12716},
  year={2020}
}

@inproceedings{wu2020adahgnn,
  title={AdaHGNN: Adaptive hypergraph neural networks for multi-label image classification},
  author={Wu, Xiangping and Chen, Qingcai and Li, Wei and Xiao, Yulun and Hu, Baotian},
  booktitle={Proceedings of the 28th ACM international conference on multimedia},
  pages={284--293},
  year={2020}
}

@article{xu2020joint,
  title={Joint input and output space learning for multi-label image classification},
  author={Xu, Jiahao and Tian, Hongda and Wang, Zhiyong and Wang, Yang and Kang, Wenxiong and Chen, Fang},
  journal={IEEE Transactions on Multimedia},
  volume={23},
  pages={1696--1707},
  year={2020},
  publisher={IEEE}
}

@inproceedings{li2016conditional,
  title={Conditional graphical lasso for multi-label image classification},
  author={Li, Qiang and Qiao, Maoying and Bian, Wei and Tao, Dacheng},
  booktitle={Proceedings of the IEEE/CVF conference on computer vision and pattern recognition},
  pages={2977--2986},
  year={2016}
}

@inproceedings{marino2017more,
  title={The More You Know: Using Knowledge Graphs for Image Classification},
  author={Marino, Kenneth and Salakhutdinov, Ruslan and Gupta, Abhinav},
  booktitle={Proceedings of the IEEE/CVF conference on computer vision and pattern recognition},
  pages={20--28},
  year={2017},
  organization={IEEE Computer Society}
}

@inproceedings{chen2019learning,
  title={Learning semantic-specific graph representation for multi-label image recognition},
  author={Chen, Tianshui and Xu, Muxin and Hui, Xiaolu and Wu, Hefeng and Lin, Liang},
  booktitle={Proceedings of the IEEE/CVF International Conference on Computer Vision},
  pages={522--531},
  year={2019}
}

@inproceedings{ye2020attention,
  title={Attention-driven dynamic graph convolutional network for multi-label image recognition},
  author={Ye, Jin and He, Junjun and Peng, Xiaojiang and Wu, Wenhao and Qiao, Yu},
  booktitle={European Conference on Computer Vision},
  pages={649--665},
  year={2020},
  organization={Springer}
}

@inproceedings{cheng2022mltr,
  title={Mltr: Multi-label classification with transformer},
  author={Cheng, Xing and Lin, Hezheng and Wu, Xiangyu and Shen, Dong and Yang, Fan and Liu, Honglin and Shi, Nian},
  booktitle={2022 IEEE International Conference on Multimedia and Expo (ICME)},
  pages={1--6},
  year={2022},
  organization={IEEE}
}

@inproceedings{lanchantin2021general,
  title={General Multi-label Image Classification with Transformers},
  author={Lanchantin, Jack and Wang, Tianlu and Ordonez, Vicente and Qi, Yanjun},
  booktitle={Proceedings of the IEEE/CVF Conference on Computer Vision and Pattern Recognition},
  pages={16478--16488},
  year={2021}
}

@article{liu2021query2label,
  title={Query2label: A simple transformer way to multi-label classification},
  author={Liu, Shilong and Zhang, Lei and Yang, Xiao and Su, Hang and Zhu, Jun},
  journal={arXiv preprint arXiv:2107.10834},
  year={2021}
}

@inproceedings{zhao2021transformer,
  title={Transformer-based dual relation graph for multi-label image recognition},
  author={Zhao, Jiawei and Yan, Ke and Zhao, Yifan and Guo, Xiaowei and Huang, Feiyue and Li, Jia},
  booktitle={Proceedings of the IEEE/CVF international conference on computer vision},
  pages={163--172},
  year={2021}
}

@inproceedings{zhao2021m3tr,
  title={M3TR: Multi-modal Multi-label Recognition with Transformer},
  author={Zhao, Jiawei and Zhao, Yifan and Li, Jia},
  booktitle={Proceedings of the 29th ACM International Conference on Multimedia},
  pages={469--477},
  year={2021}
}

@article{chen2022sst,
  title={SST: Spatial and semantic transformers for multi-label image recognition},
  author={Chen, Zhao-Min and Cui, Quan and Zhao, Borui and Song, Renjie and Zhang, Xiaoqin and Yoshie, Osamu},
  journal={IEEE Transactions on Image Processing},
  volume={31},
  pages={2570--2583},
  year={2022},
  publisher={IEEE}
}

@article{zhou2023datran,
  title={DATran: Dual Attention Transformer for Multi-Label Image Classification},
  author={Zhou, Wei and Zheng, Zhijie and Su, Tao and Hu, Haifeng},
  journal={IEEE Transactions on Circuits and Systems for Video Technology},
  volume={34},
  number={1},
  pages={342--356},
  year={2023},
  publisher={IEEE}
}

@article{wu2023transformer,
  title={Transformer driven matching selection mechanism for multi-label image classification},
  author={Wu, Yanan and Feng, Songhe and Zhao, Gongpei and Jin, Yi},
  journal={IEEE Transactions on Circuits and Systems for Video Technology},
  volume={34},
  number={2},
  pages={924--937},
  year={2023},
  publisher={IEEE}
}

@inproceedings{ouyang2023hsvlt,
  title={HSVLT: Hierarchical Scale-Aware Vision-Language Transformer for Multi-Label Image Classification},
  author={Ouyang, Shuyi and Wang, Hongyi and Niu, Ziwei and Bai, Zhenjia and Xie, Shiao and Xu, Yingying and Tong, Ruofeng and Chen, Yen-Wei and Lin, Lanfen},
  booktitle={Proceedings of the 31st ACM International Conference on Multimedia},
  pages={4768--4777},
  year={2023}
}

@inproceedings{li2024pyramidal,
  title={Pyramidal Cross-Modal Transformer with Sustained Visual Guidance for Multi-Label Image Classification},
  author={Li, Zhuohua and Wang, Ruyun and Zhu, Fuqing and Han, Jizhong and Hu, Songlin},
  booktitle={Proceedings of the International Conference on Multimedia Retrieval},
  pages={740--748},
  year={2024}
}

@inproceedings{guo2019visual,
  title={Visual attention consistency under image transforms for multi-label image classification},
  author={Guo, Hao and Zheng, Kang and Fan, Xiaochuan and Yu, Hongkai and Wang, Song},
  booktitle={Proceedings of the IEEE/CVF conference on computer vision and pattern recognition},
  pages={729--739},
  year={2019}
}

@inproceedings{ridnik2021asymmetric,
  title={Asymmetric loss for multi-label classification},
  author={Ridnik, Tal and Ben-Baruch, Emanuel and Zamir, Nadav and Noy, Asaf and Friedman, Itamar and Protter, Matan and Zelnik-Manor, Lihi},
  booktitle={Proceedings of the IEEE/CVF International Conference on Computer Vision},
  pages={82--91},
  year={2021}
}

@inproceedings{zhu2021residual,
  title={Residual attention: A simple but effective method for multi-label recognition},
  author={Zhu, Ke and Wu, Jianxin},
  booktitle={Proceedings of the IEEE/CVF International Conference on Computer Vision},
  pages={184--193},
  year={2021}
}

@article{dao2021multi,
  title={Multi-label image classification with contrastive learning},
  author={Dao, Son D and Zhao, Ethan and Phung, Dinh and Cai, Jianfei},
  journal={arXiv preprint arXiv:2107.11626},
  year={2021}
}

@inproceedings{ma2023semantic,
  title={Semantic-Aware Dual Contrastive Learning for Multi-label Image Classification},
  author={Ma, Leilei and Sun, Dengdi and Wang, Lei and Zhao, Haifeng and Luo, Bin},
  booktitle={European Conference on Artificial Intelligence},
  pages={1656--1663},
  year={2023},
}

@inproceedings{liu2022contextual,
  title={Contextual debiasing for visual recognition with causal mechanisms},
  author={Liu, Ruyang and Liu, Hao and Li, Ge and Hou, Haodi and Yu, TingHao and Yang, Tao},
  booktitle={Proceedings of the IEEE/CVF Conference on Computer Vision and Pattern Recognition},
  pages={12755--12765},
  year={2022}
}

@article{tian2023causal,
  title={Causal multi-label learning for image classification},
  author={Tian, Yingjie and Bai, Kunlong and Yu, Xiaotong and Zhu, Siyu},
  journal={Neural Networks},
  volume={167},
  pages={626--637},
  year={2023},
  publisher={Elsevier}
}

@inproceedings{liu2023causality,
  title={Causality compensated attention for contextual biased visual recognition},
  author={Liu, Ruyang and Huang, Jingjia and Li, Thomas H and Li, Ge},
  booktitle={The eleventh international conference on learning representations},
  year={2023}
}

@article{xie2024counterfactual,
  title={Counterfactual Reasoning for Multi-Label Image Classification via Patching-Based Training},
  author={Xie, Ming-Kun and Xiao, Jia-Hao and Peng, Pei and Niu, Gang and Sugiyama, Masashi and Huang, Sheng-Jun},
  journal={arXiv preprint arXiv:2404.06287},
  year={2024}
}

@inproceedings{yin2024hybrid,
  title={Hybrid Sharing for Multi-Label Image Classification},
  author={Yin, Zihao and Gan, Chen and He, Kelei and Gao, Yang and Zhang, Junfeng},
  booktitle={The Twelfth International Conference on Learning Representations},
  year={2024}
}

@article{zhu2024semantic,
  title={Semantic-Guided Representation Enhancement for Multi-Label Image Classification}, 
  author={Zhu, Xuelin and Li, Jianshu and Cao, Jiuxin and Tang, Dongqi and Liu, Jian and Liu, Bo},
  journal={IEEE Transactions on Circuits and Systems for Video Technology}, 
  year={2024},
  volume={34},
  number={10},
  pages={10036-10049}
}

@article{yuan2023balanced,
  title={Balanced masking strategy for multi-label image classification},
  author={Yuan, Jin and Zhang, Yao and Shi, Zhongchao and Geng, Xin and Fan, Jianping and Rui, Yong},
  journal={Neurocomputing},
  volume={522},
  pages={64--72},
  year={2023},
  publisher={Elsevier}
}

@article{zhou2023mining,
  title={Mining semantic information with dual relation graph network for multi-label image classification},
  author={Zhou, Wei and Jiang, Weitao and Chen, Dihu and Hu, Haifeng and Su, Tao},
  journal={IEEE Transactions on Multimedia},
  volume={26},
  pages={1143--1157},
  year={2023},
  publisher={IEEE}
}

@inproceedings{li2023patchct,
  title={Patchct: Aligning patch set and label set with conditional transport for multi-label image classification},
  author={Li, Miaoge and Wang, Dongsheng and Liu, Xinyang and Zeng, Zequn and Lu, Ruiying and Chen, Bo and Zhou, Mingyuan},
  booktitle={Proceedings of the IEEE/CVF International Conference on Computer Vision},
  pages={15348--15358},
  year={2023}
}

@inproceedings{zhu2023scene,
  title={Scene-aware label graph learning for multi-label image classification},
  author={Zhu, Xuelin and Liu, Jian and Liu, Weijia and Ge, Jiawei and Liu, Bo and Cao, Jiuxin},
  booktitle={Proceedings of the IEEE/CVF International Conference on Computer Vision},
  pages={1473--1482},
  year={2023}
}

@article{zhang2023spatial,
  title={Spatial context-aware object-attentional network for multi-label image classification},
  author={Zhang, Jialu and Ren, Jianfeng and Zhang, Qian and Liu, Jiang and Jiang, Xudong},
  journal={IEEE Transactions on Image Processing},
  volume={32},
  pages={3000--3012},
  year={2023},
  publisher={IEEE}
}

@article{deng2022beyond,
  title={Beyond word embeddings: Heterogeneous prior knowledge driven multi-label image classification},
  author={Deng, Xiang and Feng, Songhe and Lyu, Gengyu and Wang, Tao and Lang, Congyan},
  journal={IEEE Transactions on Multimedia},
  volume={25},
  pages={4013--4025},
  year={2022},
  publisher={IEEE}
}

@inproceedings{xu2022boosting,
  author={Jiazhi Xu and Sheng Huang and Fengtao Zhou and Luwen Huangfu and Daniel Zeng and Bo Liu},
  title={Boosting Multi-Label Image Classification with Complementary Parallel Self-Distillation},
  booktitle={Proceedings of the Thirty-First International Joint Conference on Artificial Intelligence},
  pages={1495--1501},
  year={2022},
  publisher={ijcai.org}
}

@article{zhou2023double,
  title={Double attention based on graph attention network for image multi-label classification},
  author={Zhou, Wei and Xia, Zhiwu and Dou, Peng and Su, Tao and Hu, Haifeng},
  journal={ACM Transactions on Multimedia Computing, Communications and Applications},
  volume={19},
  number={1},
  pages={1--23},
  year={2023},
  publisher={ACM New York, NY}
}

@inproceedings{zhan2022global,
  title={Global Meets Local: Effective Multi-Label Image Classification via Category-Aware Weak Supervision},
  author={Zhan, Jiawei and Liu, Jun and Tang, Wei and Jiang, Guannan and Wang, Xi and Gao, Bin-Bin and Zhang, Tianliang and Wu, Wenlong and Zhang, Wei and Wang, Chengjie and others},
  booktitle={Proceedings of the 30th ACM International Conference on Multimedia},
  pages={6318--6326},
  year={2022}
}

@inproceedings{huang2024tag2text,
  title={Tag2text: Guiding vision-language model via image tagging},
  author={Huang, Xinyu and Zhang, Youcai and Ma, Jinyu and Tian, Weiwei and Feng, Rui and Zhang, Yuejie and Li, Yaqian and Guo, Yandong and Zhang, Lei},
  booktitle={The twelfth international conference on learning representations},
  year={2024}
}

@article{chen2022label,
  title={Label-aware graph representation learning for multi-label image classification},
  author={Chen, Yilu and Zou, Changzhong and Chen, Jianli},
  journal={Neurocomputing},
  volume={492},
  pages={50--61},
  year={2022},
  publisher={Elsevier}
}

@inproceedings{zhu2023multi,
  title={Multi-label self-supervised learning with scene images},
  author={Zhu, Ke and Fu, Minghao and Wu, Jianxin},
  booktitle={Proceedings of the IEEE/CVF International Conference on Computer Vision},
  pages={6694--6703},
  year={2023}
}

@article{zhou2021multi,
  title={Multi-label image classification via category prototype compositional learning},
  author={Zhou, Fengtao and Huang, Sheng and Liu, Bo and Yang, Dan},
  journal={IEEE Transactions on Circuits and Systems for Video Technology},
  volume={32},
  number={7},
  pages={4513--4525},
  year={2021},
  publisher={IEEE}
}

@article{pu2023semantic,
  title={Semantic representation and dependency learning for multi-label image recognition},
  author={Pu, Tao and Sun, Mingzhan and Wu, Hefeng and Chen, Tianshui and Tian, Ling and Lin, Liang},
  journal={Neurocomputing},
  volume={526},
  pages={121--130},
  year={2023},
  publisher={Elsevier}
}

@article{wang2021semantic,
  title={Semantic supplementary network with prior information for multi-label image classification},
  author={Wang, Zhe and Fang, Zhongli and Li, Dongdong and Yang, Hai and Du, Wenli},
  journal={IEEE Transactions on Circuits and Systems for Video Technology},
  volume={32},
  number={4},
  pages={1848--1859},
  year={2021},
  publisher={IEEE}
}

@inproceedings{zhu2022two,
  title={Two-stream transformer for multi-label image classification},
  author={Zhu, Xuelin and Cao, Jiuxin and Ge, Jiawei and Liu, Weijia and Liu, Bo},
  booktitle={Proceedings of the 30th ACM International Conference on Multimedia},
  pages={3598--3607},
  year={2022}
}

@inproceedings{song2021handling,
  title={Handling difficult labels for multi-label image classification via uncertainty distillation},
  author={Song, Liangchen and Wu, Jialian and Yang, Ming and Zhang, Qian and Li, Yuan and Yuan, Junsong},
  booktitle={Proceedings of the 29th ACM International Conference on Multimedia},
  pages={2410--2419},
  year={2021}
}

@article{chen2021learning,
  title={Learning graph convolutional networks for multi-label recognition and applications},
  author={Chen, Zhao-Min and Wei, Xiu-Shen and Wang, Peng and Guo, Yanwen},
  journal={IEEE Transactions on Pattern Analysis and Machine Intelligence},
  volume={45},
  number={6},
  pages={6969--6983},
  year={2021},
  publisher={IEEE}
}

@article{gao2021learning,
  title={Learning to discover multi-class attentional regions for multi-label image recognition},
  author={Gao, Bin-Bin and Zhou, Hong-Yu},
  journal={IEEE Transactions on Image Processing},
  volume={30},
  pages={5920--5932},
  year={2021},
  publisher={IEEE}
}

@inproceedings{ridnik2023ml,
  title={Ml-decoder: Scalable and versatile classification head},
  author={Ridnik, Tal and Sharir, Gilad and Ben-Cohen, Avi and Ben-Baruch, Emanuel and Noy, Asaf},
  booktitle={Proceedings of the IEEE/CVF winter conference on applications of computer vision},
  pages={32--41},
  year={2023}
}

@inproceedings{zhang2022spatial,
  title={Spatial-context-aware deep neural network for multi-class image classification},
  author={Zhang, Jialu and Zhang, Qian and Ren, Jianfeng and Zhao, Yitian and Liu, Jiang},
  booktitle={ICASSP 2022-2022 IEEE International Conference on Acoustics, Speech and Signal Processing (ICASSP)},
  pages={1960--1964},
  year={2022},
  organization={IEEE}
}

@inproceedings{wu2020distribution,
  title={Distribution-balanced loss for multi-label classification in long-tailed datasets},
  author={Wu, Tong and Huang, Qingqiu and Liu, Ziwei and Wang, Yu and Lin, Dahua},
  booktitle={Computer Vision--ECCV 2020: 16th European Conference, Glasgow, UK, August 23--28, 2020, Proceedings, Part IV 16},
  pages={162--178},
  year={2020},
  organization={Springer}
}

@article{chen2020knowledge,
  title={Knowledge-guided multi-label few-shot learning for general image recognition},
  author={Chen, Tianshui and Lin, Liang and Chen, Riquan and Hui, Xiaolu and Wu, Hefeng},
  journal={IEEE Transactions on Pattern Analysis and Machine Intelligence},
  volume={44},
  number={3},
  pages={1371--1384},
  year={2020},
  publisher={IEEE}
}

@inproceedings{wang2020multi,
  title={Multi-label classification with label graph superimposing},
  author={Wang, Ya and He, Dongliang and Li, Fu and Long, Xiang and Zhou, Zhichao and Ma, Jinwen and Wen, Shilei},
  booktitle={Proceedings of the AAAI Conference on Artificial Intelligence},
  volume={34},
  number={07},
  pages={12265--12272},
  year={2020}
}

@article{yeh2019multilabel,
  title={Multilabel deep visual-semantic embedding},
  author={Yeh, Mei-Chen and Li, Yi-Nan},
  journal={IEEE transactions on pattern analysis and machine intelligence},
  volume={42},
  number={6},
  pages={1530--1536},
  year={2019},
  publisher={IEEE}
}

@article{li2019reconstruction,
  title={Reconstruction regularized deep metric learning for multi-label image classification},
  author={Li, Changsheng and Liu, Chong and Duan, Lixin and Gao, Peng and Zheng, Kai},
  journal={IEEE transactions on neural networks and learning systems},
  volume={31},
  number={7},
  pages={2294--2303},
  year={2019},
  publisher={IEEE}
}

@article{chen2020disentangling,
  title={Disentangling, embedding and ranking label cues for multi-label image recognition},
  author={Chen, Zhao-Min and Cui, Quan and Wei, Xiu-Shen and Jin, Xin and Guo, Yanwen},
  journal={IEEE Transactions on Multimedia},
  volume={23},
  pages={1827--1840},
  year={2020},
  publisher={IEEE}
}

@inproceedings{yan2019imbalance,
  title={Imbalance Rectification in Deep Logistic Regression for Multi-Label Image Classification Using Random Noise Samples},
  author={Yan, Wenjin and Li, Ruixuan and Wang, Jun and Li, Yuhua and Wang, Jinyang and Zhou, Pan and Gu, Xiwu},
  booktitle={Proceedings of the 28th ACM International Conference on Information and Knowledge Management},
  pages={1131--1140},
  year={2019}
}

@article{markatopoulou2018implicit,
  title={Implicit and explicit concept relations in deep neural networks for multi-label video/image annotation},
  author={Markatopoulou, Foteini and Mezaris, Vasileios and Patras, Ioannis},
  journal={IEEE transactions on circuits and systems for video technology},
  volume={29},
  number={6},
  pages={1631--1644},
  year={2018},
  publisher={IEEE}
}

@article{song2018deep,
  title={A deep multi-modal CNN for multi-instance multi-label image classification},
  author={Song, Lingyun and Liu, Jun and Qian, Buyue and Sun, Mingxuan and Yang, Kuan and Sun, Meng and Abbas, Samar},
  journal={IEEE Transactions on Image Processing},
  volume={27},
  number={12},
  pages={6025--6038},
  year={2018},
  publisher={IEEE}
}

@article{wei2015hcp,
  title={HCP: A flexible CNN framework for multi-label image classification},
  author={Wei, Yunchao and Xia, Wei and Lin, Min and Huang, Junshi and Ni, Bingbing and Dong, Jian and Zhao, Yao and Yan, Shuicheng},
  journal={IEEE transactions on pattern analysis and machine intelligence},
  volume={38},
  number={9},
  pages={1901--1907},
  year={2015},
  publisher={IEEE}
}

@article{gong2013deep,
  title={Deep convolutional ranking for multilabel image annotation},
  author={Gong, Yunchao and Jia, Yangqing and Leung, Thomas and Toshev, Alexander and Ioffe, Sergey},
  journal={arXiv preprint arXiv:1312.4894},
  year={2013}
}

@inproceedings{wang2019baseline,
  title={A baseline for multi-label image classification using an ensemble of deep convolutional neural networks},
  author={Wang, Qian and Jia, Ning and Breckon, Toby P},
  booktitle={2019 IEEE International Conference on Image Processing (ICIP)},
  pages={644--648},
  year={2019},
  organization={IEEE}
}

@inproceedings{zhang2019multi,
  title={Multi-view metric learning for multi-label image classification},
  author={Zhang, Mengying and Li, Changsheng and Wang, Xiangfeng},
  booktitle={2019 IEEE International Conference on Image Processing (ICIP)},
  pages={2134--2138},
  year={2019},
  organization={IEEE}
}

@inproceedings{yin2024tfad,
  title={TFAD: An Image Multi-Label Recognition Method with Image-Text Powered Attention},
  author={Yin, Haoran},
  booktitle={2024 International Joint Conference on Neural Networks (IJCNN)},
  pages={1--8},
  year={2024},
  organization={IEEE}
}

@article{wang2021instance,
  title={Instance-aware deep graph learning for multi-label classification},
  author={Wang, Yun and Zhang, Tong and Zhou, Chuanwei and Cui, Zhen and Yang, Jian},
  journal={IEEE Transactions on Multimedia},
  volume={25},
  pages={90--99},
  year={2021},
  publisher={IEEE}
}

@inproceedings{li2017improving,
  title={Improving pairwise ranking for multi-label image classification},
  author={Li, Yuncheng and Song, Yale and Luo, Jiebo},
  booktitle={Proceedings of the IEEE conference on computer vision and pattern recognition},
  pages={3617--3625},
  year={2017}
}

@article{coulibaly2022deep,
  title={Deep Convolution Neural Network sharing for the multi-label images classification},
  author={Coulibaly, Solemane and Kamsu-Foguem, Bernard and Kamissoko, Dantouma and Traore, Daouda},
  journal={Machine learning with applications},
  volume={10},
  pages={100422},
  year={2022},
  publisher={Elsevier}
}

@inproceedings{zhou2021deep,
  title={Deep semantic dictionary learning for multi-label image classification},
  author={Zhou, Fengtao and Huang, Sheng and Xing, Yun},
  booktitle={Proceedings of the AAAI conference on artificial intelligence},
  volume={35},
  number={4},
  pages={3572--3580},
  year={2021}
}

@inproceedings{liu2018multi,
  title={Multi-label image classification via knowledge distillation from weakly-supervised detection},
  author={Liu, Yongcheng and Sheng, Lu and Shao, Jing and Yan, Junjie and Xiang, Shiming and Pan, Chunhong},
  booktitle={Proceedings of the 26th ACM international conference on Multimedia},
  pages={700--708},
  year={2018}
}

@inproceedings{he2018reinforced,
  title={Reinforced multi-label image classification by exploring curriculum},
  author={He, Shiyi and Xu, Chang and Guo, Tianyu and Xu, Chao and Tao, Dacheng},
  booktitle={Proceedings of the AAAI conference on artificial intelligence},
  volume={32},
  number={1},
  year={2018}
}

@inproceedings{luo2019visual,
  title={Visual attention in multi-label image classification},
  author={Luo, Yan and Jiang, Ming and Zhao, Qi},
  booktitle={Proceedings of the IEEE/CVF Conference on Computer Vision and Pattern Recognition Workshops},
  pages={0--0},
  year={2019}
}

@article{zhou2023feature,
  title={Feature learning network with transformer for multi-label image classification},
  author={Zhou, Wei and Dou, Peng and Su, Tao and Hu, Haifeng and Zheng, Zhijie},
  journal={Pattern Recognition},
  volume={136},
  pages={109203},
  year={2023},
  publisher={Elsevier}
}

@article{qu2023multi,
  title={Multi-layered semantic representation network for multi-label image classification},
  author={Qu, Xiwen and Che, Hao and Huang, Jun and Xu, Linchuan and Zheng, Xiao},
  journal={International Journal of Machine Learning and Cybernetics},
  volume={14},
  number={10},
  pages={3427--3435},
  year={2023},
  publisher={Springer}
}

@article{yu2019delta,
  title={DELTA: A deep dual-stream network for multi-label image classification},
  author={Yu, Wan-Jin and Chen, Zhen-Duo and Luo, Xin and Liu, Wu and Xu, Xin-Shun},
  journal={Pattern Recognition},
  volume={91},
  pages={322--331},
  year={2019},
  publisher={Elsevier}
}

@article{yuan2023graph,
  title={Graph attention transformer network for multi-label image classification},
  author={Yuan, Jin and Chen, Shikai and Zhang, Yao and Shi, Zhongchao and Geng, Xin and Fan, Jianping and Rui, Yong},
  journal={ACM Transactions on Multimedia Computing, Communications and Applications},
  volume={19},
  number={4},
  pages={1--16},
  year={2023},
  publisher={ACM New York, NY}
}

@article{dao2023contrastively,
  title={Contrastively enforcing distinctiveness for multi-label image classification},
  author={Dao, Son D and Zhao, He and Phung, Dinh and Cai, Jianfei},
  journal={Neurocomputing},
  volume={555},
  pages={126605},
  year={2023},
  publisher={Elsevier}
}

@inproceedings{nguyen2021modular,
  title={Modular graph transformer networks for multi-label image classification},
  author={Nguyen, Hoang D and Vu, Xuan-Son and Le, Duc-Trong},
  booktitle={Proceedings of the AAAI conference on artificial intelligence},
  volume={35},
  number={10},
  pages={9092--9100},
  year={2021}
}

@article{yu2017combining,
  title={Combining local and global hypotheses in deep neural network for multi-label image classification},
  author={Yu, Qinghua and Wang, Jinjun and Zhang, Shizhou and Gong, Yihong and Zhao, Jizhong},
  journal={Neurocomputing},
  volume={235},
  pages={38--45},
  year={2017},
  publisher={Elsevier}
}

@article{singh2024multi,
  title={Multi-label image classification using adaptive graph convolutional networks: from a single domain to multiple domains},
  author={Singh, Inder Pal and Ghorbel, Enjie and Oyedotun, Oyebade and Aouada, Djamila},
  journal={Computer Vision and Image Understanding},
  volume={247},
  pages={104062},
  year={2024},
  publisher={Elsevier}
}

@inproceedings{lin2023probability,
  title={Probability guided loss for long-tailed multi-label image classification},
  author={Lin, Dekun},
  booktitle={Proceedings of the AAAI Conference on Artificial Intelligence},
  volume={37},
  number={2},
  pages={1577--1585},
  year={2023}
}

@inproceedings{Zhao2016RegionalGN,
  title={Regional Gating Neural Networks for Multi-label Image Classification},
  author={Rui-Wei Zhao and Jianguo Li and Yurong Chen and Jia-Ming Liu and Yu-Gang Jiang and X. Xue},
  booktitle={British Machine Vision Conference},
  year={2016},
  url={https://api.semanticscholar.org/CorpusID:8388788}
}

@article{zhou2023aligning,
  title={Aligning image semantics and label concepts for image multi-label classification},
  author={Zhou, Wei and Xia, Zhiwu and Dou, Peng and Su, Tao and Hu, Haifeng},
  journal={ACM Transactions on Multimedia Computing, Communications and Applications},
  volume={19},
  number={2},
  pages={1--23},
  year={2023},
  publisher={ACM New York, NY}
}

@article{jing2016multi,
  title={Multi-label dictionary learning for image annotation},
  author={Jing, Xiao-Yuan and Wu, Fei and Li, Zhiqiang and Hu, Ruimin and Zhang, David},
  journal={IEEE Transactions on Image Processing},
  volume={25},
  number={6},
  pages={2712--2725},
  year={2016},
  publisher={IEEE}
}

@article{wang2022hierarchical,
  title={Hierarchical GAN-Tree and Bi-Directional Capsules for multi-label image classification},
  author={Wang, Boyan and Hu, Xuegang and Zhang, Chenwei and Li, Peipei and Philip, S Yu},
  journal={Knowledge-Based Systems},
  volume={238},
  pages={107882},
  year={2022},
  publisher={Elsevier}
}

@inproceedings{kuang2023multi,
  title={Multi-label Image Classification with Multi-scale Global-Local Semantic Graph Network},
  author={Kuang, Wenlan and Zhu, Qiangxi and Li, Zhixin},
  booktitle={Joint European Conference on Machine Learning and Knowledge Discovery in Databases},
  pages={53--69},
  year={2023},
  organization={Springer}
}

@article{dai2023global,
  title={Global-guided weakly-supervised learning for multi-label image classification},
  author={Dai, Yong and Song, Weiwei and Gao, Zhi and Fang, Leyuan},
  journal={Journal of Visual Communication and Image Representation},
  volume={93},
  pages={103823},
  year={2023},
  publisher={Elsevier}
}

@inproceedings{singh2022multi,
  title={Multi label image classification using adaptive graph convolutional networks (ml-agcn)},
  author={Singh, Inder Pal and Ghorbel, Enjie and Oyedotun, Oyebade and Aouada, Djamila},
  booktitle={2022 IEEE International Conference on Image Processing (ICIP)},
  pages={1806--1810},
  year={2022},
  organization={IEEE}
}

@inproceedings{wang2023mumic,
  title={Mumic--multimodal embedding for multi-label image classification with tempered sigmoid},
  author={Wang, Fengjun and Mizrachi, Sarai and Beladev, Moran and Nadav, Guy and Amsalem, Gil and Assaraf, Karen Lastmann and Boker, Hadas Harush},
  booktitle={Proceedings of the AAAI Conference on Artificial Intelligence},
  volume={37},
  number={13},
  pages={15603--15611},
  year={2023}
}

@inproceedings{hu2023atom,
  title={ATOM: Automated Black-Box Testing of Multi-Label Image Classification Systems},
  author={Hu, Shengyou and Wu, Huayao and Wang, Peng and Chang, Jing and Tu, Yongjun and Jiang, Xiu and Niu, Xintao and Nie, Changhai},
  booktitle={2023 38th IEEE/ACM International Conference on Automated Software Engineering (ASE)},
  pages={230--242},
  year={2023},
  organization={IEEE}
}

@inproceedings{cheng2018multi,
  title={Multi-scale and discriminative part detectors based features for multi-label image classification.},
  author={Cheng, Gong and Gao, Decheng and Liu, Yang and Han, Junwei},
  booktitle={IJCAI},
  pages={649--655},
  year={2018}
}

@article{wang2024semantic,
  title={A semantic guidance-based fusion network for multi-label image classification},
  author={Wang, Jiuhang and Tang, Hongying and Luo, Shanshan and Yang, Liqi and Liu, Shusheng and Hong, Aoping and Li, Baoqing},
  journal={Pattern Recognition Letters},
  volume={185},
  pages={254--261},
  year={2024},
  publisher={Elsevier}
}

@inproceedings{liu2019decoupling,
  title={Decoupling category-wise independence and relevance with self-attention for multi-label image classification},
  author={Liu, Luchen and Guo, Sheng and Huang, Weilin and Scott, Matthew R},
  booktitle={ICASSP 2019-2019 IEEE International Conference on Acoustics, Speech and Signal Processing (ICASSP)},
  pages={1682--1686},
  year={2019},
  organization={IEEE}
}

@inproceedings{lin2025distributionally,
  title={Distributionally Robust Loss for Long-Tailed Multi-label Image Classification},
  author={Lin, Dekun and Peng, Tailai and Chen, Rui and Xie, Xinran and Qin, Xiaolin and Cui, Zhe},
  booktitle={European Conference on Computer Vision},
  pages={417--433},
  year={2025},
  organization={Springer}
}

@article{kuang2024multi,
  title={Multi-label image classification with multi-layered multi-perspective dynamic semantic representation},
  author={Kuang, Wenlan and Li, Zhixin},
  journal={Machine Learning},
  volume={113},
  number={6},
  pages={3443--3461},
  year={2024},
  publisher={Springer}
}

@article{ma2022deep,
  title={Deep Large-Margin Rank Loss for Multi-Label Image Classification},
  author={Ma, Zhongchen and Li, Zongpeng and Zhan, Yongzhao},
  journal={Mathematics},
  volume={10},
  number={23},
  pages={4584},
  year={2022},
  publisher={MDPI}
}

@inproceedings{sajedi2024probmcl,
  title={Probmcl: Simple probabilistic contrastive learning for multi-label visual classification},
  author={Sajedi, Ahmad and Khaki, Samir and Lawryshyn, Yuri A and Plataniotis, Konstantinos N},
  booktitle={ICASSP 2024-2024 IEEE International Conference on Acoustics, Speech and Signal Processing (ICASSP)},
  pages={5115--5119},
  year={2024},
  organization={IEEE}
}

@article{hu2024dual,
  title={Dual-stream multi-label image classification model enhanced by feature reconstruction},
  author={Hu, Liming and Chen, Mingxuan and Wang, Anjie and Fang, Zhijun},
  journal={Multimedia Systems},
  volume={30},
  number={5},
  pages={281},
  year={2024},
  publisher={Springer}
}

@inproceedings{zhang2024recognize,
    author    = {Zhang, Youcai and Huang, Xinyu and Ma, Jinyu and Li, Zhaoyang and Luo, Zhaochuan and Xie, Yanchun and Qin, Yuzhuo and Luo, Tong and Li, Yaqian and Liu, Shilong and Guo, Yandong and Zhang, Lei},
    title     = {Recognize Anything: A Strong Image Tagging Model},
    booktitle = {Proceedings of the IEEE/CVF Conference on Computer Vision and Pattern Recognition (CVPR) Workshops},
    month     = {June},
    year      = {2024},
    pages     = {1724-1732}
}

@article{yuan2024open,
  title={Open-vocabulary SAM: Segment and recognize twenty-thousand classes interactively},
  author={Yuan, Haobo and Li, Xiangtai and Zhou, Chong and Li, Yining and Chen, Kai and Loy, Chen Change},
  journal={arXiv preprint arXiv:2401.02955},
  year={2024}
}

@article{li2023semantic,
  title={Semantic-sam: Segment and recognize anything at any granularity},
  author={Li, Feng and Zhang, Hao and Sun, Peize and Zou, Xueyan and Liu, Shilong and Yang, Jianwei and Li, Chunyuan and Zhang, Lei and Gao, Jianfeng},
  journal={arXiv preprint arXiv:2307.04767},
  year={2023}
}

@article{wang2023all,
  title={The all-seeing project: Towards panoptic visual recognition and understanding of the open world},
  author={Wang, Weiyun and Shi, Min and Li, Qingyun and Wang, Wenhai and Huang, Zhenhang and Xing, Linjie and Chen, Zhe and Li, Hao and Zhu, Xizhou and Cao, Zhiguo and others},
  journal={arXiv preprint arXiv:2308.01907},
  year={2023}
}

@inproceedings{yan2024category,
  title = {Category-Prompt Refined Feature Learning for Long-Tailed Multi-Label Image Classification},
  author = {Yan, Jiexuan and Huang, Sheng and Mu, NanKun and Huangfu, Luwen and Liu, Bo},
  booktitle={Proceedings of the 32nd ACM international conference on Multimedia},
  pages = {2146–2155},
  year = {2024},
}

@inproceedings{meng2019multi,
  title={Multi-label image classification with attention mechanism and graph convolutional networks},
  author={Meng, Quanling and Zhang, Weigang},
  booktitle={Proceedings of the 1st ACM International Conference on Multimedia in Asia},
  pages={1--6},
  year={2019}
}

@article{li2020learning,
  title={Learning label correlations for multi-label image recognition with graph networks},
  author={Li, Qing and Peng, Xiaojiang and Qiao, Yu and Peng, Qiang},
  journal={Pattern Recognition Letters},
  volume={138},
  pages={378--384},
  year={2020},
  publisher={Elsevier}
}

@article{xu2022privacy,
  title={Privacy-preserving mechanisms for multi-label image recognition},
  author={Xu, Honghui and Cai, Zhipeng and Li, Wei},
  journal={ACM Transactions on Knowledge Discovery from Data (TKDD)},
  volume={16},
  number={4},
  pages={1--21},
  year={2022},
  publisher={ACM New York, NY}
}

@article{hassanin2022learning,
  title={Learning discriminative representations for multi-label image recognition},
  author={Hassanin, Mohammed and Radwan, Ibrahim and Khan, Salman and Tahtali, Murat},
  journal={Journal of Visual Communication and Image Representation},
  volume={83},
  pages={103448},
  year={2022},
  publisher={Elsevier}
}

@article{wang2016beyond,
  title={Beyond object proposals: Random crop pooling for multi-label image recognition},
  author={Wang, Meng and Luo, Changzhi and Hong, Richang and Tang, Jinhui and Feng, Jiashi},
  journal={IEEE Transactions on Image Processing},
  volume={25},
  number={12},
  pages={5678--5688},
  year={2016},
  publisher={IEEE}
}

@article{liang2022multi,
  title={A multi-scale semantic attention representation for multi-label image recognition with graph networks},
  author={Liang, Jun and Xu, Feiteng and Yu, Songsen},
  journal={Neurocomputing},
  volume={491},
  pages={14--23},
  year={2022},
  publisher={Elsevier}
}

@inproceedings{wang2020fast,
  title={Fast graph convolution network based multi-label image recognition via cross-modal fusion},
  author={Wang, Yangtao and Xie, Yanzhao and Liu, Yu and Zhou, Ke and Li, Xiaocui},
  booktitle={Proceedings of the 29th ACM International Conference on Information \& Knowledge Management},
  pages={1575--1584},
  year={2020}
}

@article{wang2022stmg,
  title={STMG: Swin transformer for multi-label image recognition with graph convolution network},
  author={Wang, Yangtao and Xie, Yanzhao and Fan, Lisheng and Hu, Guangxing},
  journal={Neural Computing and Applications},
  volume={34},
  number={12},
  pages={10051--10063},
  year={2022},
  publisher={Springer}
}

@article{nie2022multi,
  title={Multi-label image recognition with attentive transformer-localizer module},
  author={Nie, Lin and Chen, Tianshui and Wang, Zhouxia and Kang, Wenxiong and Lin, Liang},
  journal={Multimedia Tools and Applications},
  volume={81},
  number={6},
  pages={7917--7940},
  year={2022},
  publisher={Springer}
}

@inproceedings{yao2025gkgnet,
  title={Gkgnet: Group k-nearest neighbor based graph convolutional network for multi-label image recognition},
  author={Yao, Ruijie and Jin, Sheng and Xu, Lumin and Zeng, Wang and Liu, Wentao and Qian, Chen and Luo, Ping and Wu, Ji},
  booktitle={European Conference on Computer Vision},
  pages={91--107},
  year={2025},
  organization={Springer}
}

@article{yan2019multi,
  title={Multi-label image classification by feature attention network},
  author={Yan, Zheng and Liu, Weiwei and Wen, Shiping and Yang, Yin},
  journal={IEEE Access},
  volume={7},
  pages={98005--98013},
  year={2019},
  publisher={IEEE}
}

@article{chen2024msfa,
  title={MSFA: Multi-stage feature aggregation network for multi-label image recognition},
  author={Chen, Jiale and Xu, Feng and Zeng, Tao and Li, Xin and Chen, Shangjing and Yu, Jie},
  journal={IET Image Processing},
  year={2024},
  publisher={Wiley Online Library}
}

@article{chen2013multi,
  title={Multi-instance multi-label image classification: A neural approach},
  author={Chen, Zenghai and Chi, Zheru and Fu, Hong and Feng, Dagan},
  journal={Neurocomputing},
  volume={99},
  pages={298--306},
  year={2013},
  publisher={Elsevier}
}

@article{vallet2015multi,
  title={A multi-label convolutional neural network for automatic image annotation},
  author={Vallet, Alexis and Sakamoto, Hiroyasu},
  journal={Journal of information processing},
  volume={23},
  number={6},
  pages={767--775},
  year={2015},
  publisher={Information Processing Society of Japan}
}

@article{yao2022m,
  title={M-GCN: Brain-inspired memory graph convolutional network for multi-label image recognition},
  author={Yao, Xiao and Xu, Feiyang and Gu, Min and Wang, Peipei},
  journal={Neural Computing and Applications},
  pages={1--14},
  year={2022},
  publisher={Springer}
}

@article{park2020marsnet,
  title={MarsNet: multi-label classification network for images of various sizes},
  author={Park, Ju-Youn and Hwang, Yewon and Lee, Dukyoung and Kim, Jong-Hwan},
  journal={IEEE Access},
  volume={8},
  pages={21832--21846},
  year={2020},
  publisher={IEEE}
}

@article{wu2019tencent,
  title={Tencent ml-images: A large-scale multi-label image database for visual representation learning},
  author={Wu, Baoyuan and Chen, Weidong and Fan, Yanbo and Zhang, Yong and Hou, Jinlong and Liu, Jie and Zhang, Tong},
  journal={IEEE access},
  volume={7},
  pages={172683--172693},
  year={2019},
  publisher={IEEE}
}

@article{xie2022label,
  title={Label graph learning for multi-label image recognition with cross-modal fusion},
  author={Xie, Yanzhao and Wang, Yangtao and Liu, Yu and Zhou, Ke},
  journal={Multimedia Tools and Applications},
  volume={81},
  number={18},
  pages={25363--25381},
  year={2022},
  publisher={Springer}
}

@article{wu2022smart,
  title={SMART: Semantic-aware masked attention relational transformer for multi-label image recognition},
  author={Wu, Hongjun and Xu, Cheng and Liu, Hongzhe},
  journal={IEEE Signal Processing Letters},
  volume={29},
  pages={2158--2162},
  year={2022},
  publisher={IEEE}
}

@inproceedings{lydia2020multi,
  title={Multi-label classification using deep convolutional neural network},
  author={Lydia, A Agnes and Francis, F Sagayaraj},
  booktitle={2020 international conference on innovative trends in information technology (ICITIIT)},
  pages={1--6},
  year={2020},
  organization={IEEE}
}

@article{tan2024pvlr,
  title={PVLR: Prompt-driven Visual-Linguistic Representation Learning for Multi-Label Image Recognition},
  author={Tan, Hao and Tan, Zichang and Li, Jun and Wan, Jun and Lei, Zhen},
  journal={arXiv preprint arXiv:2401.17881},
  year={2024}
}

@inproceedings{chen2025modeling,
  title={Modeling Label Correlations with Latent Context for Multi-label Recognition},
  author={Chen, Zhaomin and Cui, Quan and Deng, Ruoxi and Hu, Jie and Zhang, Guodao},
  booktitle={European Conference on Computer Vision},
  pages={218--234},
  year={2025},
  organization={Springer}
}

@inproceedings{li2020multi,
  title={Multi-scale cross-modal spatial attention fusion for multi-label image recognition},
  author={Li, Junbing and Zhang, Changqing and Wang, Xueman and Du, Ling},
  booktitle={Artificial Neural Networks and Machine Learning--ICANN 2020: 29th International Conference on Artificial Neural Networks, Bratislava, Slovakia, September 15--18, 2020, Proceedings, Part I 29},
  pages={736--747},
  year={2020},
  organization={Springer}
}

@article{sun2023attention,
  title={An attention-driven multi-label image classification with semantic embedding and graph convolutional networks},
  author={Sun, Dengdi and Ma, Leilei and Ding, Zhuanlian and Luo, Bin},
  journal={Cognitive Computation},
  pages={1--12},
  year={2023},
  publisher={Springer}
}

@article{wang2022cross,
  title={Cross-modal fusion for multi-label image classification with attention mechanism},
  author={Wang, Yangtao and Xie, Yanzhao and Zeng, Jiangfeng and Wang, Hanpin and Fan, Lisheng and Song, Yufan},
  journal={Computers and Electrical Engineering},
  volume={101},
  pages={108002},
  year={2022},
  publisher={Elsevier}
}

@article{zhao2020double,
  title={Double attention for multi-label image classification},
  author={Zhao, Haiying and Zhou, Wei and Hou, Xiaogang and Zhu, Hui},
  journal={IEEE Access},
  volume={8},
  pages={225539--225550},
  year={2020},
  publisher={IEEE}
}

@article{tan2024sspa,
  title={SSPA: Split-and-Synthesize Prompting with Gated Alignments for Multi-Label Image Recognition},
  author={Tan, Hao and Tan, Zichang and Li, Jun and Wan, Jun and Lei, Zhen and Li, Stan Z},
  journal={arXiv preprint arXiv:2407.20920},
  year={2024}
}

@article{yan2016new,
  title={A New multi-instance multi-label learning approach for image and text classification},
  author={Yan, Kaobi and Li, Zhixin and Zhang, Canlong},
  journal={Multimedia Tools and Applications},
  volume={75},
  pages={7875--7890},
  year={2016},
  publisher={Springer}
}

@inproceedings{wang2021g,
  title={G-cam: graph convolution network based class activation mapping for multi-label image recognition},
  author={Wang, Yangtao and Xie, Yanzhao and Liu, Yu and Fan, Lisheng},
  booktitle={Proceedings of the 2021 International Conference on Multimedia Retrieval},
  pages={322--330},
  year={2021}
}

@article{cao2021multi,
  title={Multi-label image recognition with two-stream dynamic graph convolution networks},
  author={Cao, Pingping and Chen, Pengpeng and Niu, Qiang},
  journal={Image and Vision Computing},
  volume={113},
  pages={104238},
  year={2021},
  publisher={Elsevier}
}

@article{guo2018deep,
  title={Deep multi-instance multi-label learning for image annotation},
  author={Guo, Hai-Feng and Han, Lixin and Su, Shoubao and Sun, Zhou-Bao},
  journal={International Journal of Pattern Recognition and Artificial Intelligence},
  volume={32},
  number={03},
  pages={1859005},
  year={2018},
  publisher={World Scientific}
}

@article{chu2021multi,
  title={Multi-label image recognition by using semantics consistency, object correlation, and multiple samples},
  author={Chu, Wei-Ta and Huang, Si-Heng},
  journal={Journal of Visual Communication and Image Representation},
  volume={77},
  pages={103067},
  year={2021},
  publisher={Elsevier}
}

@inproceedings{yuan2024positive,
  title={Positive label is all you need for multi-label classification},
  author={Yuan, Zhixiang and Zhang, Kaixin and Huang, Tao},
  booktitle={2024 IEEE International Conference on Multimedia and Expo (ICME)},
  pages={1--6},
  year={2024},
  organization={IEEE}
}

@inproceedings{cheng2024towards,
  title={Towards Calibrated Multi-label Deep Neural Networks},
  author={Cheng, Jiacheng and Vasconcelos, Nuno},
  booktitle={Proceedings of the IEEE/CVF Conference on Computer Vision and Pattern Recognition},
  pages={27589--27599},
  year={2024}
}

@article{wu2024multi,
  title={A multi-label image classification method combining multi-stage image semantic information and label relevance},
  author={Wu, Liwen and Zhao, Lei and Tang, Peigeng and Pu, Bin and Jin, Xin and Zhang, Yudong and Yao, Shaowen},
  journal={International Journal of Machine Learning and Cybernetics},
  pages={1--15},
  year={2024},
  publisher={Springer}
}

@article{rawlekar2024improving,
  title={Improving Multi-label Recognition using Class Co-Occurrence Probabilities},
  author={Rawlekar, Samyak and Bhatnagar, Shubhang and Srinivasulu, Vishnuvardhan Pogunulu and Ahuja, Narendra},
  journal={arXiv preprint arXiv:2404.16193},
  year={2024}
}

@article{li2024multi,
  title={Multi-label category enhancement fusion distillation based on variational estimation},
  author={Li, Li and Xu, Jingzhou},
  journal={Knowledge-Based Systems},
  pages={112092},
  year={2024},
  publisher={Elsevier}
}

@inproceedings{dutta2020recurrent,
  title={Recurrent image annotation with explicit inter-label dependencies},
  author={Dutta, Ayushi and Verma, Yashaswi and Jawahar, CV},
  booktitle={European Conference on Computer Vision},
  pages={191--207},
  year={2020},
  organization={Springer}
}

@inproceedings{cheng2014bing,
  title={BING: Binarized normed gradients for objectness estimation at 300fps},
  author={Cheng, Ming-Ming and Zhang, Ziming and Lin, Wen-Yan and Torr, Philip},
  booktitle={Proceedings of the IEEE conference on computer vision and pattern recognition},
  pages={3286--3293},
  year={2014}
}

@inproceedings{zitnick2014edge,
  title={Edge boxes: Locating object proposals from edges},
  author={Zitnick, C Lawrence and Doll{\'a}r, Piotr},
  booktitle={Computer Vision--ECCV 2014: 13th European Conference, Zurich, Switzerland, September 6-12, 2014, Proceedings, Part V 13},
  pages={391--405},
  year={2014},
  organization={Springer}
}

@article{uijlings2013selective,
  title={Selective search for object recognition},
  author={Uijlings, Jasper RR and Van De Sande, Koen EA and Gevers, Theo and Smeulders, Arnold WM},
  journal={International journal of computer vision},
  volume={104},
  pages={154--171},
  year={2013},
  publisher={Springer}
}

@article{jaderberg2015spatial,
  title={Spatial transformer networks},
  author={Jaderberg, Max and Simonyan, Karen and Zisserman, Andrew and others},
  journal={Advances in neural information processing systems},
  volume={28},
  year={2015}
}

@article{ba2014multiple,
  title={Multiple object recognition with visual attention},
  author={Ba, Jimmy and Mnih, Volodymyr and Kavukcuoglu, Koray},
  journal={arXiv preprint arXiv:1412.7755},
  year={2014}
}

@article{ren2016faster,
  title={Faster R-CNN: Towards real-time object detection with region proposal networks},
  author={Ren, Shaoqing and He, Kaiming and Girshick, Ross and Sun, Jian},
  journal={IEEE transactions on pattern analysis and machine intelligence},
  volume={39},
  number={6},
  pages={1137--1149},
  year={2016},
  publisher={IEEE}
}

@inproceedings{lin2014microsoft,
  title={Microsoft coco: Common objects in context},
  author={Lin, Tsung-Yi and Maire, Michael and Belongie, Serge and Hays, James and Perona, Pietro and Ramanan, Deva and Doll{\'a}r, Piotr and Zitnick, C Lawrence},
  booktitle={European Conference on Computer Vision},
  pages={740--755},
  year={2014},
  organization={Springer}
}

@article{everingham2010pascal,
  title={The pascal visual object classes (voc) challenge},
  author={Everingham, Mark and Van Gool, Luc and Williams, Christopher KI and Winn, John and Zisserman, Andrew},
  journal={International journal of computer vision},
  volume={88},
  number={2},
  pages={303--338},
  year={2010},
  publisher={Springer}
}

@inproceedings{chua2009nus,
  title={Nus-wide: a real-world web image database from national university of singapore},
  author={Chua, Tat-Seng and Tang, Jinhui and Hong, Richang and Li, Haojie and Luo, Zhiping and Zheng, Yantao},
  booktitle={Proceedings of the ACM international conference on image and video retrieval},
  pages={1--9},
  year={2009}
}

@inproceedings{li2016human,
  title={Human attribute recognition by deep hierarchical contexts},
  author={Li, Yining and Huang, Chen and Loy, Chen Change and Tang, Xiaoou},
  booktitle={Computer Vision--ECCV 2016: 14th European Conference, Amsterdam, The Netherlands, October 11-14, 2016, Proceedings, Part VI 14},
  pages={684--700},
  year={2016},
  organization={Springer}
}

@article{kuznetsova2020open,
  title={The open images dataset v4: Unified image classification, object detection, and visual relationship detection at scale},
  author={Kuznetsova, Alina and Rom, Hassan and Alldrin, Neil and Uijlings, Jasper and Krasin, Ivan and Pont-Tuset, Jordi and Kamali, Shahab and Popov, Stefan and Malloci, Matteo and Kolesnikov, Alexander and others},
  journal={International journal of computer vision},
  volume={128},
  number={7},
  pages={1956--1981},
  year={2020},
  publisher={Springer}
}

@article{hochreiter1997long,
  title={Long Short-term Memory},
  author={Hochreiter, S},
  journal={Neural Computation MIT-Press},
  year={1997}
}

@article{wen2020multilabel,
  title={Multilabel image classification via feature/label co-projection},
  author={Wen, Shiping and Liu, Weiwei and Yang, Yin and Zhou, Pan and Guo, Zhenyuan and Yan, Zheng and Chen, Yiran and Huang, Tingwen},
  journal={IEEE Transactions on Systems, Man, and Cybernetics: Systems},
  volume={51},
  number={11},
  pages={7250--7259},
  year={2020},
  publisher={IEEE}
}

@article{li2024noah,
  title={NOAH: Learning Pairwise Object Category Attentions for Image Classification},
  author={Li, Chao and Zhou, Aojun and Yao, Anbang},
  journal={arXiv preprint arXiv:2402.02377},
  year={2024}
}

@inproceedings{wang2024beyond,
  title={Beyond Direct Relationships: Exploring Multi-Order Label Pair Dependencies for Knowledge Distillation},
  author={Wang, Jingchao and Deng, Zhengnan and Lin, Tongxu and Li, Wenyuan and Ling, Shaobin and Lin, Junyu},
  booktitle={Proceedings of the 32nd ACM International Conference on Multimedia},
  pages={8527--8535},
  year={2024}
}

@article{liu2024multi,
  title={Multi-Label Image Classification Based on Object Detection and Dynamic Graph Convolutional Networks.},
  author={Liu, Xiaoyu and Hu, Yong},
  journal={Computers, Materials \& Continua},
  volume={80},
  number={3},
  year={2024}
}

@inproceedings{sovatzidi2023towards,
  title={Towards the Interpretation of Multi-Label Image Classification Using Transformers and Fuzzy Cognitive Maps},
  author={Sovatzidi, Georgia and Vasilakakis, Michael D and Iakovidis, Dimitris K},
  booktitle={2023 IEEE International Conference on Fuzzy Systems (FUZZ)},
  pages={1--7},
  year={2023},
  organization={IEEE}
}

@article{huang2023cross,
  title={Cross-modality semantic guidance for multi-label image classification},
  author={Huang, Jun and Wang, Dian and Hong, Xudong and Qu, Xiwen and Xue, Wei},
  journal={Intelligent Data Analysis},
  number={Preprint},
  pages={1--14},
  year={2023},
  publisher={IOS Press}
}

@inproceedings{zhou2016learning,
  title={Learning deep features for discriminative localization},
  author={Zhou, Bolei and Khosla, Aditya and Lapedriza, Agata and Oliva, Aude and Torralba, Antonio},
  booktitle={Proceedings of the IEEE conference on computer vision and pattern recognition},
  pages={2921--2929},
  year={2016}
}

@article{kim2016hadamard,
  title={Hadamard product for low-rank bilinear pooling},
  author={Kim, Jin-Hwa and On, Kyoung-Woon and Lim, Woosang and Kim, Jeonghee and Ha, Jung-Woo and Zhang, Byoung-Tak},
  journal={arXiv preprint arXiv:1610.04325},
  year={2016}
}

@inproceedings{woo2018cbam,
  title={Cbam: Convolutional block attention module},
  author={Woo, Sanghyun and Park, Jongchan and Lee, Joon-Young and Kweon, In So},
  booktitle={Proceedings of the European conference on computer vision (ECCV)},
  pages={3--19},
  year={2018}
}

@article{zhou2020classify,
  title={Classify multi-label images via improved CNN model with adversarial network},
  author={Zhou, Tao and Li, Zhixin and Zhang, Canlong and Ma, Huifang},
  journal={Multimedia Tools and Applications},
  volume={79},
  number={9},
  pages={6871--6890},
  year={2020},
  publisher={Springer}
}

@article{zhou2023attention,
  title={Attention-augmented memory network for image multi-label classification},
  author={Zhou, Wei and Hou, Yanke and Chen, Dihu and Hu, Haifeng and Su, Tao},
  journal={ACM Transactions on Multimedia Computing, Communications and Applications},
  volume={19},
  number={3},
  pages={1--24},
  year={2023},
  publisher={ACM New York, NY}
}

@inproceedings{ablavatski2017enriched,
  title={Enriched deep recurrent visual attention model for multiple object recognition},
  author={Ablavatski, Artsiom and Lu, Shijian and Cai, Jianfei},
  booktitle={2017 IEEE Winter Conference on Applications of Computer Vision (WACV)},
  pages={971--978},
  year={2017},
  organization={IEEE}
}

@article{vaswani2017attention,
  title={Attention is all you need},
  author={Vaswani, A},
  journal={Advances in Neural Information Processing Systems},
  year={2017}
}

@inproceedings{liu2017semantic,
  title={Semantic regularisation for recurrent image annotation},
  author={Liu, Feng and Xiang, Tao and Hospedales, Timothy M and Yang, Wankou and Sun, Changyin},
  booktitle={Proceedings of the IEEE Conference on Computer Vision and Pattern Recognition},
  pages={2872--2880},
  year={2017}
}

@article{krishna2017visual,
  title={Visual genome: Connecting language and vision using crowdsourced dense image annotations},
  author={Krishna, Ranjay and Zhu, Yuke and Groth, Oliver and Johnson, Justin and Hata, Kenji and Kravitz, Joshua and Chen, Stephanie and Kalantidis, Yannis and Li, Li-Jia and Shamma, David A and others},
  journal={International journal of computer vision},
  volume={123},
  pages={32--73},
  year={2017},
  publisher={Springer}
}

@inproceedings{jin2016annotation,
  title={Annotation order matters: Recurrent image annotator for arbitrary length image tagging},
  author={Jin, Jiren and Nakayama, Hideki},
  booktitle={2016 23rd international conference on pattern recognition (ICPR)},
  pages={2452--2457},
  year={2016},
  organization={IEEE}
}

@inproceedings{pennington2014glove,
  title={Glove: Global vectors for word representation},
  author={Pennington, Jeffrey and Socher, Richard and Manning, Christopher D},
  booktitle={Proceedings of the 2014 conference on empirical methods in natural language processing (EMNLP)},
  pages={1532--1543},
  year={2014}
}

@article{qing2019learning,
  title={Learning category correlations for multi-label image recognition with graph networks},
  author={Qing, Li and Xiaojiang, Peng and Yu, Qiao and Qiang, Peng},
  journal={arXiv preprint arXiv:1909.13005},
  year={2019}
}

@inproceedings{he2017mask,
  title={Mask r-cnn},
  author={He, Kaiming and Gkioxari, Georgia and Doll{\'a}r, Piotr and Girshick, Ross},
  booktitle={Proceedings of the IEEE international conference on computer vision},
  pages={2961--2969},
  year={2017}
}

@article{devlin2018bert,
  title={Bert: Pre-training of deep bidirectional transformers for language understanding},
  author={Devlin, Jacob},
  journal={arXiv preprint arXiv:1810.04805},
  year={2018}
}

@inproceedings{vu2020privacy,
  title={Privacy-preserving visual content tagging using graph transformer networks},
  author={Vu, Xuan-Son and Le, Duc-Trong and Edlund, Christoffer and Jiang, Lili and Nguyen, Hoang D},
  booktitle={Proceedings of the 28th ACM International Conference on Multimedia},
  pages={2299--2307},
  year={2020}
}

@inproceedings{chen2024pursuit,
  title={In pursuit of causal label correlations for multi-label image recognition},
  author={Chen, Zhao-Min and Jin, Xin and Chan, Sixian and others},
  journal={Advances in neural information processing systems},
  volume={38},
  year={2024}
}

@article{sajedi2023end,
  title={End-to-end supervised multilabel contrastive learning},
  author={Sajedi, Ahmad and Khaki, Samir and Plataniotis, Konstantinos N and Hosseini, Mahdi S},
  journal={arXiv preprint arXiv:2307.03967},
  year={2023}
}

@inproceedings{singh2022iml,
  title={Iml-gcn: Improved multi-label graph convolutional network for efficient yet precise image classification},
  author={Singh, Inder Pal and Oyedotun, Oyebade and Ghorbel, Enjie and Aouada, Djamila},
  booktitle={AAAI-22 Workshop Program-Deep Learning on Graphs: Methods and Applications},
  year={2022}
}

@inproceedings{seymour2018multi,
  title={Multi-label triplet embeddings for image annotation from user-generated tags},
  author={Seymour, Zachary and Zhang, Zhongfei},
  booktitle={Proceedings of the 2018 ACM on International Conference on Multimedia Retrieval},
  pages={249--256},
  year={2018}
}

@article{dosovitskiy2020image,
  title={An image is worth 16x16 words: Transformers for image recognition at scale},
  author={Dosovitskiy, Alexey and Beyer, Lucas and Kolesnikov, Alexander and Weissenborn, Dirk and Zhai, Xiaohua and Unterthiner, Thomas and Dehghani, Mostafa and Minderer, Matthias and Heigold, Georg and Gelly, Sylvain and others},
  journal={arXiv preprint arXiv:2010.11929},
  year={2020}
}

@inproceedings{lyu2018coarse,
  title={Coarse to fine: Multi-label image classification with global/local attention},
  author={Lyu, Fan and Hu, Fuyuan and Sheng, Victor S and Wu, Zhengtian and Fu, Qiming and Fu, Baochuan},
  booktitle={2018 IEEE International Smart Cities Conference (ISC2)},
  pages={1--7},
  year={2018},
  organization={IEEE}
}

@inproceedings{zhang2024multi,
  title={Multi-label supervised contrastive learning},
  author={Zhang, Pingyue and Wu, Mengyue},
  booktitle={Proceedings of the AAAI Conference on Artificial Intelligence},
  volume={38},
  number={15},
  pages={16786--16793},
  year={2024}
}

@article{lyu2019multi,
  title={Multi-label image classification via Coarse-to-Fine attention},
  author={Lyu, Fan and Li, Linyan and Victor, S Sheng and Fu, Qiming and Hu, Fuyuan},
  journal={Chinese Journal of Electronics},
  volume={28},
  number={6},
  pages={1118--1126},
  year={2019},
  publisher={Wiley Online Library}
}

@inproceedings{le2016fully,
  title={Fully automated multi-label image annotation by convolutional neural network and adaptive thresholding},
  author={Le, Hoa M and Nguyen, Thi-Oanh and Ngo-Tien, Dung},
  booktitle={Proceedings of the 7th Symposium on Information and Communication Technology},
  pages={323--330},
  year={2016}
}

@inproceedings{yang2023multi,
  title={Multi-label knowledge distillation},
  author={Yang, Penghui and Xie, Ming-Kun and Zong, Chen-Chen and Feng, Lei and Niu, Gang and Sugiyama, Masashi and Huang, Sheng-Jun},
  booktitle={Proceedings of the IEEE/CVF international conference on computer vision},
  pages={17271--17280},
  year={2023}
}

@inproceedings{chen2019multi2,
  title={Multi-label image recognition with joint class-aware map disentangling and label correlation embedding},
  author={Chen, Zhao-Min and Wei, Xiu-Shen and Jin, Xin and Guo, Yanwen},
  booktitle={2019 IEEE International Conference on Multimedia and Expo (ICME)},
  pages={622--627},
  year={2019},
  organization={IEEE}
}

@inproceedings{niu2019coupled,
  title={Coupled Dictionary Learning for Multi-label Embedding},
  author={Niu, Sijia and Xu, Qian and Zhu, Pengfei and Hu, Qinghua and Shi, Hong},
  booktitle={2019 International Joint Conference on Neural Networks (IJCNN)},
  pages={1--8},
  year={2019},
  organization={IEEE}
}

@article{cao2015sled,
  title={SLED: semantic label embedding dictionary representation for multilabel image annotation},
  author={Cao, Xiaochun and Zhang, Hua and Guo, Xiaojie and Liu, Si and Meng, Dan},
  journal={IEEE Transactions on Image Processing},
  volume={24},
  number={9},
  pages={2746--2759},
  year={2015},
  publisher={IEEE}
}

@inproceedings{zhang2024dbddl,
  title={DBDDL: Double-Branch Deep Dictionary Learning for Multi-Label Image Classification},
  author={Zhang, Wenke and Liao, MengMeng},
  booktitle={Proceedings of the 2024 6th International Conference on Big-data Service and Intelligent Computation},
  pages={35--40},
  year={2024}
}

@article{gu2022multilabel,
  title={Multilabel convolutional network with feature denoising and details supplement},
  author={Gu, Tianhao and Wang, Zhe and Fang, Zhongli and Zhu, Zonghai and Yang, Hai and Li, Dongdong and Du, Wenli},
  journal={IEEE Transactions on Neural Networks and Learning Systems},
  volume={34},
  number={11},
  pages={8349--8361},
  year={2022},
  publisher={IEEE}
}

@article{zhou2025drtn,
  title={DRTN: Dual Relation Transformer Network with feature erasure and contrastive learning for multi-label image classification},
  author={Zhou, Wei and Lin, Kang and Zheng, Zhijie and Chen, Dihu and Su, Tao and Hu, Haifeng},
  journal={Neural Networks},
  pages={107309},
  year={2025},
  publisher={Elsevier}
}

@inproceedings{zhong2020random,
  title={Random erasing data augmentation},
  author={Zhong, Zhun and Zheng, Liang and Kang, Guoliang and Li, Shaozi and Yang, Yi},
  booktitle={Proceedings of the AAAI conference on artificial intelligence},
  volume={34},
  number={07},
  pages={13001--13008},
  year={2020}
}

@inproceedings{guo2021long,
  title={Long-tailed multi-label visual recognition by collaborative training on uniform and re-balanced samplings},
  author={Guo, Hao and Wang, Song},
  booktitle={Proceedings of the IEEE/CVF Conference on Computer Vision and Pattern Recognition},
  pages={15089--15098},
  year={2021}
}

@article{xia2023lmpt,
  title={LMPT: prompt tuning with class-specific embedding loss for long-tailed multi-label visual recognition},
  author={Xia, Peng and Xu, Di and Hu, Ming and Ju, Lie and Ge, Zongyuan},
  journal={arXiv preprint arXiv:2305.04536},
  year={2023}
}

@inproceedings{radford2021learning,
  title={Learning transferable visual models from natural language supervision},
  author={Radford, Alec and Kim, Jong Wook and Hallacy, Chris and Ramesh, Aditya and Goh, Gabriel and Agarwal, Sandhini and Sastry, Girish and Askell, Amanda and Mishkin, Pamela and Clark, Jack and others},
  booktitle={International conference on machine learning},
  pages={8748--8763},
  year={2021},
  organization={PmLR}
}

@inproceedings{xianhua2021image,
  title={Image Multi-Label Classification Based on Pyramid Convolution and Split-Attention Mechanism},
  author={Xianhua, Yang and Yi, Yang and Juan, Yang and Han, Yao and Zheng, Wang and Shuquan, Long},
  booktitle={2021 18th International Computer Conference on Wavelet Active Media Technology and Information Processing (ICCWAMTIP)},
  pages={534--538},
  year={2021},
  organization={IEEE}
}

@inproceedings{dong2022towards,
  title={Towards class interpretable vision transformer with multi-class-tokens},
  author={Dong, Bowen and Zhou, Pan and Yan, Shuicheng and Zuo, Wangmeng},
  booktitle={Chinese Conference on Pattern Recognition and Computer Vision (PRCV)},
  pages={609--622},
  year={2022},
  organization={Springer}
}

@article{chauhan2023tackling,
  title={Tackling over-smoothing in multi-label image classification using graphical convolution neural network},
  author={Chauhan, Vikas and Tiwari, Aruna and Venkata, Boppudi and Naik, Vislavath},
  journal={Evolving Systems},
  volume={14},
  number={5},
  pages={771--781},
  year={2023},
  publisher={Springer}
}

@inproceedings{zhou2021multiple,
  title={Multiple Semantic Embedding with Graph Convolutional Networks for Multi-Label Image Classification},
  author={Zhou, Tong and Feng, Songhe},
  booktitle={Pattern Recognition and Computer Vision: 4th Chinese Conference, PRCV 2021, Beijing, China, October 29--November 1, 2021, Proceedings, Part II 4},
  pages={449--461},
  year={2021},
  organization={Springer}
}

@inproceedings{wang2024dynamic,
  title={Dynamic Label Bidirectional Fusion Based Cross-modal Multi-label Image Classification},
  author={Wang, Rui and Tian, Mei and Luo, Yan and Kou, Zhe},
  booktitle={2024 IEEE 17th International Conference on Signal Processing (ICSP)},
  pages={395--399},
  year={2024},
  organization={IEEE}
}

@inproceedings{fu2024generative,
  title={Generative Contrastive Learning for Multi-Label Image Classification},
  author={Fu, ShengWu and Gu, ZhengShen and Wang, Dong and Xu, Songhua},
  booktitle={2024 International Symposium on Digital Home (ISDH)},
  pages={73--78},
  year={2024},
  organization={IEEE}
}

@inproceedings{kuang2025two,
  title={Two-stream Semantic Alignment Networks for Multi-label Image Classification},
  author={Kuang, Wenlan and Li, Zhixin},
  booktitle={ICASSP 2025-2025 IEEE International Conference on Acoustics, Speech and Signal Processing (ICASSP)},
  pages={1--5},
  year={2025},
  organization={IEEE}
}

@inproceedings{xu2023research,
  title={Research on Multi-Labels Image Classification Based on Self-Supervised Model},
  author={Xu, Xuetian},
  booktitle={2022 International Conference on Image Processing and Computer Vision (IPCV)},
  pages={56--59},
  year={2023},
  organization={IEEE}
}

@inproceedings{li2019classification,
  title={A classification method for small sample multi-label images},
  author={Li, Ruohan and Jiang, Zengru and Dai, Wei and Nie, Yongkang and Liu, Liang and Dai, Yaping},
  booktitle={2019 Chinese Control And Decision Conference (CCDC)},
  pages={1365--1370},
  year={2019},
  organization={IEEE}
}

@inproceedings{shen2023improved,
  title={An improved multi-label classification algorithm based on YOLOV5s},
  author={Shen, Chen and Yi, Mingfa and Fu, Maosheng},
  booktitle={2023 3rd International Conference on Neural Networks, Information and Communication Engineering (NNICE)},
  pages={674--678},
  year={2023},
  organization={IEEE}
}

@inproceedings{gu2024hypergraph,
  title={Hypergraph Transformer for Multi-Label Image Classification},
  author={Gu, ZhengShen and Fu, ShengWu and Wang, Dong and Xu, SongHua},
  booktitle={2024 International Symposium on Digital Home (ISDH)},
  pages={79--84},
  year={2024},
  organization={IEEE}
}

@inproceedings{li2023improved,
  title={An Improved Attention Mechanism Fusion for Multi-Label Image Classification},
  author={Li, Zhongliang and Yuan, Haoliang},
  booktitle={2023 International Conference on Wavelet Analysis and Pattern Recognition (ICWAPR)},
  pages={103--108},
  year={2023},
  organization={IEEE}
}

@article{liu2025transformer,
  title={Transformer-driven feature fusion network and visual feature coding for multi-label image classification},
  author={Liu, Pingzhu and Qian, Wenbin and Huang, Jintao and Tu, Yanqiang and Cheung, Yiu-Ming},
  journal={Pattern Recognition},
  pages={111584},
  year={2025},
  publisher={Elsevier}
}

@article{wang2025establishing,
  title={Establishing Two-Dimensional Dependencies for Multi-Label Image Classification},
  author={Wang, Jiuhang and Zhang, Yuewen and Wang, Tengjing and Tang, Hongying and Li, Baoqing},
  journal={Applied Sciences},
  volume={15},
  number={5},
  pages={2845},
  year={2025},
  publisher={MDPI}
}

@article{lin2025adaptive,
  title={Adaptive knowledge graph for multi-label image classification},
  author={Lin, Zhihong and Tang, Xue-song and Hao, Kuangrong and Zhao, Mingbo and Li, Yubing},
  journal={Applied Intelligence},
  volume={55},
  number={1},
  pages={20},
  year={2025},
  publisher={Springer}
}

@inproceedings{huang2025multi,
  title={Multi-label image classification with semantic information and upsampling},
  author={Huang, Shuang},
  booktitle={Fourth International Conference on Computer Vision, Application, and Algorithm (CVAA 2024)},
  volume={13486},
  pages={25--30},
  year={2025},
  organization={SPIE}
}

@inproceedings{jiu2025deep,
  title={Deep Multi-order Context-Aware Kernel Network for Multi-label Classification},
  author={Jiu, Mingyuan and Zhu, Hailong and Sahbi, Hichem},
  booktitle={International Conference on Pattern Recognition},
  pages={1--17},
  year={2025},
  organization={Springer}
}

@article{chen2025towards,
  title={Towards Gradient Equalization and Feature Diversification for Long-Tailed Multi-Label Image Recognition},
  author={Chen, Zhao-Min and Cui, Quan and Zhang, Xiaoqin and Deng, Ruoxi and Xia, Chaoqun and Lu, Shijian},
  journal={IEEE Transactions on Multimedia},
  year={2025},
  publisher={IEEE}
}

@article{chen2019multi3,
  title={Multi-label image classification with recurrently learning semantic dependencies},
  author={Chen, Long and Wang, Ronggui and Yang, Juan and Xue, Lixia and Hu, Min},
  journal={The Visual Computer},
  volume={35},
  pages={1361--1371},
  year={2019},
  publisher={Springer}
}

@inproceedings{li2018multi,
  title={A multi-label image classification algorithm based on attention model},
  author={Li, Yugang and Wang, Yongbin},
  booktitle={2018 IEEE/ACIS 17th International Conference on Computer and Information Science (ICIS)},
  pages={728--731},
  year={2018},
  organization={IEEE}
}

@inproceedings{li2018attentive,
  title={Attentive recurrent neural network for weak-supervised multi-label image classification},
  author={Li, Liang and Wang, Shuhui and Jiang, Shuqiang and Huang, Qingming},
  booktitle={Proceedings of the 26th ACM international conference on Multimedia},
  pages={1092--1100},
  year={2018}
}

@article{li2018improving,
  title={Improving multi-label classification using scene cues},
  author={Li, Zhao and Lu, Wei and Sun, Zhanquan and Xing, Weiwei},
  journal={Multimedia Tools and Applications},
  volume={77},
  pages={6079--6094},
  year={2018},
  publisher={Springer}
}

@article{krizhevsky2012imagenet,
  title={Imagenet classification with deep convolutional neural networks},
  author={Krizhevsky, Alex and Sutskever, Ilya and Hinton, Geoffrey E},
  journal={Advances in neural information processing systems},
  volume={25},
  year={2012}
}

@inproceedings{liu2021swin,
  title={Swin transformer: Hierarchical vision transformer using shifted windows},
  author={Liu, Ze and Lin, Yutong and Cao, Yue and Hu, Han and Wei, Yixuan and Zhang, Zheng and Lin, Stephen and Guo, Baining},
  booktitle={Proceedings of the IEEE/CVF international conference on computer vision},
  pages={10012--10022},
  year={2021}
}

@inproceedings{lin2017feature,
  title={Feature pyramid networks for object detection},
  author={Lin, Tsung-Yi and Doll{\'a}r, Piotr and Girshick, Ross and He, Kaiming and Hariharan, Bharath and Belongie, Serge},
  booktitle={Proceedings of the IEEE conference on computer vision and pattern recognition},
  pages={2117--2125},
  year={2017}
}

@inproceedings{simonyan2015very,
  title={Very deep convolutional networks for large-scale image recognition},
  author={Simonyan, Karen and Zisserman, Andrew},
  booktitle={The third International Conference on Learning Representations},
  year={2015}
}

@inproceedings{he2016deep,
  title={Deep residual learning for image recognition},
  author={He, Kaiming and Zhang, Xiangyu and Ren, Shaoqing and Sun, Jian},
  booktitle={Proceedings of the IEEE conference on computer vision and pattern recognition},
  pages={770--778},
  year={2016}
}

@INPROCEEDINGS{afxentiou2023multi,
  title={Multi-Label Weather Image Classification for Embedded Computing Platforms}, 
  author={Afxentiou, Viktoria and Vladimirova, Tanya},
  booktitle={2023 IEEE Smart World Congress}, 
  year={2023},
  pages={1-8}
}

@inproceedings{pal2025label,
  title={Label Dependency Aware Loss for Reliable Multi-Label Medical Image Classification},
  author={Pal, Aditya Shankar and Panda, Arkapal and Garain, Utpal},
  booktitle={ICASSP 2025-2025 IEEE International Conference on Acoustics, Speech and Signal Processing (ICASSP)},
  pages={1--5},
  year={2025},
  organization={IEEE}
}

@article{holste2024towards,
  title={Towards long-tailed, multi-label disease classification from chest X-ray: Overview of the CXR-LT challenge},
  author={Holste, Gregory and Zhou, Yiliang and Wang, Song and Jaiswal, Ajay and Lin, Mingquan and Zhuge, Sherry and Yang, Yuzhe and Kim, Dongkyun and Nguyen-Mau, Trong-Hieu and Tran, Minh-Triet and others},
  journal={Medical Image Analysis},
  pages={103224},
  year={2024},
  publisher={Elsevier}
}

@article{dai2022feature,
  title={Feature disentangling and reciprocal learning with label-guided similarity for multi-label image retrieval},
  author={Dai, Yong and Song, Weiwei and Li, Yi and Di Stefano, Luigi},
  journal={Neurocomputing},
  volume={511},
  pages={353--365},
  year={2022},
  publisher={Elsevier}
}

@article{li2021ml,
  title={ML-ANet: A transfer learning approach using adaptation network for multi-label image classification in autonomous driving},
  author={Li, Guofa and Ji, Zefeng and Chang, Yunlong and Li, Shen and Qu, Xingda and Cao, Dongpu},
  journal={Chinese Journal of Mechanical Engineering},
  volume={34},
  pages={1--11},
  year={2021},
  publisher={Springer}
}

@article{ge2021multi,
  title={Multi-label correlation guided feature fusion network for abnormal ECG diagnosis},
  author={Ge, Zhaoyang and Jiang, Xiaoheng and Tong, Zhuang and Feng, Panpan and Zhou, Bing and Xu, Mingliang and Wang, Zongmin and Pang, Yanwei},
  journal={Knowledge-Based Systems},
  volume={233},
  pages={107508},
  year={2021},
  publisher={Elsevier}
}

@inproceedings{huang2013multi,
  title={Multi-task deep neural network for multi-label learning},
  author={Huang, Yan and Wang, Wei and Wang, Liang and Tan, Tieniu},
  booktitle={2013 IEEE International conference on image processing},
  pages={2897--2900},
  year={2013},
  organization={IEEE}
}

@article{devkar2017survey,
  title={A survey on multi-label classification for images},
  author={Devkar, Radhika and Shiravale, Sankirti},
  journal={International Journal of Computer Application},
  volume={162},
  number={8},
  pages={39--42},
  year={2017}
}

@article{tarekegn2024deep,
  title={Deep learning for multi-label learning: a comprehensive survey},
  author={Tarekegn, Adane Nega and Ullah, Mohib and Cheikh, Faouzi Alaya},
  journal={arXiv preprint arXiv:2401.16549},
  year={2024}
}

@article{liu2021emerging,
  title={The emerging trends of multi-label learning},
  author={Liu, Weiwei and Wang, Haobo and Shen, Xiaobo and Tsang, Ivor W},
  journal={IEEE transactions on pattern analysis and machine intelligence},
  volume={44},
  number={11},
  pages={7955--7974},
  year={2021},
  publisher={IEEE}
}

@article{han2023survey,
  title={A survey of multi-label classification based on supervised and semi-supervised learning},
  author={Han, Meng and Wu, Hongxin and Chen, Zhiqiang and Li, Muhang and Zhang, Xilong},
  journal={International Journal of Machine Learning and Cybernetics},
  volume={14},
  number={3},
  pages={697--724},
  year={2023},
  publisher={Springer}
}

@article{wei2022survey,
  title={A survey on extreme multi-label learning},
  author={Wei, Tong and Mao, Zhen and Shi, Jiang-Xin and Li, Yu-Feng and Zhang, Min-Ling},
  journal={arXiv preprint arXiv:2210.03968},
  year={2022}
}

@article{tarekegn2021review,
  title={A review of methods for imbalanced multi-label classification},
  author={Tarekegn, Adane Nega and Giacobini, Mario and Michalak, Krzysztof},
  journal={Pattern Recognition},
  volume={118},
  pages={107965},
  year={2021},
  publisher={Elsevier}
}

@inproceedings{chen2022survey,
  title={A survey of multi-label text classification based on deep learning},
  author={Chen, Xiaolong and Cheng, Jieren and Liu, Jingxin and Xu, Wenghang and Hua, Shuai and Tang, Zhu and Sheng, Victor S},
  booktitle={International Conference on Adaptive and Intelligent Systems},
  pages={443--456},
  year={2022},
  organization={Springer}
}

@article{wever2021automl,
  title={AutoML for multi-label classification: Overview and empirical evaluation},
  author={Wever, Marcel and Tornede, Alexander and Mohr, Felix and H{\"u}llermeier, Eyke},
  journal={IEEE transactions on pattern analysis and machine intelligence},
  volume={43},
  number={9},
  pages={3037--3054},
  year={2021},
  publisher={IEEE}
}

@article{qian2023survey,
  title={A survey on multi-label feature selection from perspectives of label fusion},
  author={Qian, Wenbin and Huang, Jintao and Xu, Fankang and Shu, Wenhao and Ding, Weiping},
  journal={Information Fusion},
  volume={100},
  pages={101948},
  year={2023},
  publisher={Elsevier}
}

@article{bogatinovski2022comprehensive,
  title={Comprehensive comparative study of multi-label classification methods},
  author={Bogatinovski, Jasmin and Todorovski, Ljup{\v{c}}o and D{\v{z}}eroski, Sa{\v{s}}o and Kocev, Dragi},
  journal={Expert Systems with Applications},
  volume={203},
  pages={117215},
  year={2022},
  publisher={Elsevier}
}

@inproceedings{kirillov2023segment,
  title={Segment anything},
  author={Kirillov, Alexander and Mintun, Eric and Ravi, Nikhila and Mao, Hanzi and Rolland, Chloe and Gustafson, Laura and Xiao, Tete and Whitehead, Spencer and Berg, Alexander C and Lo, Wan-Yen and others},
  booktitle={Proceedings of the IEEE/CVF international conference on computer vision},
  pages={4015--4026},
  year={2023}
}

@inproceedings{guo2023texts,
  title={Texts as images in prompt tuning for multi-label image recognition},
  author={Guo, Zixian and Dong, Bowen and Ji, Zhilong and Bai, Jinfeng and Guo, Yiwen and Zuo, Wangmeng},
  booktitle={Proceedings of the IEEE/CVF Conference on Computer Vision and Pattern Recognition},
  pages={2808--2817},
  year={2023}
}

@inproceedings{ridnik2021tresnet,
  title={Tresnet: High performance gpu-dedicated architecture},
  author={Ridnik, Tal and Lawen, Hussam and Noy, Asaf and Ben Baruch, Emanuel and Sharir, Gilad and Friedman, Itamar},
  booktitle={proceedings of the IEEE/CVF winter conference on applications of computer vision},
  pages={1400--1409},
  year={2021}
}

@inproceedings{xie2017aggregated,
  title={Aggregated residual transformations for deep neural networks},
  author={Xie, Saining and Girshick, Ross and Doll{\'a}r, Piotr and Tu, Zhuowen and He, Kaiming},
  booktitle={Proceedings of the IEEE conference on computer vision and pattern recognition},
  pages={1492--1500},
  year={2017}
}

@inproceedings{peng2023rwkv,
  title={RWKV: Reinventing RNNs for the Transformer Era},
  author={Peng, Bo and Alcaide, Eric and Anthony, Quentin and Albalak, Alon and Arcadinho, Samuel and Biderman, Stella and Cao, Huanqi and Cheng, Xin and Chung, Michael and Grella, Matteo and others},
  booktitle={Findings of the Association for Computational Linguistics: EMNLP 2023},
  pages={14048--14077},
  year={2023}
}

@article{gu2023mamba,
  title={Mamba: Linear-time sequence modeling with selective state spaces},
  author={Gu, Albert and Dao, Tri},
  journal={arXiv preprint arXiv:2312.00752},
  year={2023}
}

@article{liu2024kan,
  title={Kan: Kolmogorov-arnold networks},
  author={Liu, Ziming and Wang, Yixuan and Vaidya, Sachin and Ruehle, Fabian and Halverson, James and Solja{\v{c}}i{\'c}, Marin and Hou, Thomas Y and Tegmark, Max},
  journal={arXiv preprint arXiv:2404.19756},
  year={2024}
}

@inproceedings{zhu2025mambaml,
  title={MambaML: Exploring State Space Models for Multi-Label Image Classification},
  author={Zhu, Xuelin and Liu, Jian and Cao, Jiuxin and Wang, Bing},
  booktitle={Proceedings of the IEEE/CVF International Conference on Computer Vision},
  pages={4743--4753},
  year={2025}
}

@article{fang2025amita,
  title={AMITA: Attribute-guided Masked Image-Text Alignment for Multi-label Image Representation},
  author={Fang, Jinyi and Zhu, Bingke and Yuan, Jingling and Chen, Yingying and Tang, Ming and Wang, Jinqiao},
  journal={IEEE Transactions on Circuits and Systems for Video Technology},
  year={2025},
  publisher={IEEE}
}

@inproceedings{zhao2025towards,
  title={Towards Space and Semantics: Object-Purified Representation Learning for Multi-Label Image Classification},
  author={Zhao, Haifeng and Xu, Shuo and Ma, Leilei and Zhang, Yufei and Wang, Lei and Sun, Dengdi},
  booktitle={Proceedings of the 33rd ACM International Conference on Multimedia},
  pages={3270--3279},
  year={2025}
}

@article{jiu2025multi,
  title={Multi-label Classification with Panoptic Context Aggregation Networks},
  author={Jiu, Mingyuan and Zhu, Hailong and Wei, Wenchuan and Sahbi, Hichem and Ji, Rongrong and Xu, Mingliang},
  journal={arXiv preprint arXiv:2512.23486},
  year={2025}
}

@article{ye2025multi,
  title={Multi-label image classification model based on multiscale fusion and adaptive label correlation},
  author={Ye, Jihua and Jiang, Lu and Xiao, Shunjie and Zong, Ye and Jiang, Aiwen},
  journal={Journal of Shanghai Jiaotong University (Science)},
  volume={30},
  number={5},
  pages={889--898},
  year={2025},
  publisher={Springer}
}

@article{zhong2025multi,
  title={Multi-Scale Feature Fusion and Advanced Representation Learning for Multi Label Image Classification.},
  author={Zhong, Naikang and Lin, Xiao and Du, Wen and Shi, Jin},
  journal={Computers, Materials \& Continua},
  volume={82},
  number={3},
  year={2025}
}

@article{wang2025splicemix,
  title={Splicemix: A cross-scale and semantic blending augmentation strategy for multi-label image classification},
  author={Wang, Lei and Zhan, Yibing and Ma, Leilei and Tao, Dapeng and Ding, Liang and Gong, Chen},
  journal={IEEE Transactions on Multimedia},
  year={2025},
  volume={27},
  number={},
  pages={3251-3265},
  publisher={IEEE}
}

@article{du2025mlmamba,
  title={MLMamba: A Mamba-based Efficient Network for Multi-label Remote Sensing Scene Classification},
  author={Du, Ruiqi and Tang, Xu and Ma, Jingjing and Zhang, Xiangrong and Jiao, Licheng},
  journal={IEEE Transactions on Circuits and Systems for Video Technology},
  year={2025},
  volume={35},
  number={7},
  pages={6245-6258},
  publisher={IEEE}
}

@inproceedings{deng2009imagenet,
  title={Imagenet: A large-scale hierarchical image database},
  author={Deng, Jia and Dong, Wei and Socher, Richard and Li, Li-Jia and Li, Kai and Fei-Fei, Li},
  booktitle={2009 IEEE conference on computer vision and pattern recognition},
  pages={248--255},
  year={2009},
  organization={Ieee}
}

@article{abulfaraj2025deep,
  title={A deep ensemble learning approach based on a vision transformer and neural network for multi-label image classification},
  author={Abulfaraj, Anas W and Binzagr, Faisal},
  journal={Big Data and Cognitive Computing},
  volume={9},
  number={2},
  pages={39},
  year={2025},
  publisher={MDPI}
}

@inproceedings{tao2025mlc,
  title={Mlc-nc: Long-tailed multi-label image classification through the lens of neural collapse},
  author={Tao, Zijian and Li, Shao-Yuan and Wan, Wenhai and Zheng, Jinpeng and Chen, Jia-Yao and Li, Yuchen and Huang, Sheng-Jun and Chen, Songcan},
  booktitle={Proceedings of the AAAI Conference on Artificial Intelligence},
  volume={39},
  number={19},
  pages={20850--20858},
  year={2025}
}

@article{kuang2026multi,
  title={Multi-modal Feature Alignment Networks for Multi-label Image Classification},
  author={Kuang, Wenlan and Li, Zhixin},
  journal={Neural Networks},
  pages={108629},
  year={2026},
  publisher={Elsevier}
}

@article{wang2026topic,
  title={Topic-aware transformer with hierarchical prompting learning for multi-label image classification},
  author={Wang, Jiangpeng and Dai, Xiao and Shi, Benyun and Liu, Miao and Peng, Yue},
  journal={PeerJ Computer Science},
  volume={12},
  pages={e3552},
  year={2026},
  publisher={PeerJ Inc.}
}

@article{li2026multi,
  title={A multi-label image classification method via graph attention network with dynamic and static label correlations},
  author={Li, Zhiming and Zhou, Kai and Chen, Bingnan and Lv, Yuchen and He, Yihao and You, Dianlong},
  journal={International Journal of Machine Learning and Cybernetics},
  volume={17},
  number={2},
  pages={46},
  year={2026},
  publisher={Springer}
}

@article{fan2026exploring,
  title={Exploring Prompt Distributions and Probability Bias for Long-Tailed Multi-Label Image Recognition},
  author={Fan, Liuyi and Ai, Xinbo},
  journal={Knowledge-Based Systems},
  pages={115363},
  year={2026},
  publisher={Elsevier}
}

@article{fan2026channel,
  title={Channel-hierarchical graph convolutional network with semantic alignment for long-tailed multi-label image recognition},
  author={Fan, Liuyi and Ai, Xinbo},
  journal={Neurocomputing},
  pages={132832},
  year={2026},
  publisher={Elsevier}
}

@inproceedings{du2025category,
  title={Category-Specific Selective Feature Enhancement for Long-Tailed Multi-Label Image Classification},
  author={Du, Ruiqi and Tang, Xu and Zhang, Xiangrong and Ma, Jingjing},
  booktitle={Proceedings of the IEEE/CVF International Conference on Computer Vision},
  pages={3757--3766},
  year={2025}
}

@inproceedings{park2023robust,
  title={Robust asymmetric loss for multi-label long-tailed learning},
  author={Park, Wongi and Park, Inhyuk and Kim, Sungeun and Ryu, Jongbin},
  booktitle={Proceedings of the IEEE/CVF international conference on computer vision},
  pages={2711--2720},
  year={2023}
}

@article{zhang2025federated,
  title={Federated Chain Context Optimization for Long-Tailed Multi-Label Image Classification},
  author={Zhang, Libao and Zhu, Suxia and Yao, Wenjie and Sun, Guanglu},
  journal={IEEE Transactions on Mobile Computing},
  year={2025},
  publisher={IEEE}
}

@article{timmermann2025lm,
  title={LM-CLIP: Adapting Positive Asymmetric Loss for Long-Tailed Multi-Label Classification},
  author={Timmermann, Christoph and Jung, Seunghyeon and Kim, Miso and Lee, Woojin},
  journal={IEEE Access},
  year={2025},
  publisher={IEEE}
}

@article{tang2025unleashing,
  title={Unleashing the Power of Vision-Language Models for Long-Tailed Multi-Label Visual Recognition},
  author={Tang, Wei and Wang, Zuo-Zheng and Zhang, Kun and Wei, Tong and Zhang, Min-Ling},
  journal={arXiv preprint arXiv:2511.20641},
  year={2025}
}

@article{mikolov2013efficient,
  title={Efficient estimation of word representations in vector space},
  author={Mikolov, Tomas and Chen, Kai and Corrado, Greg and Dean, Jeffrey},
  journal={arXiv preprint arXiv:1301.3781},
  year={2013}
}

@article{zhu2026autoit,
  title={AutoIT: Automated Image Tagging with Random Perturbation: X. Zhu et al.},
  author={Zhu, Xuelin and Li, Jianshu and Liu, Jian and Tang, Dongqi and Ge, Jiawei and Liu, Weijia and Liu, Bo and Cao, Jiuxin},
  journal={International Journal of Computer Vision},
  volume={134},
  number={3},
  pages={110},
  year={2026},
  publisher={Springer}
}

@inproceedings{zhou2022conditional,
  title={Conditional prompt learning for vision-language models},
  author={Zhou, Kaiyang and Yang, Jingkang and Loy, Chen Change and Liu, Ziwei},
  booktitle={Proceedings of the IEEE/CVF conference on computer vision and pattern recognition},
  pages={16816--16825},
  year={2022}
}

@inproceedings{yang2015pinterest,
  title={Pinterest board recommendation for twitter users},
  author={Yang, Xitong and Li, Yuncheng and Luo, Jiebo},
  booktitle={Proceedings of the 23rd ACM international conference on Multimedia},
  pages={963--966},
  year={2015}
}

@article{wei2021fine,
  title={Fine-grained image analysis with deep learning: A survey},
  author={Wei, Xiu-Shen and Song, Yi-Zhe and Mac Aodha, Oisin and Wu, Jianxin and Peng, Yuxin and Tang, Jinhui and Yang, Jian and Belongie, Serge},
  journal={IEEE transactions on pattern analysis and machine intelligence},
  volume={44},
  number={12},
  pages={8927--8948},
  year={2021},
  publisher={IEEE}
}

@article{wei2025delving,
  title={Delving deep into simplicity bias for long-tailed image recognition},
  author={Wei, Xiu-Shen and Sun, Xuhao and Shen, Yang and Wang, Peng},
  journal={International Journal of Computer Vision},
  volume={133},
  number={6},
  pages={3349--3366},
  year={2025},
  publisher={Springer}
}

@inproceedings{cartucho2018robust,
  title={Robust object recognition through symbiotic deep learning in mobile robots},
  author={Cartucho, Joao and Ventura, Rodrigo and Veloso, Manuela},
  booktitle={2018 IEEE/RSJ international conference on intelligent robots and systems (IROS)},
  pages={2336--2341},
  year={2018},
  organization={IEEE}
}

\end{document}


\title{Appendix\\Rethinking Multi-Label Image Classification With Deep Learning: Taxonomy, Challenge, and Outlook}

\author{

{
Xuelin Zhu$^{\orcidlink{0000-0001-7676-2843}}$, 
Xiu-Shen Wei$^{\orcidlink{0000-0002-8200-1845}}$, 
Jiawei Ge$^{\orcidlink{0000-0001-7268-7815}}$, 
Shuai Xu$^{\orcidlink{0000-0002-5734-3616}}$, 
Bing Wang$^{\orcidlink{0000-0003-0977-0426}}$
}

\thanks{This work is supported by the National Natural Science Foundation of China (42301520), the Research Grants Council of Hong Kong (25206524, 15212925), the Unmanned Autonomous Systems Research Centre (P0049516), and the Smart Cities Research Institute (P0051028). \textit{(Corresponding author: Bing Wang.)}}

\thanks{Xuelin Zhu and Bing Wang are with the Spatial Intelligence Group, The Hong Kong Polytechnic University, Hong Kong, China (e-mail: zhuxuelin23@gmail.com; bingwang@polyu.edu.hk).}

\thanks{Xiu-Shen Wei is with the School of Computer Science and Engineering, Southeast University, Nanjing, China (e-mail: weixs@seu.edu.cn).}

\thanks{Jiawei Ge is with the School of Cyber Science and Engineering, Southeast University, Nanjing, China (e-mail: jiawei\_ge@seu.edu.cn).}

\thanks{Shuai Xu is with the College of Computer Science and Technology, Nanjing University of Aeronautics and Astronautics, Nanjing, China (e-mail: xushuai7@nuaa.edu.cn).}


}
\markboth{}%
{Shell \MakeLowercase{\textit{et al.}}: A Sample Article Using IEEEtran.cls for IEEE Journals}


\maketitle

\subsection{Differences With Earlier Publications}
Over the past decade, a large body of work have been devoted to multi-label image classification (MLIC), largely driven by the emergence of diverse deep learning techniques, yet the literature remains fragmented and loosely uncoordinated. Existing surveys \cite{liu2021emerging,wever2021automl,tarekegn2021review,wei2021fine,bogatinovski2022comprehensive,wei2022survey,chen2022survey,han2023survey,qian2023survey,tarekegn2024deep,holste2024towards} review multi-label classification across broad domains spanning computer vision, natural language processing, and data mining. However, they typically concentrate on specific problem settings, (e.g., imbalanced learning \cite{tarekegn2021review}, semi-supervised learning \cite{han2023survey}, extreme classification \cite{wei2022survey}), particular technique paradigms (e.g., automated machine learning \cite{wever2021automl}, feature selection \cite{qian2023survey}), or specialized application areas (e.g., chest X-rays \cite{holste2024towards}). As a result, they do not provide a systematic and in-depth treatment of MLIC itself, including its model designs, learning mechanisms, and unique challenges.

The only survey \cite{devkar2017survey} dedicated exclusively to MLIC is now considerably outdated. It covers merely 37 studies published before 2017 and therefore overlooks the explosive growth of deep learning–based MLIC methods, the advent of large-scale benchmarks, and the recent shift toward transformer architectures. In particular, it does not capture the open problems, key challenges, and promising research directions specific to MLIC, nor does it offer a forward-looking perspective on how this field may develop in the era of large language models and multimodal foundation models. These limitations highlight an urgent need for a comprehensive and up-to-date survey of MLIC that can synthesize recent advances, clarify open challenges, and offer coherent guidance for future research and practical system design.

\subsection{Datasets\label{sec:datasets}}

Table \ref{tab:datasets} summarizes the commonly used datasets in MLIC. 
Pascal VOC \cite{everingham2010pascal} is a foundational dataset in CV that provides well-annotated labels for MLIC. MS-COCO \cite{lin2014microsoft} is the most popular benchmark in MLIC, offering large-scale images with high-quality annotations. In practice, its validation set is used as a test set for performance evaluation. NUS-WIDE \cite{chua2009nus} extends MLIC to large-scale, real-world web images accompanied by user-generated tags. VG500 \cite{chen2019learning} is constructed by selecting images with the 500 most frequent labels from Visual Genome \cite{krishna2017visual}.
WIDER-Attribute \cite{li2016human} focuses on human-centric attributes, providing comprehensive data for more nuanced and detailed classification tasks.
PA-100K \cite{yazici2020orderless} caters specifically to fashion-related attributes, driving innovation in fashion-specific applications and recommendation systems.
Open Images (v6) \cite{kuznetsova2020open} is a large-scale dataset partially annotated with human labels and machine-generated labels, featuring diverse object classes and supporting broad exploration across various visual domains.

\begin{table}[t]
\centering
\caption{The commonly used datasets in the MLIC task.}
\begin{tabularx}{\linewidth}{X<{\centering}m{1cm}<{\centering}m{1cm}<{\centering}m{1cm}<{\centering}m{0.6cm}<{\centering}}
  \toprule
    Dataset &  \#Train & \#Valid & \#Test  & \#Label \\
  \midrule
    Pascal VOC 2007 \cite{everingham2010pascal} & 5,011 & - & 4,952 & 20 \\
    Pascal VOC 2012 \cite{everingham2010pascal} & 11,540 & - & 10,991 & 20 \\
    MS-COCO 2014 \cite{lin2014microsoft} & 82,081 & 40,137 & - & 80 \\
    NUS-WIDE \cite{chua2009nus} & 150,000 & - & 59,347 & 81 \\
    VG500 \cite{chen2019learning} & 82,904 & - & 10,000 & 500 \\
    PA-100K \cite{yazici2020orderless} & 80,000 & 10,000 & 10,000 & 26 \\
    WIDER-Attribute \cite{li2016human} & 28,345 & - & 29,179 & 14 \\
    Open Images (v6) \cite{kuznetsova2020open} & 9,000,000 & 41,620 & 125,436 & 9,600 \\
  \bottomrule
\end{tabularx}
\label{tab:datasets}
\end{table}

\subsection{Backbones\label{sec:backbones}}

Most MLIC frameworks are built on backbone networks pretrained on ImageNet \cite{deng2009imagenet}. AlexNet \cite{krizhevsky2012imagenet} marks an early milestone, delivering the first major breakthrough on ImageNet, while VGG \cite{simonyan2015very} increases network depth with small convolutional filters to obtain stronger generic visual features; both are widely adopted in early MLIC methods for feature extraction. Then, ResNet \cite{he2016deep} reshapes backbone design with residual connections, enabling very deep architectures and quickly becoming the default choice for MLIC, along with its variants such as ResNeXt \cite{xie2017aggregated} and TResNet \cite{ridnik2021tresnet}. Inspired by the success of Transformer \cite{vaswani2017attention} in NLP, Vision Transformer (ViT) \cite{dosovitskiy2020image} treats an image as a patch sequence and uses self-attention to model their dependencies, and has become a popular backbone for MLIC, with variants like Swin Transformer \cite{liu2021swin} further boosting performance. Overall, these backbone families underpin modern MLIC approaches by providing increasingly powerful image representations.

\subsection{Evaluation Metrics}
\label{sec:metrics}

In MLIC, the most common evaluation metrics are the average precision (AP) per label and the mean average precision (mAP) over all labels, measuring label-wise performance and overall performance, respectively. Given predicted scores $\mathbf{s}_i\in\mathbb{R}^N$ for label $c_i$, AP and mAP are defined as:
\begin{gather}
    \text{AP}_i = \sum_{k=1}^MP_k(\mathbf{s}_i)(R_k(\mathbf{s}_i)-R_{k-1}(\mathbf{s}_i)), \\
    \text{mAP} = \frac{1}{N}\sum_{i=0}^{N-1}\text{AP}_i,
\end{gather}
where $P_k(\cdot)$ and $R_k(\cdot)$ are the precision and recall for label $c_i$ based on the top-$k$ scores from $\mathbf{s}_i$ in descending order.

In addition, the precision (P), recall (R), and F1-measure (F1) under the settings of top-3 highest scores and all scores are adopted for evaluation. Both overall (OP, OR, OF1) and per-class (CP, CR, CF1) metrics are computed as:
\begin{equation}
\begin{aligned}
    \mathrm{CP}&=\frac{1}{C}\sum_i\frac{N_i^c}{N_i^p}, \qquad &\mathrm{OP}&=\frac{\sum_iN_i^c}{\sum_iN_i^p}, \\
    \mathrm{CR}&=\frac{1}{C}\sum_i\frac{N_i^c}{N_i^g},  &\mathrm{OR}&=\frac{\sum_iN_i^c}{\sum_iN_i^g}, \\
    \mathrm{CF1}&=\frac{2\times\mathrm{CP}\times\mathrm{CR}}{\mathrm{CP}+\mathrm{CR}}, &\mathrm{OF1}&=\frac{2\times\mathrm{OP}\times\mathrm{OR}}{\mathrm{OP}+\mathrm{OR}},
\end{aligned}
\end{equation}
where $C$ is the number of labels; $N_i^c$, $N_i^p$, and $N_i^g$ are the numbers of correctly predicted, predicted, and ground-truth images for label $c_i$, respectively. To handle images with varying label counts, it is common to report these metrics by treating a label as positive if its score exceeds a threshold. Among all metrics, mAP, OF1, and CF1 are the most important.\renewcommand{\tabcolsep}{1.2pt}
\newcolumntype{Y}{p{0.68cm}<{\centering}}

\bibliographystyle{IEEEtran}
\bibliography{reference}